\documentclass[preprint,twoside,11pt,theapa]{article}
\usepackage{preprint, theapa, rawfonts}
\usepackage[implicit=false]{hyperref}
\usepackage{amsmath,amsfonts, mathtools, bm}
\usepackage[dvipsnames]{xcolor}
\usepackage{colortbl}
\usepackage{algorithm}
\usepackage{algorithmic}
\usepackage{makecell}
\usepackage{tikz}
\usepackage{pifont}
\usepackage{pgfplots}
\usepackage{booktabs, enumitem}
\pgfplotsset{compat=1.7}
\usepackage{bbm}
\usepackage{subcaption} 
\usetikzlibrary{arrows,matrix, positioning, shapes.geometric}
\usepackage{listofitems} 
\tikzstyle{mynode}=[thick,inner sep={.75\pgflinewidth}, draw=black,fill=white,circle,minimum size=3]
\jairheading{1}{2024}{}{}{}
\ShortHeadings{Decision-Focused Learning: A Survey}{Mandi, Kotary, Berden, Mulamba, Bucarey, Guns, \& Fioretto}
\firstpageno{1}
 
\DeclareMathOperator*{\argmax}{argmax}
\DeclareMathOperator*{\argmin}{argmin}
\newcommand{\pto}{\textit{Predict-Then-Optimize}\ }

\newcommand{\coloredasterisk}[1]{\textcolor{#1}{\ding{81}}}
\newcommand{\coloredx}[1]{\textcolor{#1}{\ding{54}}}
\newcommand{\coloredplus}[1]{\textcolor{#1}{\ding{58}}}
\newcommand{\coloredtriangle}[1]{\textcolor{#1}{\ding{115}}}
\newcommand{\coloredtriangledown}[1]{\textcolor{#1}{\ding{116}}}
\newcommand{\coloredtickmark}[1]{\textcolor{#1}{\ding{52}}}
\newcommand{\coloredcross}[1]{\textcolor{#1}{\ding{56}}}
\newcommand{\rev}[1]{{\color{purple}{#1}}}

\newcommand{\jay}[1]{{\color{blue}{jay: #1}}}
\newcommand{\senne}[1]{{\color{red}{Senne: #1}}}
\newcommand{\victor}[1]{{\color{RoyalPurple}{Victor: #1}}}
\newcommand{\nando}[1]{{\color{purple}{\textbf{[}#1\textbf{]}}}}
\newcommand{\tias}[1]{{\color{teal}{Tias: #1}}}

\firstpageno{1}
\begin{document}

\title{Decision-Focused Learning: Foundations, State of the Art, Benchmark and Future Opportunities}

\author{\name Jayanta Mandi {\thanks{JM was affiliated to Vrije Universiteit Brussel,
	Belgium during the submision of this article.}} \email jayanta.mandi@kuleuven.be \\
	\addr KU Leuven,
	Belgium
	\AND
	\name James Kotary {\thanks{JM and JK should both be considered first authors.}} \email jk4pn@virginia.edu \\
	\addr University of Virginia, 
	USA
	\AND
	\name Senne Berden \email senne.berden@kuleuven.be \\
	\addr KU Leuven,
	Belgium
	\AND
	\name Maxime Mulamba \email maxime.mulamba@vub.be \\
	\addr Vrije Universiteit Brussel,
	Belgium
	\AND
	\name V\'ictor Bucarey \email victor.bucarey@uoh.cl \\
	\addr Universidad de O'Higgins,
	Chile
	\AND
	\name Tias Guns \email tias.guns@kuleuven.be \\
	\addr KU Leuven,
	Belgium
	\AND
	\name Ferdinando Fioretto \email fioretto@virginia.edu \\
	\addr University of Virginia,
	USA
	}


\maketitle
\begin{abstract}
\emph{Decision-focused learning} (DFL) is an emerging paradigm that integrates machine learning (ML) and constrained optimization to enhance decision quality by training ML models in an end-to-end system. This approach shows significant potential to revolutionize combinatorial decision-making in real-world applications that operate under uncertainty, where estimating unknown parameters within decision models is a major challenge. This paper presents a comprehensive review of DFL, providing an in-depth analysis of both gradient-based and gradient-free techniques used to combine ML and constrained optimization. It evaluates the strengths and limitations of these techniques and includes an extensive empirical evaluation of eleven methods across seven problems. The survey also offers insights into recent advancements and future research directions in DFL.

\noindent
{\bf Code and benchmark}: \url{https://github.com/PredOpt/predopt-benchmarks}
\end{abstract}


\section{Introduction}
Real-world applications frequently confront the task of decision-making under uncertainty, such as planning the shortest route in a city, determining optimal power generation schedules, or managing investment portfolios \cite{SAHINIDIS2004971,liu2009theory,kim2005optimal,hu2016toward,delage2010distributionally,hhl003}. In such scenarios, estimating unknown parameters often poses a significant challenge. 

\begin{figure}
    \centering



  \includegraphics[width=0.8\textwidth]{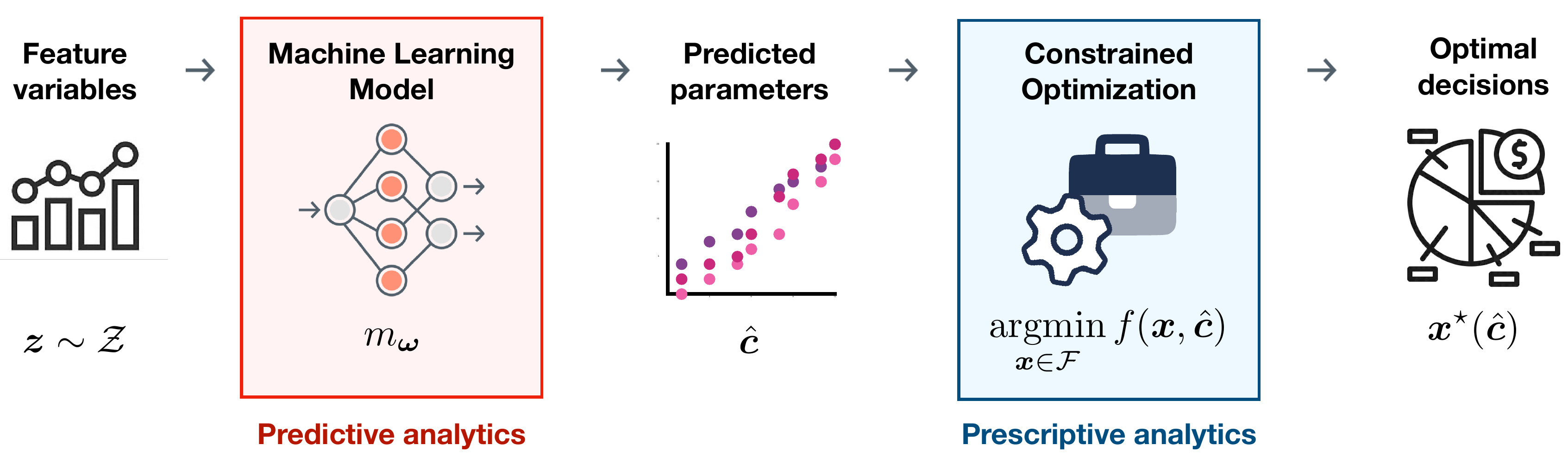}
    \caption{Decision-making under uncertainty involves both predictive and prescriptive analytics. In the predictive stage, the uncertain parameters are predicted from the features using an ML model. In the prescriptive stage, a decision is prescribed by solving a CO problem using the predicted parameters.
    }
    \label{fig:problem_desc}
\end{figure}

Machine Learning (ML) and Constrained Optimization (CO) serve as two key tools for these complex problems. ML models estimate uncertain quantities, while CO models optimize objectives within constrained spaces. This sequential process, commonly referred to as \emph{predictive} and \emph{prescriptive} modeling, as illustrated in Figure~\ref{fig:problem_desc}, is prevalent in fields like operations research and business analytics \cite{den2016bridging}. For instance, in portfolio management, the prediction stage forecasts asset returns, while the prescriptive phase optimizes returns based on these predictions.
The terminology \pto problem will be used in this survey paper to refer to the problem setting where the uncertain parameter has to be predicted first, followed by solving the CO problem using the predicted parameter to make a decision.

A commonly adopted approach to tackle \pto problems involves handling these two stages---prediction and optimization---separately and independently. 
This ``two-stage'' process first involves training an ML model to create a mapping between observed features and the relevant parameters of a CO problem.
Subsequently, and independently, a specialized optimization algorithm is used to solve the decision problem, which is specified by the predicted problem parameters.
The underlying assumption in this methodology is that superior predictions would lead to precise predictive models and consequently, high-quality decisions. Indeed, if the predictions of the parameters were perfectly accurate, they would enable the correct specification of CO models which can be solved to yield fully optimal decisions.
However, ML models often fall short of perfect accuracy, leading to suboptimal decisions due to propagated prediction errors. Thus, in many applications, the predictive and prescriptive modelings are not isolated but rather, deeply interconnected, and hence should ideally be modeled jointly. 

This is the goal of the \textbf{decision-focused learning} (DFL) paradigm, which directly trains the ML model to make predictions that lead to good decisions. In other words, DFL integrates prediction and optimization in an end-to-end system trained to optimize a criterion (i.e., a loss function) that is based on the resulting decisions.

Since many ML models, including neural networks (NNs), are trained via gradient-based optimization, the gradients of the loss must be backpropagated through each constituent operation of the model.  
In DFL, the loss function is dependent on the solution of the CO problem, thus the CO solver is \emph{embedded} as a component of the ML model. In this integration of prediction and optimization, a key challenge is \emph{differentiating through the optimization problem}. 
An additional challenge arises from decision models operating on discrete variables, which produce discontinuous mappings and hinder gradient-based learning. Hence, examining \emph{smooth} surrogate models for these discrete mappings, along with their differentiation, becomes crucial. These two challenges are the core emphasis and central focal points in DFL.

This survey paper presents a comprehensive overview of decision-focused learning and makes several key contributions. First, to navigate the complex methodologies developed in recent years, the survey differentiates gradient-based DFL methodologies from `gradient-free' methodologies,
which do not rely on computing gradients for learning.
Given their compatibility with neural networks, which are the predominant ML architectures, there has been a greater focus on gradient-based DFL methods.
To facilitate a comprehensive understanding of this field, we propose categorizing gradient-based learning methods into four distinct classes:
\textbf{(1)} analytical differentiation of optimization mappings, \textbf{(2)} analytical smoothing of optimization mappings, \textbf{(3)} smoothing by random perturbations,
and \textbf{(4)} differentiation of surrogate loss functions.
This categorization, illustrated in Figure \ref{Fig:DFLMethodologyTable} lower in the paper, serves as a framework for comprehending and organizing various gradient-based DFL methodologies.

In the second part, this paper compiles a selection of problem-specific DFL models, making them publicly available to facilitate broader access and usage.
As part of that, we benchmark the performance of the various methods on \emph{seven}
distinct problems. This provides an opportunity for comparative understanding and assists in identifying the relative strengths and weaknesses of each approach.
The code and data used in the benchmarking are accessible through \url{https://github.com/PredOpt/predopt-benchmarks}.
Finally, this survey looks forward and discusses open challenges and offering an outlook on potential future directions in the field of DFL.

\paragraph{Positioning with respect to other overview papers.}
With the growing interest of the operations research (OR) and artificial intelligence (AI) communities in integrating ML and CO, various survey, review, and tutorial papers have emerged.  
\citeA{MisicINFORMS} provide applications of \pto problems in various fields of OR, but offer limited discussion on DFL methodologies.
DFL is also a part of a short review article by \citeA{kotary2021end}, where the primary focus is on recent developments in differentiable optimization to integrate CO problems into neural network architectures.
The tutorial by \citeA{Qi2022Integrating} introduces the notion of DFL to the OR community and discusses
methods that fall under our category of \emph{differentiating surrogate loss functions} below.
The PyEPO library \cite{pyepo} offers a common code base and an implementation of a selected number of DFL techniques, along with a common interface for benchmarking on various datasets. 
\citeA{pyepo} also provide an overview of the DFL techniques implemented in the library, but offer limited discussion on the broader literature of existing DFL techniques.
Finally the recent survey by~\citeA{sadana2023survey}, which appeared online concurrently with the submission of this article, proposes the umbrella term of \emph{contextual stochastic optimization} and discusses three families of techniques namely decision rule optimization, sequential learning and (stochastic) optimization and integrated learning and (stochastic) optimization; with DFL belonging to the latter. 

%
In contrast, our survey offers a distinct perspective by adopting an ML oriented overview that extensively surveys gradient-based and gradient-free DFL techniques for \pto problems, where uncertain parameters appear in the objective function of the CO problems. The proposed categorization of gradient-based DFL techniques into four classes allows for a systematic discussion and comparison of the different methods. Moreover, this article stands out as an experimental survey by performing comparative evaluations of some widely-used DFL techniques on benchmark problems; thereby highlighting the feasibility of implementing, comparing and deploying practical DFL methods. The aim of this dual focus is to enhance understanding of both specific DFL techniques while also showing empirical insights into their effectiveness across various applications.




\paragraph{Paper organization.}
Following this introduction, the paper is structured as follows. Preliminary concepts are discussed in Section \ref{sect:prelim}, which introduces the problem setting and explains the challenges in implementing DFL. The subsequent Section \ref{sect:review} offers a comprehensive review of existing gradient-free and gradient-based DFL methodologies for handling these challenges, further categorizing the gradient-based methodologies into four distinct classes. 
Section \ref{sect:applications} presents interesting real-world examples of DFL applications.
Section \ref{sect:data} brings forth seven benchmark DFL tasks from public datasets, with a comparative evaluation of eleven DFL techniques. Finally, the survey paper concludes by providing a discourse on the current challenges and possible future directions in DFL research.

\section{Preliminaries}\label{sect:prelim}

This section presents an overview of the problem setting, along with preliminary concepts and essential terminology. Then, the central modeling challenges are discussed, setting the stage for a review of current methodologies in the design and implementation of DFL.
Throughout the survey paper, vectors are denoted by boldface lowercase letters, such as $\mathbf{x}$, while scalar components within the vector $\mathbf{x}$ are represented with a subscript $i$, denoting the $i^{\text{th}}$ item within $\mathbf{x}$ as $x_i$. Similarly, the vectors $\mathbf{1}$ and $\mathbf{0}$ symbolize the vector of all-ones and all-zeros, respectively.
Moreover, $\mathbf{I}$ denotes an identity matrix of appropriate dimension.

\subsection{Problem Setting}
\label{subsec:problem_description}
In operations research and business analytics, decisions are often quantitatively modeled using CO problems. 
In many real-world applications, it happens that some parameters of the CO problem are uncertain and must be inferred from contextual data (hereafter referred to as \emph{features}).
The settings considered in this survey paper involve estimating those parameters through predictive inferences made by ML models, and subsequently, the final decisions are modeled as the solution to the CO problems based on those inferences. 
%
In this setting, the decision-making processes can be described by \emph{parametric} CO problems, defined as,
\begin{subequations}
\label{eq:opt_generic}
\begin{align}
    \label{eq:opt_generic_obj}
    \mathbf{x}^\star( \mathbf{c} ) = \argmin_{ \mathbf{x}} &\;\;
    f(\mathbf{x}, \mathbf{c})\\
    \label{eq:opt_generic_ineq}
    \texttt{s.t.} &\;\;
    \bm{g}( \mathbf{x}, \mathbf{c}) \leq \mathbf{0}\\
    \label{eq:opt_generic_eq}
    &\;\; \bm{h}(\mathbf{x}, \mathbf{c}) = \mathbf{0}.
\end{align}
\end{subequations}
The goal of the CO problem above is to find \emph{a} solution $\mathbf{x}^\star(\mathbf{c}) \in \mathbb{R}^n$ ($n$ being the dimension of the decision variable $x$), which minimizes the objective function $f$, subject to equality and inequality constraints as defined by the functions $\bm{h}$ and $\bm{g}$.
This \emph{parametric} problem formulation defines $\mathbf{x}^\star(\mathbf{c})$ as a function of the parameters $\mathbf{c} \in \mathbb{R}^m$. 

CO problems can be categorized in terms of the forms taken by the functions defining their objectives (\ref{eq:opt_generic_obj}) and constraints (\ref{eq:opt_generic_ineq}-\ref{eq:opt_generic_eq}). These forms also determine important properties of the optimization mapping $\mathbf{c} \rightarrow \mathbf{x}^\star(\mathbf{c})$, when viewed as a function from problem parameters to optimal solutions, such as its continuity, differentiability, and injectivity.

In this survey paper, it is assumed that the constraints are fully known prior to solving, i.e., $\bm{h}(\mathbf{x},\mathbf{c}) = \bm{h}(\mathbf{x})$ and $\bm{g}( \mathbf{x},\mathbf{c}) = \bm{g}( \mathbf{x})$, indicating the constraints do not depend on the parameter $\mathbf{c}$, which is uncertain.
Rather the dependence on $\mathbf{c}$ is restricted solely to the objective function. 
This is the setting considered by almost all existing works surveyed. While it is also possible to consider uncertainty in the constraints, this leads to the possibility of predicting parameters that lead to solutions that are infeasible with respect to the ground-truth parameters. The learning problem has not yet been well-defined in this setting (unless a recourse action to correct infeasible solutions is used \cite{PredictOptimizeforPacking,hu2023branch}). For this reason, in the following sections, only $f$ is assumed to depend on $\mathbf{c}$, so that $\bm{g}(\mathbf{x}) \leq \mathbf{0}$ and $\bm{h}(\mathbf{x}) = \mathbf{0}$ are satisfied for all outputs of the decision model. For notational convenience, the feasible region of the CO problem in \eqref{eq:opt_generic}, will be denoted by $\mathcal{F}$ (i.e., $\mathbf{x} \in \mathcal{F}$ if and only if $\bm{g}(\mathbf{x}) \leq \mathbf{0}$ and $\bm{h}(\mathbf{x}) = \mathbf{0}$).

If the true parameters $\mathbf{c}$ are known exactly, the corresponding `true' optimal decisions may be computed by solving \eqref{eq:opt_generic}. In such scenarios, $\mathbf{x}^\star (\mathbf{c})$ will be referred to as the \emph{full-information optimal decisions} \cite{bertsimas2020predictive}. 
This paper, instead, considers problems where the parameters $\mathbf{c}$ are unknown but can be estimated as a function of observed features $\mathbf{z}$. The problem of estimating $\mathbf{c}$ falls under the category of supervised ML problems. In this setting, a set of past observation pairs $\{(\mathbf{z_i}, \mathbf{c_i})\}_{i=1}^N$ is available as a training dataset, $\mathcal{D}$, and used to train an ML model $m_{\bm{\omega}}$ (with trainable ML parameters $\bm{\omega}$), so that parameter predictions take the form $\mathbf{\hat{ \mathbf{c}}} = m_{\bm{\omega}}( \mathbf{z})$. Then, a decision $\mathbf{x}^\star ( \mathbf{ \hat{c} })$ can be made based on the predicted parameters. $\mathbf{x}^\star ( \mathbf{ \hat{c} })$  is referred to as a \emph{prescriptive decision}. 
The overall learning goal is to optimize the set of prescriptive decisions made over a distribution of features $\mathbf{z} \sim \mathcal{Z}$, with respect to some evaluation criterion on those decisions. Thus, while the ML model $m_{\bm{\omega}}$ is trained to predict $\mathbf{\hat{c}}$, its performance is evaluated based on $\mathbf{x}^\star ( \mathbf{\hat{c}})$.
This paper uses the terminology \pto problem to refer to the problem of predicting $\mathbf{\hat{c}}$, to improve the evaluation of $\mathbf{x}^\star ( \mathbf{\hat{c}})$.

\subsection{Learning Paradigms}\label{sect:learningparadigm}
The defining challenge of the \pto problem setting is the gap in modeling between the prediction and the optimization components: while $m_{\bm{\omega}}$ is trained to predict $\mathbf{\hat{c}}$, it is evaluated based on the subsequently computed $\mathbf{x}^\star(\mathbf{\hat{c}})$. 
Standard ML approaches, based on the \emph{empirical risk minimization} (ERM) \cite{Vapnik99}, use standard loss functions $\mathcal{L}$, such as mean squared error or cross-entropy, in order to learn to predict $\mathbf{\hat{c}} = m_{\bm{\omega}}(\mathbf{z})$.
The learning is supervised by by the ground-truth $\mathbf{c}$.
%
However, in principle, for \pto problems, it is desirable to train $m_{\bm{\omega}}$ to make predictions $\mathbf{\hat{c}}$ that optimize the evaluation criterion on $\mathbf{x}^{\star}(\mathbf{\hat{c}})$ directly. This distinction motivates the definition of two alternative learning paradigms for \pto problems.   

\paragraph{Prediction-focused learning (PFL).}  A straightforward approach to this supervised ML problem is to train the model to generate accurate  parameter predictions $\mathbf{\hat{c}}$ with respect to ground-truth values $\mathbf{c}$.
This paper introduces the term \emph{prediction-focused learning} to refer to this approach (also called two-stage learning \cite{aaai/WilderDT19}) because the model is trained with a focus on the accuracy of the parameter predictions preceding the decision model. Here, the training is agnostic of the downstream CO problem. At the time of making the decision, the pre-trained model's predictions $\mathbf{\hat{c}}$ are passed to the CO solvers which solve \eqref{eq:opt_generic} to return $\mathbf{x}^\star (\mathbf{\hat{c}})$. Typical ML losses, such as the mean squared error (MSE) or binary cross entropy (BCE), are used to train the prediction model in this case.
\begin{equation}
    MSE\; (\mathbf{\hat{c}}, \mathbf{c}) = \frac{1}{N} \| \mathbf{c} - \mathbf{\hat{c}} \| ^2
    \label{eq:prediction_focused}
\end{equation}
\noindent 
Such loss functions, like Eq. \eqref{eq:prediction_focused}, which measure the prediction error of $\mathbf{\hat{c}}$ with respect to $\mathbf{c}$, are referred to as \textit{prediction losses}.
Algorithm~\ref{alg:prediction_focused} illustrates PFL with MSE loss.

\paragraph{Decision-focused learning (DFL).} 
By contrast, in \textit{decision-focused} learning, the ML model is trained to optimize the evaluation criteria which measure the quality of the resulting decisions. As the decisions are realized after the optimization stage, this requires the integration of prediction and optimization components, into a composite framework which produces full decisions. From this point of view, generating the predicted parameters $\mathbf{\hat{c}}$ is an intermediary step of the integrated approach, and the accuracy of $\mathbf{\hat{c}}$ is not the primary focus in training. The focus, rather, is on the error incurred after optimization. A measure of error with respect to the integrated model's prescriptive decisions, when used as a loss function for training, is henceforth referred to as a \textit{task loss}. The essential difference from the aforementioned prediction loss is that it measures the error in $\mathbf{x}^{\star}(\mathbf{\hat{c}})$, rather than in $\mathbf{\hat{c}}$.

The objective value achieved by using the predicted $\mathbf{x}^\star (\mathbf{\hat{c}})$ is generally suboptimal
 with respect to the true objective parameters $\mathbf{c}$. Often, the end goal is to generate predictions $\mathbf{\hat{c}}$ with an optimal solution $\mathbf{x}^\star (\mathbf{\hat{c}})$ whose objective value in practice (i.e., $f(\mathbf{x}^\star(\mathbf{\hat{c}}),\mathbf{ c})$) comes close to the full-information optimal value $f(\mathbf{x}^\star( \mathbf{c}), \mathbf{c})$. In such cases, a salient notion of task loss is the \emph{regret}, defined as  the difference between the full-information optimal objective value and the objective value realized by the prescriptive decision. Equivalently, it is the magnitude of suboptimality of the decision $\mathbf{x}^\star(\mathbf{\hat{c}})$ with respect to the optimal solution $\mathbf{x}^\star( \mathbf{c})$ under ground-truth parameters $\mathbf{c}$:
\begin{equation}
\mathit{Regret} \; (\mathbf{x}^\star(\mathbf{\hat{c}}),\mathbf{ c })= f(\mathbf{x}^\star (\mathbf{\hat{c}}), \mathbf{c}) - f(\mathbf{x}^\star( \mathbf{c}), \mathbf{c})
\label{eq:regret}
\end{equation}
\noindent
Note that minimizing regret is equivalent to minimizing the value of $f(\mathbf{x}^\star (\mathbf{\hat{c}}), \mathbf{c})$, since the term $f(\mathbf{x}^\star( \mathbf{c}), \mathbf{c})$ is constant with respect to the prediction model.  While regret may be considered the quintessential example of a task loss, other task losses can arise in practice. For example, when the ground-truth target data are observed in terms of decision values $\mathbf{x}^\star$, rather than parameter values $\mathbf{c}$, they may be targeted using the typical training loss functions such as $MSE \; (\mathbf{x}^{\star}(\mathbf{\hat{c}}), \mathbf{x}^\star)$. 

\raggedbottom
 \paragraph{Relationship between prediction and task losses.} As previously mentioned, an ML model is trained without considering the downstream CO problem in prediction-focused learning for \pto tasks; still the ML model is evaluated at test time on the basis of its resulting CO problem solutions.
This is based on an underlying assumption that generating accurate predictions with respect to a standard prediction loss will result in good prescriptive decisions. Note that zero prediction loss always implies zero task loss, since $\mathbf{\hat{c}} = \mathbf{c}$ implies $\mathbf{x}^{\star}(\mathbf{\hat{c}}) = \mathbf{x}^{\star}(\mathbf{c})$. However, in practice, it is impossible to learn an ML model that makes no prediction error on any sample. 
The model error can only be minimized in one metric, and the minimization of the prediction error and the resulting decision error do not in general coincide \cite{demirovic2019investigation}. 
Furthermore, the prediction loss and the task loss are, in general, not  continuously related. These principles are illustrated by the following example.
\paragraph{Example.}  The shortcomings of training with respect to prediction errors can be illustrated with a simple CO problem. For this illustration, consider a knapsack problem \cite{pisinger1998knapsack}. 
The objective of the knapsack problem is to select a maximal-value subset from an overall set of items, each having its own value and unit weight, subject to a capacity constraint,
which imposes that the number of selected items cannot be higher than the capacity $C$. 
This knapsack problem with unit weights can be formulated as follows:
\begin{equation}
    \label{eq:unitknapsackformulation}
   \mathbf{x}^{\star}(\mathbf{c}) = \argmax_{\mathbf{x} \in \{0,1\} } \mathbf{c}^\top \mathbf{x} \;\; \texttt{s.t.} \sum_i x_i \leq \mathit{C}
\end{equation}

In a \pto variant of this knapsack problem, the item weights and knapsack capacity are known, but the item values are unknown and must be predicted using observed features.
The ground-truth item value $\mathbf{c}$ implies the ground-truth solution $\mathbf{x}^\star (\mathbf{c})$. Overestimating the values of the items that are chosen in $\mathbf{x}^\star (\mathbf{c})$ (or underestimating the values of the items that are not chosen) increases the prediction error. Note that these kind of prediction errors, even if they are high,  
do not affect the solution, and thus do not affect the task loss either.  On the other hand, even low prediction errors for some item values may change the solution, affecting the task loss. That is why after a certain point, reducing prediction errors does not decrease task loss, and sometimes may increase it. DFL aims to address this shortcoming of PFL: by minimizing the task loss directly, prediction errors are implicitly traded off on the basis of how they affect the resulting decision errors.

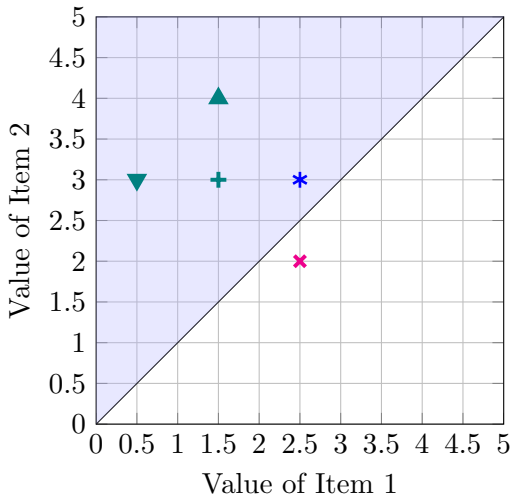
\begin{figure}
    \centering
\begin{tikzpicture}
\begin{axis}[
    xlabel={Value of Item 1 },
    ylabel={Value of Item 2},
    title={},
    xmin=0, xmax=5,
    ymin=0, ymax=5,
    grid=major,
    xtick distance=0.5,  
    ytick distance=0.5,  
    width=7cm,
    height=7cm,
]
\addplot[
    color=black,
]
coordinates {
    (0,0)
    (1,1)
    (2,2)
    (3, 3)
    (4, 4)
    (5,5)
    (6,6)
};
\begin{scope}
\clip (axis cs: 0,0) -- (axis cs: 6,6) -- (axis cs: 0,6) -- cycle;
\fill[blue!30, opacity=0.3] (axis cs:0,0) rectangle (axis cs:6,6);
\end{scope}

\addplot[
    color=blue,
    mark=asterisk,  
    mark size=3pt,  
     mark options={
        line width=1pt, 
    },
]
coordinates {
    (2.5,3)
};

\addplot[
    color=magenta,
    mark=x,  
    mark size=3pt, 
    mark options={
        line width=1.5pt,  
    },
]
coordinates {
    (2.5,2)
};

\addplot[
    color=teal,
    mark=+,  
    mark size=3pt, 
    mark options={
        line width=1.5pt,  
    },
]
coordinates {
    (1.5,3)
};

\addplot[
    color=teal,
    only marks, 
    mark=triangle*, 
    mark size=3pt,  
    mark options={
        line width=1.5pt,  
    },
]
coordinates {
    (1.5,4)
};
\addplot[
    color=teal,
    only marks, 
    mark=triangle*, 
    mark size=3pt,  
    mark options={
        line width=1.5pt,  
        rotate=180
    },
]
coordinates {
    (0.5,3)
};
\end{axis}
\end{tikzpicture}
\caption{An illustrative numerical example with a knapsack problem with two items to exemplify the discrepancy between prediction error and regret. 
The figure illustrates that two points can have the same prediction error but different regret.
Furthermore, it demonstrates that overestimating the values of the selected items or underestimating the values of the items that are left out does not change the solution, and thus does not increase the regret, even though the prediction error does increase.}
    \label{fig:knapsack_example}
\end{figure}


The discrepancy between the prediction loss and the task loss has been exemplified in Figure \ref{fig:knapsack_example} for a very simple knapsack problem with only two items. For this illustration, assume that both the items are of unit weights and the capacity of the knapsack is one, i.e., only one of the two items can be selected.
 The true values of the first and second items are $2.5$ and $3$ respectively. The point $(2.5,3)$, marked with \coloredasterisk{blue}, represents the true item values.
In this case the true solution is $(0,1)$, which corresponds to selecting only the second item. It is evident that any prediction in the blue shaded region leads to this solution. 
For instance, the point $(1.5,3)$, marked with \coloredplus{teal}, corresponds to predicting $1.5$ and $3$ as values of the two items respectively and this results in selecting the second item.
On the other hand, the point $(2.5,2)$, marked with \coloredx{magenta}, triggers the wrong solution $(1,0)$, although the squared error values of \coloredplus{teal} and \coloredx{magenta} are identical. 
Also, note that overestimating the value of the second item does not change the solution. For instance, the point $(1.5,4)$, marked with \coloredtriangle{teal}, corresponds to overestimating the value of the second item to $4$ while keeping the value of the first item the same as the point in \coloredplus{teal}.
 This point is positioned directly above the point in \coloredplus{teal} and still stays in the blue-shaded region.
 Similarly, the point $(0.5,3)$, marked with \coloredtriangledown{teal}, results from underestimating the value of the first item and is in the blue shaded region too. 
 Although these two points have higher values of squared error than the point marked with \coloredx{magenta}, they trigger the right solution, resulting in zero regret. 

\paragraph{Empirical risk minimization and bilevel form of DFL.} The minimization of either the expected prediction loss in PFL or the expected task loss in DFL, can be expressed as an Empirical Risk Minimisation (ERM) problem over a training dataset $\mathcal{D} \equiv \{(\mathbf{z_i}, \mathbf{c_i})\}_{i=1}^N$.
The desired objective of training is to learn a model, $m_{\bm{\omega}}$, that minimizes the \emph{expected loss}. However, as the joint probability distribution of $(\mathbf{z}, \mathbf{c})$ is often unknown, ERM minimizes the empirical loss instead, i.e., ERM minimizes the average loss, calculated over $\mathcal{D}$.

Note that ERM typically involves making a point prediction $\mathbf{\hat{c}}$ to minimize expected loss. However, besides point predictions, it is also possible to estimate the distribution of $\mathbf{\hat{c}}$. The advantage of distributional estimation is that it allows the decision-maker to consider extreme cases of the parameter distribution, leading to more robust decisions. However, minimizing distributionally robust loss \cite{levy2020dro} is inherently more challenging than ERM, even for standard ML problems. As a result, most of the surveyed works focus on minimizing expected task loss in DFL. 
Hence, learning techniques for minimizing \emph{expected} task loss are the main focus in this survey paper.

The respective ERM problems below assume the use of the MSE and regret loss functions for PFL and DFL respectively, but the principles described here hold for a wide range of alternative loss functions. 
PFL, by minimizing the prediction error with respect to the ground-truth parameters directly, takes the form of a standard regression problem:
\begin{equation}
    \min_{\bm{\omega}} \frac{1}{N} \sum_{i=1}^N \| m_{\bm{\omega}} (\mathbf{z_i})  - \mathbf{c_i} \|^2 \label{eq:ERM_PFL_outer},\\
\end{equation}
which is an instance of unconstrained optimization. In the case of DFL, it is natural to view the ERM as a bilevel optimization problem: 
\begin{subequations}
\label{eq:bilevel}
\begin{align}
    \min_{\bm{\omega}} \frac{1}{N} \sum_{i=1}^N \Big( f(\mathbf{x}^\star(\mathbf{\hat{c}_i}), \mathbf{c_i}) - f(  \mathbf{x}^\star(\mathbf{c_i}), \mathbf{c_i}) \Big) \label{eq:bilevel_outer}\\
    \texttt{s.t.} \;\; \mathbf{\hat{c}_i} = m_{\bm{\omega} }(\mathbf{z_i})  ;\ \mathbf{\mathbf{x}^\star}(\mathbf{ \hat{c}_i }) = \argmin_{ \mathbf{x} \in \mathcal{F}}   f(\mathbf{x}, \mathbf{\hat{c}_i}) 
   \label{eq:bilevel_inner}
\end{align}
\end{subequations}
The outer-level problem \eqref{eq:bilevel_outer} minimizes task loss on the training set while the inner-level problem \eqref{eq:bilevel_inner} computes the mapping $\mathbf{c} \to \mathbf{x}^{\star}(\mathbf{c})$. 
Solving \eqref{eq:bilevel} is computationally more challenging than solving \eqref{eq:ERM_PFL_outer} in the prediction-focused paradigm. In both cases, optimization by stochastic gradient descent (SGD) is the preferred method for training neural networks.

Algorithms~\ref{alg:prediction_focused} and~\ref{alg:decision_focused} compare the gradient descent training schemes for PFL and DFL. Algorithm ~\ref{alg:prediction_focused} is a standard application of gradient descent, in which the derivatives of Line \rev{6} are generally well-defined and can be computed straightforwardly (typically by automatic differentiation \cite{JMLR:autodiff}).
Line \rev{7} of Algorithm ~\ref{alg:decision_focused} shows that direct differentiation of the mapping $\mathbf{c} \to \mathbf{x}^{\star}(\mathbf{c})$ can be used to form the overall task loss gradient $\frac{d  \mathcal{L}} {d  \bm{\omega}}$,
by providing the required chain rule term $\frac{d  \mathbf{x}^\star( \mathbf{\hat{c}} )}{d  \mathbf{\hat{c}}}$. However, this differentiation is nontrivial as the mapping itself lacks a closed-form representation. Furthermore, many interesting and practical optimization problems are inherently nondifferentiable and even discontinuous as functions of their parameters, precluding the direct application of Algorithm ~\ref{alg:decision_focused} to optimize~\eqref{eq:bilevel} by gradient descent. The following subsections review the main challenges of implementing Algorithm~\ref{alg:decision_focused}.

\subsection{Challenges to Implement Decision-Focused Learning}
\label{sect:challenge}
\paragraph{Differentiation of CO mappings.} When considering gradient-based learning techniques, one needs to minimize the task loss by backpropagating the error over it. Hence, the partial derivatives of the task loss with respect to the prediction model parameters $\bm{\omega}$ must be computed to carry out the parameter update at Line \rev{7} of Algorithm \ref{alg:decision_focused}. Since the task loss $\mathcal{L}$ is a function of $\mathbf{x}^\star ( \mathbf{ \hat{c}} )$, the gradient of $\mathcal{L}$ with respect to $\bm{\omega}$ can be expressed in the following terms by using the chain rule of differentiation:
\begin{equation}
    \frac{d \mathcal{L}(\mathbf{x}^\star(\mathbf{\hat{c}}), \mathbf{c} )}{d \bm{\omega}} = \frac{d  \mathcal{L}(\mathbf{x}^\star (\mathbf{\hat{c}}), \mathbf{c})}{d  \mathbf{x}^\star (\mathbf{\hat{c}})} \frac{ d \mathbf{x}^\star(\mathbf{\hat{c}}) }{ d  \mathbf{\hat{c}}} \frac{d \mathbf{ \hat{c}}}{d  \bm{\omega}}
    \label{eq:chain_rule}
\end{equation}
The first term in the right side of \eqref{eq:chain_rule}, can be computed directly as $\mathcal{L}(\mathbf{x}^\star(\mathbf{\hat{c}}), \mathbf{c})$ is typically a differentiable function of $\mathbf{x}^\star (\mathbf{\hat{c}})$. 
A deep learning library (such as 
PyTorch \cite{paszke2019pytorch}) computes the last term by representing the neural network as a computational graph and applying automatic differentiation (autodiff) in the reverse mode \cite{JMLR:autodiff}.
However, the second term, $\frac{ d  \mathbf{x}^\star (\mathbf{\hat{c}})}{ d  \mathbf{\hat{c}}} $, may be nontrivial to compute given the presence of two major challenges: \textbf{(1)} The mapping $\mathbf{\hat{c}} \to \mathbf{x}^\star (\mathbf{\hat{c}})$, as defined by the solution to an optimization problem, \emph{lacks a closed form} which can be differentiated directly, and \textbf{(2)} for many interesting and useful optimization models, the mapping is \emph{nondifferentiable} in some points, and has zero-valued gradients in others, precluding the straightforward use of gradient descent. As shown in the next subsection, even the class of linear programming problems, widely used in decision modeling, is affected by both issues. Section \ref{sect:review} details the various existing approaches aimed at overcoming these challenges.

\begin{minipage}[t]{0.4\textwidth}
    \begin{algorithm}[H]
\caption{Gradient-descent in prediction-focused learning}
\label{alg:prediction_focused}
\textbf{Input}: training data D$ \equiv \{(\mathbf{z_i},\mathbf{c_i})\}_{i=1}^N$ \\
\textbf{Hyperparams}: $\alpha$- learning rate\\
\begin{algorithmic}[1] 
\STATE Initialize $\bm{\omega}$.
\FOR{each epoch}
\FOR{each instance $(\mathbf{z},\mathbf{c})$}
\STATE $\mathbf{\hat{c}} = m_{\mathbf{\bm{\omega}}} ( \mathbf{z})$ \label{alg:ln:pred} 
\STATE $\mathcal{L} = (\mathbf{\hat{c}} -\mathbf{ c})^2$ \\
\STATE $\bm{\omega} \leftarrow \bm{\omega} - \alpha \frac{d  \mathcal{L}}{d  \mathbf{\hat{c}}} \frac{d  \mathbf{\hat{c}}}{d  \bm{\omega}}$

\ENDFOR
\ENDFOR
\end{algorithmic}
\end{algorithm}
\end{minipage}
\hspace{3mm}
\begin{minipage}[t]{0.55\textwidth}

\begin{algorithm}[H]
\caption{Gradient-descent in decision-focused learning with regret as task loss}
\label{alg:decision_focused}
\textbf{Input}: $\mathcal{F}$, training data D $\equiv \{(\mathbf{z_i}, \mathbf{c_i}, \mathbf{x}^\star(\mathbf{c_i})\}_{i=1}^N$; \\
\textbf{Hyperparams}: $\alpha$- learning rate\\
\begin{algorithmic}[1] 
\STATE Initialize $\bm{\omega}$.
\FOR{each epoch}
\FOR{each instance $(\mathbf{z},\mathbf{c}, \mathbf{x}^\star(\mathbf{c}))$}
\STATE $\mathbf{\hat{c}} = m_{\mathbf{\bm{\omega}}} (\mathbf{ z})$ \\
\STATE $\mathbf{x}^\star(\mathbf{\hat{c}} ) = \argmin_{\mathbf{x} \in \mathcal{F} } f( \mathbf{x}, \mathbf{\hat{c}})$ \\
\STATE $\mathcal{L} = f(\mathbf{x}^\star(\mathbf{\hat{c}}), \mathbf{c}) - f(\mathbf{x}^\star (\mathbf{c}), \mathbf{c})$ \\
\STATE $\bm{\omega} \leftarrow \bm{\omega} - \alpha \frac{d  \mathcal{L}}{d  \mathbf{x}^\star(\mathbf{\hat{c}} )}   \frac{d  \mathbf{x}^\star(\mathbf{\hat{c}} )}{d \mathbf{ \hat{c}}} \frac{d  \mathbf{\hat{c}}}{d  \bm{\omega}}$

\ENDFOR
\ENDFOR
\end{algorithmic}
\end{algorithm}
\end{minipage}
\paragraph{Computational cost.}
\label{subsection:scalability}
Another major challenge in DFL is the computational resources required to train the integrated prediction and optimization model. Note that Line \rev{5} in Algorithm~\ref{alg:decision_focused} evaluates $\mathbf{x}^\star(\mathbf{\hat{c}})$. This requires solving and differentiating the underlying CO problem for each observed data sample, in each epoch. This imposes a significant computational cost even when dealing with small-scale and efficiently solvable CO problems, but can become an impediment in the case of large and (NP-)hard optimization problems. 
\newsavebox{\mybox}
\sbox{\mybox}{%
\begin{tikzpicture}[x=0.35cm,y=0.5cm]
  \readlist\Nnod{3,4,4,1} 
  \foreachitem \N \in \Nnod{ 
    \foreach \i [evaluate={\x=\Ncnt; \y=\N/2-\i+0.5; \prev=int(\Ncnt-1);}] in {1,...,\N}{ 
      \node[mynode] (N\Ncnt-\i) at (\x,\y) {};
      \ifnum\Ncnt>1 
        \foreach \j in {1,...,\Nnod[\prev]}{ 
          \draw[thick] (N\prev-\j) -- (N\Ncnt-\i); 
        }
      \fi 
    }
  }
\end{tikzpicture}
}

\begin{figure}
\centering
\includegraphics[width=0.9\textwidth]{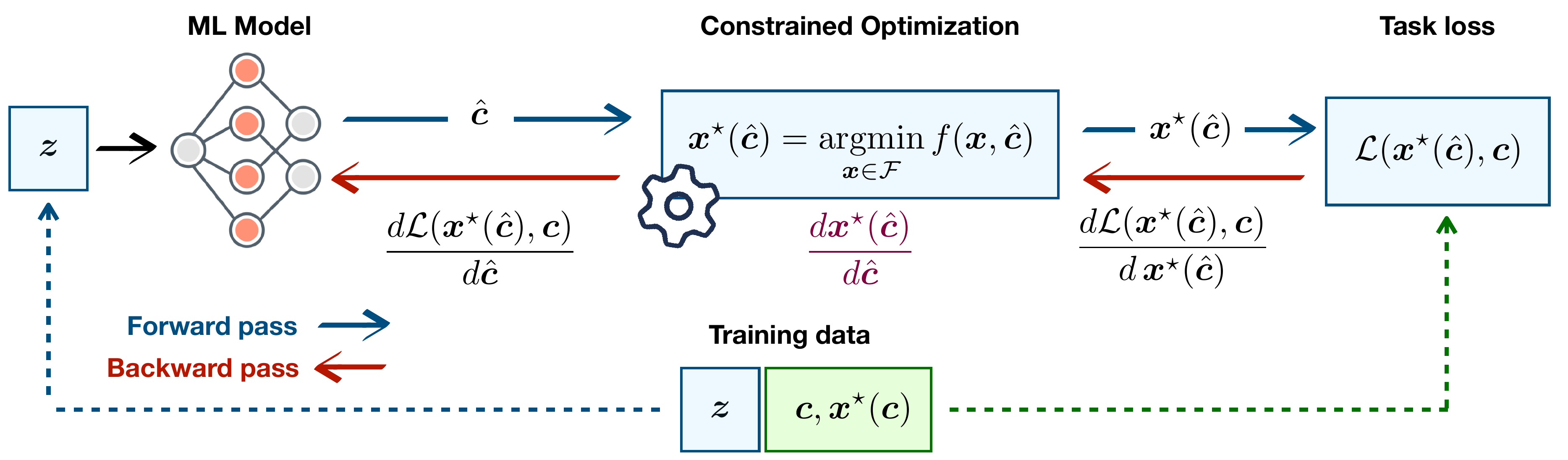}
\caption{In decision-focused learning, the neural network model is trained to minimize the task loss}
\end{figure} 

\subsection{Optimization Problem Forms}
\label{subsec:examples}
The effectiveness of solving an optimization problem depends on the specific forms of the objective and constraint functions. Considerable effort has been made to develop efficient algorithms for certain optimization forms. 
Below, the readers are provided an overview of the key and widely utilized types of optimization problem formulations.
\subsubsection{Convex Optimization}
In \textit{convex} optimization problems, a convex objective function is to be optimized over a convex feasible space. 
This class of problems is distinguished by the guarantee that any locally optimal solution is also globally optimal \cite{boyd2004convex}. Since many optimization problems converge provably to local minima, convex problems are considered to be reliably and efficiently solvable as opposed to \emph{nonconvex} problems. Despite this, convex optimization mappings still impose significant computational overhead on Algorithm~\ref{alg:decision_focused} since 
most convex optimizations are orders of magnitude more complex than conventional neural network layers \cite{amos2017optnet}. Like all parametric optimization problems, convex ones  are implicitly defined mappings from parameters to optimal solutions, lacking a closed form that can be differentiated directly. However as detailed in Section \ref{sect:kkt}, they can be canonicalized to a standard form, which facilitates automation of their solution and backpropagation by a single standardized procedure \cite{agrawal2019differentiable}.

The class of convex problems is broad enough to include some that yield mappings $\mathbf{x}^\star(\mathbf{\hat{c}})$ that are differentiable everywhere, and some that do not. The \emph{linear programs}, which are convex and form nondifferentiable mappings with respect to their objective parameters, are notable examples of the latter case and are discussed next. 

\subsubsection{Linear Programming}
\label{subsubsec:linear_programming}
Linear Programs (LPs) are convex optimization problems whose objective and constraints are composed of affine functions. These programs are predominant as decision models in operations research, and have endless industrial applications since the allocation and transfer of resources is typically modeled by linear relationships between variables \cite{bazaraa2008linear}. The parametric LPs considered in this survey paper take the following form: 
\begin{subequations}
    \label{eq:param_LP}
\begin{align}
   \label{eq:param_LP_obj}
   \mathbf{x}^\star(\mathbf{c}) = \argmin_{\mathbf{x}} \mathbf{c}^\top \mathbf{x}\\
    \label{eq:param_LP_eq}
    \text{s.t.} \;\;A\mathbf{x} = \mathbf{b} \\
    \label{eq:param_LP_ineq}
    \mathbf{x} \geq \mathbf{0}
\end{align}
\end{subequations}
\noindent
Compared to other classes of convex problems, LPs admit efficient solution methods, even for large-scale problems \cite{bazaraa2008linear,ignizio1994linear}. From a DFL standpoint, however, LPs pose a challenge, because the mapping $\mathbf{c} \rightarrow \mathbf{x}^\star(\mathbf{c} )$ is nondifferentiable. Although the  derivatives of mapping     \eqref{eq:param_LP} are defined almost everywhere, they provide no useful information for gradient descent training. To see this, first note the well-known fact that a linear program always takes its optimal value at a vertex of its feasible set \cite{bazaraa2008linear}. Since the number of vertices in any such set is finite,      \eqref{eq:param_LP} maps a continuous parameter space to a discrete set of solutions. As such, it is a piecewise constant mapping. Therefore its derivatives are zero almost everywhere, and undefined elsewhere. Prevalent strategies for incorporating linear programs in DFL thus typically rely on differentiating smooth approximations to the LP, as detailed in Section~\ref{sec:analytical_smoothing}.

Many OR problems, such as the allocation and planning of resources, can be modeled as LPs. Also many prototypical problems in algorithm design (e.g., sorting and top-\textit{k} selection) can be formulated as LPs with continuous variables, despite admitting only discrete integer solutions, by relying on the total unimodularity of the constraint matrices \cite{bazaraa2008linear}.
In what follows, some examples of machine learning models of LPs and how they might occur in a \pto context are given.
\begin{description}
    \item[Shortest paths.] Given a directed graph with a given start and end node, the goal in the shortest path problem is to find a sequence of arcs of minimal total length that connects the start and the end node. The decision variables are binary indicators of each edge's inclusion in the path. The linear constraints ensure $[0,1]$ bounds on each indicator, as well as flow balance through each node. These flow balance constraints capture that, except for the start and end node, each node has as many incoming selected arcs as outgoing selected arcs. For the start node, there is one additional outgoing selected arc, and for the end node, there is one more incoming selected arc. The parameters in the linear objective represent the arc lengths. In many realistic settings---as well as in several common DFL benchmarks \cite{elmachtoub2022smart,PogancicPMMR20}---these are unknown, requiring them to be predicted before a shortest path can be computed. This motivating example captures the realistic setting in which the shortest route between two locations has to be computed, but in which the road traversal times are uncertain (due to unknown traffic conditions, for example), but can be predicted from known features (such as day of the week, time of day and weather conditions).
    
  \item[Bipartite matching.] 
Given a bipartite graph with weighted arcs, where the weights are unknown and must be predicted, the task is to choose a subset of arcs so that each node is involved in at most one selected arc, maximizing the total weight.
The variables lie in $[0,1]$ indicating the inclusion of each edge and the weights are the objective parameters. The constraints ensure that each node is involved at most once in a selected arc. With a complete bipartite graph, matchings can be construed as permutations, which can be employed in tasks such as learning to rank \cite{kotary2022end}.
  
  \item[Sorting and Ranking.] The sorting of any list of predicted values can be posed as a linear program over a feasible region whose vertices correspond to all of the possible permutations of the list. The related ranking, or argsort problem assigns to any length-$n$ list a permutation of sequential integers $[n]$ which sorts the list. By smoothing the linear program, these basic operations can be differentiated and backpropagated \cite{blondel2020fast}.

  \item[Top-$k$ selection.] 
    Given a set of items and item values that must be predicted, the task is to choose the subset of size $k$ with the largest total value in selected items. In addition to $[0,1]$ bounds on the indicator variables, a single linear constraint ensures that the selected item indicators sum to $k$. A prevalent example can be found in multilabel classification \cite{amos2019limited,martins2016softmax}.

  \item[Computing the maximum.] 
   This is a special case of top-$k$ selection where $k=1$. When the LP's objective is regularized with the entropy term $H(\mathbf{x} ) = \mathbf{x}^\top \log \mathbf{x}$, the mapping from predicted values to optimal solutions is equivalent to a softmax function \cite{agrawal2019differentiable}. 
  
  \item[Max-flow/ min-cut.] Given a network with predefined source and sink nodes, and predicted flow capacities on each arc, the task is to find the maximum flow rate that can be channeled from source to sink. Here the predicted flow capacities occupy the right-hand side of the linear constraints, which is not in line with the DFL problem formulation introduced in subsection~\ref{subsec:problem_description}. However, in the min-cut problem---which is the dual linear program of the max-flow problem---the flow capacities are the parameters in the objective function. The max-flow problem can thus be cast as an equivalent min-cut problem allowing DFL techniques to predict the flow capacities. 
\end{description}
\subsubsection{Integer Linear Programming}
Integer Linear Programs (ILPs) are another mainstay in OR and AI research. ILPs differ from LPs in that the decision variables $\mathbf{x}$ are restricted to integer values, i.e., $\mathbf{x} \in \mathbb{Z}^k$ where $\mathbb{Z}^k$ is the set of integral vectors of appropriate dimensions. Like LPs, ILPs are challenging to use in DFL because they yield discontinuous, nondifferentiable mappings. Computationally however, they are more challenging due to their NP-hard complexity, which may preclude the exact computation of the mapping $\mathbf{\hat{c}} \rightarrow \mathbf{x}^\star(\mathbf{\hat{c}})$ at each step of Algorithm \ref{alg:decision_focused}. Their differentiation is also significantly more challenging, since the discontinuity of their feasible regions prevents many smoothing techniques that can be applied in DFL with LPs.

The following examples include ILPs in \pto problems.
\begin{description}
  \item[Knapsack.] The knapsack problem \eqref{eq:unitknapsackformulation}, discussed earlier, has been used to compose benchmark problems in several papers about DFL \cite{mandi2020smart,mandi2020interior,demirovic2019investigation}. Given are weights of a set of items, as well as a capacity. The items also have associated values, which have to be predicted from features. The optimization task involves selecting a subset of the items that maximizes the value of the selected items, whilst ensuring that the sum of the associated weights does not exceed the capacity.
  \item[Travelling salesperson problem.]
  In the travelling salesperson problem, the list of cities, and the distances between each pair of cities, is given. The goal is to find a path of minimal length that visits each city exactly once. In the \pto setting, the distances between the cities first have to be predicted \cite{PogancicPMMR20} from observable empirical data.
  \item[Combinatorial portfolio optimization.]
  Portfolio optimization involves making optimal investment decisions across a range of financial assets. In the combinatorial \pto variant, the decisions are discrete, and must be made on the basis of the predicted next period's increase in the value of several assets \cite{aaaiFerberWDT20}.
  \item[Diverse bipartite matching.]
  Diverse bipartite matching problems are similar to the bipartite matching problems described in \ref{subsubsec:linear_programming}, but are subject to additional diversity constraints \cite{aaaiFerberWDT20,mulamba2020discrete,mandi22-pmlr}. In this variant, edges have additional properties. The diversity constraints enforce lower and upper bounds on the proportion of edges selected with a certain property. These changes preclude an LP formulation, by breaking the total unimodularity property possessed by the standard bipartite matching LP.
  \item[Energy-cost aware scheduling.]
  Energy-cost aware scheduling involves scheduling a set of tasks across a set of machines minimizing the overall energy cost involved. As future energy costs are unknown, they first have to be predicted \cite{mandi2020smart}.
\end{description}
\subsubsection{Integer Nonlinear Programming}
In integer nonlinear programming, variables are defined over an integer domain whiles the objective function and/or the constraints are nonlinear. Performing DFL on integer nonlinear programs faces similar challenges as in performing DFL on ILPs: their implicitly defined mappings $ \mathbf{x}^\star (\mathbf{\hat{c}})$ result in zero-valued gradients almost everywhere. Additionally, because of their nonlinear nature, many of the techniques developed for DFL with ILPs, which assume linearity, do not directly translate to integer nonlinear programs \cite{elmachtoub2022smart,PogancicPMMR20}. To the best of our knowledge, no DFL technique has specifically been developed for or tested on integer nonlinear programs. The most closely related work is \cite{ferber2022surco}, which employs approximate ILP surrogates for integer nonlinear programs, which are then used to compose a DFL training procedure using known techniques for differentiation through the ILP surrogate model.
\tikzset{
  table/.style={
    matrix of nodes,
    row sep=0.1\pgflinewidth,
    column sep=-\pgflinewidth,
    nodes={
      rectangle,
      draw=black,
      align=center,
      minimum height=\baselineskip, 
       text width = 0.12\textwidth, text depth = 16mm, text
    height = 3mm, draw, text centered, anchor=south, font=\footnotesize 
    },
    nodes in empty cells,
   column 1/.style={
            nodes={draw=black, text width = 0.2\textwidth, font=\bfseries \footnotesize}
        },
   column 2/.style={
   	nodes={draw=black, text width = 0.15\textwidth}
   },
      column 3/.style={
   	nodes={draw=black, text width = 0.16\textwidth}
   },
  row 1/.style={
           nodes={
                fill=black,
                text=white,
                text depth = 5mm,
                text height = 2mm,
                font=\bfseries
            },
            }
},
arrow/.style={thick,->,>=stealth}
}
\begin{center}
\begin{figure}
\centering
\begin{tikzpicture}
\matrix (MT) [table]
{
&    Objective Function & Constraint Functions  & Decision Variables & CO Solver\\
\textcolor{blue}{Analytical Differentiation of Optimization Mappings} & \textcolor{blue}{Strictly Convex} & \textcolor{blue}{Convex} & \textcolor{blue}{Continuous} & \textcolor{blue}{Primal-Dual Solver}\\
\textcolor{blue}{Analytical Smoothing of Optimization Mappings} & \textcolor{blue}{Linear} & \textcolor{blue}{Linear} & \textcolor{blue}{Continuous/ Discrete} & \textcolor{blue}{Primal-Dual Solver}\\
\textcolor{violet}{Smoothing by Random Perturbations}  & \textcolor{violet}{Linear in the Predicted Parameter} & \textcolor{violet}{Not limited to specific form} & \textcolor{violet}{Continuous/ Discrete} & \textcolor{violet}{Solver agnostic}\\
\textcolor{purple}{Differentiation of Surrogate Loss Functions}  & \textcolor{purple}{Linear in the Predicted Parameter} &  \textcolor{purple}{ Not limited to specific form }  & \textcolor{purple}{Continuous/ Discrete} & \textcolor{purple}{Solver agnostic} \\
};
\draw [arrow,rounded corners,thick]  (MT-3-1.west) -- ++(-1cm,0) |- ([xshift=-1mm]MT-2-1.west)
node[midway, sloped,above]{\textcolor{blue}{\textbf{Utilize}}};
\end{tikzpicture}
\caption{An overview of gradient-based DFL methodologies categorized into four classes.}
\label{Fig:DFLMethodologyTable}
\end{figure} 
\end{center}
\raggedbottom
\section{Review of Decision-Focused Learning Methodologies}\label{sect:review}
Gradient-based learning is a popular learning approach in ML, and likewise, gradient-based DFL techniques have been extensively studied, with the added benefit that (deep) neural networks can be used as the underlying ML model. However, other ML frameworks, such as tree-based methods or search-based methods, do not require gradients at all and can side-step the issue of zero-valued gradients altogether.
Hence, DFL methodologies can be classified into two broad categories: 
\textbf{I.} Gradient-based DFL, and \textbf{II.} Gradient-free DFL.
First, gradient-based DFL will be surveyed, followed by gradient-free DFL.
\subsection{Review of Gradient-Based DFL Methodologies}
This subsection will describe several DFL techniques which address the challenge of differentiating an optimization mapping for DFL in gradient-based training.
In essence, different approaches propose different smoothed surrogate approximations of $\frac{ d  \mathbf{x}^\star (\mathbf{\hat{c}})}{ d  \mathbf{\hat{c}}}$ or $\frac{ d  \mathcal{L}(\mathbf{x}^\star (\mathbf{\hat{c}}))}{ d \mathbf{ \hat{c}}}$, which is used for backpropagation. 
\raggedbottom
This paper proposes the first categorization of existing gradient-based DFL techniques into the following four distinct classes:
\begin{description}
    \item [Analytical Differentiation of Optimization Mappings:] Techniques in this category aim to compute exact derivatives for backpropagation by differentiating the optimality conditions for certain optimization problem forms, for which the derivative exists and is non-zero.
    \item [Analytical Smoothing of Optimization Mappings:] These approaches deal with combinatorial optimization problems (for which the analytical derivatives are zero almost everywhere) by performing smoothing of combinatorial optimization problems, which results in approximate problems that can be differentiated analytically.
    \item [Smoothing by Random Perturbations:] Techniques under this category utilize implicit regularization through perturbations, constructing smooth approximations of optimization mappings.
    \item [Differentiation of Surrogate Loss Functions:] Techniques under this category propose surrogate loss functions of specific task loss such as regret. These surrogate losses reflect the quality of the decisions and they provide easy-to-compute gradients or subgradients for gradient-based training.
\end{description}
\raggedbottom
Figure~\ref{Fig:DFLMethodologyTable} presents key characteristics of these four methodology classes, 
highlighting the types of problems that can be addressed within each class. 
Next, each category is thoroughly described.

%






%

\subsubsection{Analytical Differentiation of Optimization Mappings}
\label{sect:kkt}
As discussed before, differentiating through parametric CO problems comes with two main challenges. First, since CO problems are complex, implicitly defined mappings from parameters to solutions; computing the derivatives is not straightforward. Second, since some CO problems result in piecewise-constant mappings, their derivatives are zero almost everywhere, and do not exist elsewhere.
\raggedbottom

This subsection pertains to CO problems for which the second challenge does not apply, i.e., problems that are smooth mappings. For these problems, all that is required to implement DFL is direct differentiation of the mapping in Eq.~\eqref{eq:opt_generic}.
\paragraph{Differentiating unconstrained relaxations.}
An early work discussing differentiation through constrained argmin problems in the context of machine learning is \cite{gould2016differentiating}. It first proposes a technique to differentiate the argmin of a smooth, \textit{unconstrained} convex function. When $V(\mathbf{c}) = \argmin_{\mathbf{x}} f(\mathbf{c}, \mathbf{x})$, it can be shown that when all second derivatives of $f$ exist, 
\begin{equation}
 \frac{dV(\mathbf{c})}{d \mathbf{c}} = - \frac{f_{\mathbf{cx}}( \mathbf{c},V(\mathbf{c}))}{f_{\mathbf{xx}}( \mathbf{c},V(\mathbf{c}))}
\end{equation}
where $f_{ \mathbf{cx}}$ is the second partial derivative of $f$ with respect to $\mathbf{c}$ followed by $\mathbf{x}$. This follows from implicit differentiation of the first-order optimality conditions
\begin{equation}
 \frac{d}{d \mathbf{x}}f(\mathbf{c},V(\mathbf{c})) = 0
\end{equation}
with respect to $\mathbf{c}$, and rearranging terms. Here the variables $\mathbf{c}$ are the optimization problem's defining parameters, and the variables $\mathbf{x}$ are the decision variables.

This technique is then extended to find approximate derivatives to \textit{constrained} optimization problems with inequality constraints $g_i(\mathbf{c}, \mathbf{x}) \leq 0$, 
by first relaxing the problem to an unconstrained problem, by means of the log-barrier function
\begin{equation}
 F(\mathbf{c},\mathbf{x}) =  f(\mathbf{c},\mathbf{x}) - \mu \sum_i^{M_{in}} \log(-g_i(\mathbf{c},\mathbf{x}))
\end{equation}
\noindent where $M_{in}$ is the number of inequality constraints. 
Then $\argmin_{ \mathbf{x} } F(\mathbf{c},\mathbf{x})$ is differentiated with respect to $\mathbf{c}$ for some choice of the scaling factor $\mu$. Since this approach relies on approximations and requires hyperparameter tuning for the factor $\mu$, subsequent works focus on differentiating CO problems directly via their own global conditions for optimality, as discussed next. 

\paragraph{Differentiating KKT conditions of quadratic programs}
More recent approaches are based on differentiating the optimality conditions of a CO problem directly, i.e., without first converting it to an unconstrained problem. Consider an optimization problem and its optimal solution:
\begin{subequations}
    \label{eq:general_convex}
\begin{align}
  \label{eq:general_convex_objective}
    \mathbf{x}^\star   = \argmin_{\mathbf{x}} &\;\;
    f(\mathbf{x})\\
    \label{eq:general_convex_inequality}
    \texttt{s.t.} &\;\;
    \bm{g}(\mathbf{x}) \leq 0 \\
    \label{eq:general_convex_equality}
     & \;\; \bm{h}(\mathbf{x}) = 0
\end{align}
\end{subequations}
and assume that $f$, $g$ and $h$ are differentiable functions of $\mathbf{x}$. The Karush–Kuhn–Tucker (KKT) conditions are a set of equations expressing optimality conditions for a solution $\mathbf{x}^\star$ of problem (\ref{eq:general_convex}) \cite{boyd2004convex}:
\begin{subequations}
    \label{eq:KKT}
\begin{align}
  \label{eq:KKT_stationarity}
    \nabla f(\mathbf{x}^\star) + \sum_i w_i \nabla h_i(\mathbf{x}^\star) + \sum_j  u_j \nabla g_j(\mathbf{x}^\star) = 0\\
    \label{eq:KKT_primal_feasibility_ineq}
    g_j(\mathbf{x}^\star) \leq 0 \;\; \forall j \\
    \label{eq:KKT_primal_feasibility_eq}
    h_i(\mathbf{x}^\star) = 0 \;\; \forall i \\
    \label{eq:KKT_dual_feasibility}
    u_j \geq 0 \;\; \forall j \\
    \label{eq:KKT_complementary_slackness}
     u_j g_j(\mathbf{x}^\star) = 0  \;\; \forall j
\end{align}
\end{subequations}
\textsl{OptNet} is a framework developed by \citeA{amos2017optnet} to differentiate through optimization mappings that are convex quadratic programs (QPs) by differentiating through these KKT conditions. In convex quadratic programs, the objective $f$ is a convex quadratic function and the constraint functions $g, h$ are linear over a continuous domain. In the most general case, each of $f$, $g$ and $h$ are dependent on a distinct set of parameters, in addition to the optimization variable $\mathbf{x}$:
\begin{subequations}
    \label{eq:QP_param}
\begin{align}
  \label{eq:QP_param_f}
    f(\mathbf{c},Q, \mathbf{x}) = \frac{1}{2} \mathbf{x}^\top Q \mathbf{x} + \mathbf{c}^\top \mathbf{x}\\
    \label{eq:QP_param_g}
      h(A, \mathbf{b}, \mathbf{x}) = A \mathbf{x} - \mathbf{b}\\
    \label{eq:QP_param_h}
     g(R,\mathbf{s},\mathbf{x}) = R \mathbf{x} - \mathbf{s }
\end{align}
\end{subequations}
When $\mathbf{x} \in \mathbb{R}^n$ and the number of inequality and equality constraints are $M_{in}$ and $M_{eq}$, respectively, a QP problem is specified by parameters $Q \in \mathbb{R}^{n \times n}$,  $\mathbf{c} \in \mathbb{R}^{n}$,   $R \in \mathbb{R}^{n \times M_{in}}$, $\mathbf{s} \in \mathbb{R}^{M_{in}}$, $A \in \mathbb{R}^{n \times M_{eq}}$, and $\mathbf{b} \in \mathbb{R}^{M_{eq}}$. 
The optimal solution $\mathbf{x}^{\star}$ can be implicitly differentiated with respect to each of the parameters $(Q, \mathbf{c},R,\mathbf{s},A, \mathbf{b}) $, by directly differentiating the KKT conditions of optimality \eqref{eq:KKT}. This produces a linear system of equations, which can be solved for the desired gradients $\frac{d \mathbf{x}^{\star}}{d \mathbf{c}}$.
Later, \citeA{TakuyaGradientBoosting} extended the technique of \citeA{amos2017optnet}, by also computing the QP problem's second order derivatives. This enables training with gradient boosting models, which require the gradient as well as the Hessian matrix of the loss.


\paragraph{Differentiating optimality conditions of conic programs.}
Another class of problems with a parametric canonical form are the conic programs, which take the form: 
\begin{subequations}
    \label{eq:cone}
\begin{align}
  \label{eq:cone_objective}
    \mathbf{x}^\star(A,\mathbf{b},\mathbf{c})   = \argmin_{ \mathbf{x} } &\;\;
    \mathbf{c}^\top \mathbf{x}\\
    \label{eq:cone_inequality}
    \texttt{s.t.} &\;\;
    A\mathbf{x} - \mathbf{b} \in \mathcal{K}
\end{align}
\end{subequations}
where $\mathcal{K}$ is a nonempty, closed, convex cone. 

A framework for differentiating the mapping (\ref{eq:cone}) for any $\mathcal{K}$ is proposed by \citeA{agrawal2019coneprogram}, which starts by forming the homogeneous self-dual embedding of (\ref{eq:cone}), whose parameters form an askew-symmetric block matrix composed of $A$, $\mathbf{b}$, and $\mathbf{c}$. Following the technique proposed by  \citeA{busseti2019solution}, the solution to this embedding is expressed as the problem of finding a zero of a mapping containing a skew-symmetric linear function and projections onto the cone $\mathcal{K}$ and its dual. The zero-value of this function is implicitly differentiated, in a similar manner to the KKT conditions of a quadratic program \cite{amos2017optnet}. 
The overall mapping  (\ref{eq:cone}) is viewed as the composition of a function that maps $(A,\mathbf{b},\mathbf{c})$ onto the skew-symmetric parameter space of the self-dual embedding, the rootfinding problem that produces a solution to the embedding, and a transformation back to a solution of the primal and dual problems. The overall derivative is found by a chain rule applied over this composition. 

Subsequent work by \citeA{agrawal2019differentiable} leverages the above-described differentiation of cone programs to develop a more general differentiable convex optimization solver---\textsl{cvxpylayers}. It is well known that conic programs of the form (\ref{eq:cone}) can provide canonical representations of convex programs \cite{nemirovski2007advances}. The approach described by \citeA{agrawal2019differentiable} is based on this principle; that a large class of \emph{parametric} convex optimization problems can be recast as equivalent parametric cone programs, with an appropriate choice of the cone $\mathcal{K}$. A major benefit of this representation is that it allows a convex program to be separated with respect to its defining parameters ($A,\mathbf{b},\mathbf{c}$) and its structure $\mathcal{K}$, allowing a generic procedure to be applied for solving and differentiating the transformed problem with respect to $A$, $\mathbf{b}$ and $\mathbf{c}$. 

To transform convex programs to cone programs \eqref{eq:cone}, the framework of \citeA{grant2008graph} is utilized, which is based on two related concepts. 
First is the notion of \emph{disciplined convex programming}, which assists the automation of cone transforms by imposing a set of rules or conventions on how convex programs can be represented. Second is the notion of \emph{graph implementations}, which represent functions as optimization problems over their epigraphs, for the purpose of generically representing optimization problems and assisting conversion between equivalent forms. The associated software system called $\texttt{cvx}$ allows for disciplined convex programs to be converted to cone programs via their graph implementations. Subsequently, the transformed problem is solved using conic optimization algorithms, and its optimal solution is converted to a solution of the original disciplined convex program. Differentiation is performed through each operation and combined by the chain rule. The transformation of parameters between respective problem forms, and the solution recovery step, are differentiable by virtue of being affine mappings \cite{agrawal2019differentiable}. The intermediate conic program is differentiated via the methods of \citeA{agrawal2019coneprogram}. 

\paragraph{Solver unrolling and fixed-point differentiation.}
While the methods described above for differentiation through CO problems are generic and applicable to broad classes of problems, other practical techniques have been proven effective and even advantageous in some cases. A common strategy is that of solver \emph{unrolling}, in which the solution to \eqref{eq:opt_generic} is found by executing an iterative optimization method on the computational graph of its preceding predictive model. The optimization mapping \eqref{eq:opt_generic} is then backpropagated simply by automatic differentiation through each step of the algorithm, thus avoiding the need to explicitly model $\frac{d  \mathbf{x}^{\star}(\mathbf{c} )}{d  \mathbf{c}}$ \cite{domke2012generic}. While this approach leads to accurate backpropagation in many cases, it suffers disadvantages in efficiency due to the memory and computational resources required to store and apply backpropagation over the entire computational graph of an algorithm that requires many iterations \cite{amos2017optnet}. A comprehensive survey of algorithm unrolling in image processing applications is provided by \citeA{monga2021algorithm}.

Another way in which a specific solution algorithm may provide gradients though a corresponding optimization mapping, is by implicit differentiation of its fixed-point conditions. Suppose that the solver iterations 
\begin{equation}
    \label{eq:opt_iteration}
        \mathbf{x}_{t+1}(\mathbf{c}) = \mathcal{U}(\mathbf{x}_t (\mathbf{c}) ,\;  \mathbf{c} )
\end{equation}

\noindent converge as $t \rightarrow \infty$ to a solution $\mathbf{x}^{\star}(\mathbf{c})$ of the problem \eqref{eq:opt_generic}, then the fixed-point conditions 

\begin{equation}
    \label{eq:opt_fixedpt}
       \mathbf{x}^{\star}(\mathbf{c}) = \mathcal{U}(\mathbf{x}^{\star} (\mathbf{c}) ,\;  \mathbf{c} )
\end{equation}
\noindent are satisfied. Assuming the existence of all derivatives on an open set containing $\mathbf{c}$ to satisfy the implicit function theorem, it follows  by implicit differentiation with respect to $\mathbf{c}$ that
\begin{equation}
    \label{eq:Lemma_fixedpt}  
        (\mathbf{I} - \Phi ) \frac{d \mathbf{x}^{\star}}{d \mathbf{c}} = \Psi,
\end{equation}

\noindent which is a linear system to be solved for $\frac{d \mathbf{x}^{\star}}{d \mathbf{c}}$, in terms of   $\Phi = {{\frac{d \mathcal{U} }{d \mathbf{x}^{\star}} (\mathbf{x}^{\star}(\mathbf{c}), \; c )}}$,  $\Psi = \frac{d \mathcal{U} }{d \mathbf{c}} (\mathbf{x}^{\star}(\mathbf{c}), \; \mathbf{c})$ and identity matrix $\mathbf{I}$.

\noindent The relationship between unrolling and differentiation of the fixed-point conditions is studied by \citeA{kotary2023folded}, showing that backpropagation of \eqref{eq:opt_generic} by unrolling \eqref{eq:opt_iteration} is equivalent to solving the linear system \eqref{eq:Lemma_fixedpt}  by fixed-point iteration. The convergence rate of the backward pass in unrolling is determined by that of solving the linear system \eqref{eq:Lemma_fixedpt} in such a manner, and can be calculated in terms of the spectral radius of $\Phi$. 


\paragraph{Discussion.}
In contrast to most other differentiable optimization methods surveyed in this survey paper, the analytical approaches in this subsection allow for the backpropagation of coefficients that specify the constraints as well as the objective function.
For example, \citeA{amos2017optnet} propose parametric quadratic programming layers whose linear objective parameters are predicted by previous layers, and whose constraints are learned through the layer's own embedded parameters. This is distinct from most cases of DFL, in which the CO problems have fixed constraints and no trainable parameters of their own. 

Furthermore, the techniques surveyed in this subsection are aimed at computing the exact gradients of parametric optimization mappings. However, many applications of DFL contain optimization mappings that are discontinuous and piecewise-constant. 
Such mappings, including parametric  linear programs \eqref{eq:param_LP}, have gradients that are zero almost everywhere and thus do not supply useful descent directions for SGD training. Therefore, the techniques of this subsection are often applied after regularizing the problem analytically with smooth functions, as detailed in the next subsection. 

\subsubsection{Analytical Smoothing of Optimization Mappings}
\label{sec:analytical_smoothing}
To differentiate through combinatorial optimization problems, the optimization mapping first has to be smoothed. While techniques such as noise-based gradient estimation (surveyed in Section~\ref{sect:perturb}) provide smoothing and differentiation simultaneously, analytical differentiation first incorporates smooth analytical terms in the optimization problem's formulation, and then analytically differentiates the resulting optimization problem using the techniques discussed in Section~\ref{sect:kkt}.

\paragraph{Analytical smoothing of linear programs.} Note that while an LP problem is convex and has continuous variables, only a finite number of its feasible solutions can potentially be optimal. These points coincide with the vertices of its feasible polytope \cite{bazaraa2008linear}. Therefore the mapping $\mathbf{x}^\star ( \mathbf{ \hat{c}} )$ in (\ref{eq:param_LP}), as a function of $\mathbf{ \hat{c} }$, is discontinuous and piecewise constant, and thus requires smoothing before it can be differentiated through. An approach to do so was presented in
\citeA{aaai/WilderDT19}, which proposes to  augment the linear LP objective function with the Euclidean norm of its decision variables, so that the new objective takes the following form
\begin{subequations}
    \label{eq:qptl}
    \begin{align}
    \mathbf{x}^{\star}(\mathbf{c}) &= \argmin_\mathbf{x}  \mathbf{c}^\top \mathbf{x} + \mu \lVert \mathbf{x} \rVert^2_2 \\
    &= \argmin_{\mathbf{x}} \lVert \mathbf{x} - \Big(\frac{ -\mathbf{c} }{\mu} \Big) \rVert^2_2
    \end{align}
\end{subequations}

\noindent where the above equality follows from expanding the square and cancelling constant terms, which do not affect the $\argmax$. This provides an intuition as to the effect of such a quadratic regularization: it converts a LP problem into that of projecting the point $\Big(\frac{ -\mathbf{c} }{\mu} \Big)$ onto the feasible polytope, which results in a continuous mapping $\mathbf{c} \to \mathbf{x}^\star(\mathbf{c})$.  \citeA{aaai/WilderDT19} then train models in decision-focused paradigm by solving and backpropagating the respective quadratic programming problem using the OptNet framework \cite{amos2017optnet}, in order to learn to predict objective parameters with minimal regret. At test time, the quadratic smoothing term is removed. \citeA{aaai/WilderDT19} refers to such regret-based DFL with quadratically regularized linear programs as the \emph{Quadratic Programming Task Loss} method (QPTL).


Other forms of analytical smoothing for linear programs  can be applied by adding different regularization functions to the objective function. Other regularization terms for LPs include the entropy function $H(\mathbf{x}) = \sum_i x_i \log x_i$ and the binary entropy function $H_b(\mathbf{x}) = H(\mathbf{x}) + H(\mathbf{1}-\mathbf{x})$. To differentiate the resulting smoothed optimization problems, the framework of \citeA{agrawal2019differentiable} can be used. Alternatively, problem-specific approaches, without employing this framework, have also been proposed. For example, \citeA{blondel2020fast} propose a method for problems where $H$ smooths an LP for differentiable sorting and ranking, and \citeA{amos2019limited} propose a way to differentiate using $H_b$ for multilabel classification problems. Both works propose fast implementations for both the forward and backward passes of the respective optimization problems. 

In a related approach, \citeA{mandi2020interior} propose a general, differentiable LP solver based on log-barrier regularization. For a parametrized LP of standard form \eqref{eq:param_LP}, gradients are computed for the regularized form in which the constraints $\mathbf{x} \geq \mathbf{0}$ are replaced with log-barrier approximations as follows:
\begin{subequations}
    \label{eq:intopt_LP}
\begin{align}
   \label{eq:intopt_LP_obj}
   \mathbf{x}^\star (\mathbf{c}) = \argmin_{\mathbf{x}} \mathbf{c}^\top \mathbf{x} - \mu \sum_{i=1}^{\dim (\mathbf{x})} \log(x_i) \\
    \label{eq:intopt_LP_eq}
    \text{s.t.} \;\;A \mathbf{x} = \mathbf{b}
\end{align}
\end{subequations}
\noindent While this method, in this sense,
is similar to the approach of \citeA{gould2016differentiating}, it exploits several efficiencies specific to linear programming, in which the log-barrier term serves a dual purpose of rendering \eqref{eq:intopt_LP} differentiable and also aiding its solution~\cite{mandi2020interior}. Rather than forming and solving this regularized LP problem directly, the solver uses an interior point method to produce a sequence of log-barrier approximations to the LP's homogenous self-dual (HSD) embedding. Early stopping is applied in the interior point method, producing a solution to \eqref{eq:intopt_LP} for some $\mu$, which serves as a smooth surrogate problem for differentiation. A major advantage of this technique is that it only requires optimization of a linear program, making it in general more efficient than direct solution of a regularized problem as in the approaches described above. 

\paragraph{Analytical smoothing of integer linear programs.}
To differentiate through ILPs, \citeA{aaai/WilderDT19} propose to simply drop the integrality constraints, and to then smooth and differentiate through the resulting LP relaxation, which is observed to give satisfactory performance in some cases. \citeA{aaaiFerberWDT20} later extended this work by using a more systematic approach to generate the LP relaxation of the ILP problem. They use the method of cutting planes to discover an LP problem that admits the same solution as the ILP. Subsequently, the method of \citeA{aaai/WilderDT19} is applied to approximate the LP mapping's derivatives. Although this results in enhanced performance with respect to regret, there are some practical scalability concerns, since the cut generation process is time consuming but also must be repeated for each instance in each training epoch.

Although this subsection primarily surveys the smoothing of LPs and ILPs, as these are the focal points of this paper, note that differentiable relaxation of other optimization problems has also received attention recently. For example, differential relaxations of MAXSAT~\cite{wang2019satnet,linsatnet} and submodular optimization problems \cite{NIPS2017Djolonga,ijcaiTschiatschekS018} have been developed to embed these problems into neural networks.
\subsubsection{Smoothing by Random Perturbations}
\label{sect:perturb}
A central challenge in DFL is the need for smoothing operations of non-smooth optimization mappings. 
In contrast to the DFL techniques surveyed in the previous subsection, which perform the smoothing operation by adding explicit regularization functions to the objective functions of optimization problems, this subsection focuses on techniques that use implicit regularization through perturbations.
These techniques construct smooth approximations of the optimization mappings by adopting a probabilistic point of view. To introduce this point of view, the CO problem in this section is not viewed as a mapping from $\mathbf{c}$ to $\mathbf{x}^\star (\mathbf{c})$. Rather, it is viewed as a function that maps $\mathbf{c}$ onto a probability distribution over the feasible region $\mathcal{F}$.
From this perspective, $\mathbf{x}^\star (\mathbf{c})$ can be viewed as a random variable, conditionally dependent on $\mathbf{c}$.
The motivation behind representing $\mathbf{x}^\star (\mathbf{c})$ as a random variable is that the rich literature of likelihood maximization with latent variables, in fields such as Probabilistic Graphical Models (PGMs) \cite{koller2009probabilistic},
can be exploited. 

\paragraph{Implicit differentiation by perturbation.}
One seminal work in the field of PGMs is by \citeA{Domke10}.
This work contains an important proposition, which 
deals with a setup where a variable $\bm{\theta}_1$ is conditionally dependent on another variable $\bm{\theta}_2$ and the final loss $\mathcal{L}$ is defined on the variable $\bm{\theta}_1$. Let $p(\bm{\theta_1}| \bm{\theta_2})$  and $\mathbb{E}[\bm{\theta_1}| \bm{\theta_2}]$ be the conditional distribution and the conditional mean of $\bm{\theta}_1$.
The loss $\mathcal{L}$ is measured on the conditional mean $\mathbb{E}[\bm{\theta_1}| \bm{\theta_2}]$ and the goal is to compute the derivative of $\mathcal{L}$ with respect to $\bm{\theta}_2$.
\citeA{Domke10} proposes that the derivative of $\mathcal{L}$ with respect to $\bm{\theta}_2$ can be approximated by the following finite difference method:
\begin{equation}
     \frac{d L}{d \bm{\theta_2} }\approx \frac{1}{\delta} \Bigg( \mathbb{E}[\bm{\theta_1}| \big( \bm{\theta_2} + \delta \frac{d}{d \bm{\theta_1}} \big( \mathcal{L}(\mathbb{E}[\bm{\theta_1}| \bm{\theta_2}] \big)   \big) ]- \mathbb{E}[\bm{\theta_1}| \bm{\theta_2}]   \Bigg)
     \label{eq:domke}
\end{equation}
where $ \frac{d}{d \bm{\theta_1}}[\mathcal{L}(\mathbb{E}[ \bm{\theta_1}]  ) ]$ is the derivative $\mathcal{L}$ with respect to $\bm{\theta}_1$ at $\mathbb{E}[ \bm{\theta_1}]$. 
Notice that the first term in \eqref{eq:domke} is the conditional mean after perturbing the parameter $\bm{\theta}_2$ where the magnitude of the perturbation is modulated by the derivative of $\mathcal{L}$ with respect to $\bm{\theta}_1$.
Taking inspiration from this proposition,
by defining a conditional distribution $p(\mathbf{x}^\star ( \mathbf{ \hat{c}} )|\hat{c})$, one can
compute the derivative of the regret with respect to $\mathbf{ \hat{c} }$ in the context of DFL.

To perfectly represent the deterministic mapping $\mathbf{c} \rightarrow \mathbf{x}^\star (\mathbf{c})$, the straightforward choice is to
define a Dirac mass distribution, 
which assigns all probability mass to the optimal point and none to other points, i.e.,
\begin{equation}
     p( \mathbf{x}| \mathbf{c}) = \begin{cases}
          1 & \mathbf{x}= \mathbf{x}^\star (\mathbf{c}) \\
          0 & \text{otherwise}
     \end{cases}
     \label{eq:argmax_distri}
\end{equation}

\paragraph{Differentiation of blackbox combinatorial solvers (DBB).}
Note that with the distribution in \eqref{eq:argmax_distri} $\mathbb{E}_{\mathbf{x} \sim p( \mathbf{x} |\mathbf{c})}[x|c] = \mathbf{x}^\star (\mathbf{c})$.
Hence, using conditional probability in the proposition in \eqref{eq:domke}, $\frac{d \mathcal{L}(\mathbf{x}^\star ( \mathbf{ \hat{c}} ) )} { d \mathbf{\hat{c}} }$ can be approximated in the following way:
\begin{align}
     \frac{d \mathcal{L}(\mathbf{x}^\star ( \mathbf{ \mathbf{\hat{c}}} ) )} { d \mathbf{ \hat{c} }} \approx \nabla^{(DBB)}\mathcal{L}(\mathbf{x}^\star ( \mathbf{ \hat{c}} ) ) =  \Bigg(  \mathbf{x}^\star \Big( \mathbf{ \hat{c}} + \delta \frac{d \mathcal{L}(\mathbf{x}^\star ( \mathbf{ \hat{c}} ) )}{d \mathbf{x}^\star ( \mathbf{ \hat{c}} )} \Big)-  \mathbf{x}^\star \Big(\mathbf{ \hat{c}} \Big) \Bigg)
     \label{eq:dbb}
\end{align}
The gradient computation technique proposed by \citeA{PogancicPMMR20} takes the form of \eqref{eq:dbb}. 
They interpret it as substituting the jump-discontinuous optimization mapping with a piece-wise linear interpolation. It is a linear interpolation of the mapping $\mathbf{\hat{c}} \rightarrow \mathbf{x}^\star (\mathbf{\hat{c}})$ between the points $\mathbf{ \hat{c} }$ and $\mathbf{\hat{c}} + \delta \frac{d \mathcal{L}(\mathbf{x}^\star ( \mathbf{ \mathbf{\hat{c}}} ) )}{ d \mathbf{x} }|_{ \mathbf{x} =\mathbf{x}^\star ( \mathbf{ \hat{c}} )}$. 
\citeA{PogancicPMMR20} call this `differentiation of blackbox' (DBB) solvers, because this  approach considers the CO solver as a blackbox oracle, i.e., it does not take into consideration how the solver works internally. 

In a subsequent work, \citeA{sahoo2022gradient} propose to treat the CO solver as as a negative identity matrix while backpropagating the loss, i.e., they propose the following approximation of the derivative:
\begin{align}
     \frac{d \mathcal{L}(\mathbf{x}^\star ( \mathbf{ \mathbf{\hat{c}}} ) )} { d \mathbf{ \hat{c} }} \approx    - \frac{d \mathcal{L}(\mathbf{x}^\star ( \mathbf{ \hat{c}} ) )}{d \mathbf{x}^\star ( \mathbf{ \hat{c}} )} 
     \label{eq:dbb}
\end{align}
However, they notice that such an approach might run into unstable learning for scale-invariant optimization problems such as LPs and ILPs.
To negate this effect, they suggest
multiplying the cost vector with the matrix of the invariant transformation.
In the case of LPs and ILPs this can be achieved by \textbf{normalizing the cost vector through projection onto the unit sphere}.
\paragraph{Perturb-and-MAP.}
However, at this point it is worth mentioning that \citeA{Domke10} assumes, in his proposition, that the distribution $p(\theta_1|\theta_2)$ in \eqref{eq:domke} belongs to the exponential family of distributions \cite{exponentialbarndorff}. 
Note that the distribution defined in \eqref{eq:argmax_distri} is not a distribution of the exponential family. 
Nevertheless, a tempered softmax \cite{hinton2015distilling} distribution belonging to exponential family can be defined to express the mapping in the following way:
\begin{equation}
     p_{\tau}( \mathbf{x}|\mathbf{c}) = \begin{cases}
          \frac{\exp(- f(\mathbf{x}, \mathbf{c}) / \tau  ) }{ \sum_{\mathbf{x}^\prime \in \mathcal{F}} \exp(- f(\mathbf{x}^\prime, \mathbf{c}) / \tau ) } & \mathbf{x} \in  \mathcal{F}\\
          0 & \text{otherwise}
     \end{cases}
     \label{eq:tempered_softmax}
\end{equation}
In this case, the log unnormalized probability mass at each $\mathbf{x} \in \mathcal{F}$ is proportional to $\exp(- f(\mathbf{x}, \mathbf{c}) / \tau  )$, the exponential of the negative of the tempered objective value.
The idea behind \eqref{eq:tempered_softmax} is to assign a probability to each feasible solution such that solutions with a better objective value have a larger probability. 
The parameter $\tau$ affects the way in which objective values map to probabilities. When $\tau \to 0 $, the distribution becomes the argmax distribution in \eqref{eq:argmax_distri}, when $\tau \to \infty$, the distribution becomes uniform. In other words, the value of $\tau$ determines how drastically the probability changes because of a change in objective value. Good values for $\tau$ are problem-dependent, and thus tuning $\tau$ is advised.

Note that \eqref{eq:domke} deals with conditional expectation.
As in the case of the tempered softmax distribution, the conditional expectation is not always equal to the solution to the CO problem, it must be computed first to use the finite difference method in \eqref{eq:domke}.
However, computing the probability distribution function in \eqref{eq:tempered_softmax} is not tractable, as the denominator (also called the partition function) requires iterating over all feasible points in $\mathcal{F}$. Instead, \citeA{PapandreouY11} propose a novel approach, known as \emph{perturb-and-MAP}, to estimate the probability using perturbations. 
It states that the distribution of the maximizer after perturbing the log unnormalized probability mass by i.i.d. Gumbel$(0,\epsilon)$ noise has the same exponential distribution as \eqref{eq:tempered_softmax}.
To make it more explicit,
if $\Tilde{\mathbf{c}} = \mathbf{c} + \bm{\eta}  $, where the perturbation vector $\bm{\eta} \overset{ \text{i.i.d.} }\sim \text{Gumbel}(0,\epsilon) $,
\begin{equation}
     \label{eq:perturbmap}
     \mathbb{P} [ \mathbf{x} = \argmax_{ \mathbf{x}^\prime} -f(\mathbf{x}^\prime, \Tilde{\mathbf{c}})] = p_{\epsilon}(\mathbf{x}| \mathbf{c}) 
\end{equation}
The perturb-and-MAP framework can be viewed as a method of stochastic smoothing~\cite{abernethy2016perturbation}. A smoothed approximation of the optimization mapping is created by considering the average value of the solutions of a set of \emph{nearby perturbed} points.
With the help of \eqref{eq:perturbmap}, the conditional distribution and hence the conditional mean can be approximated by Monte Carlo simulation.

\paragraph{Differentiable perturbed optimizers.}
\citeA{berthet2020learning} propose another approach for perturbation-based differentiation. They name it differentiable perturbed optimizers (DPO). 
They make use of the perturb-and-MAP framework to draw samples from the conditional distribution $p( \mathbf{x} |\mathbf{c})$. 
In particular, they use the reparameterization trick~\cite{KingmaW13,rezende14} to generate samples from $p( \mathbf{x} |\mathbf{c})$.
The reparameterization trick uses a change of variables to rewrite $\mathbf{x}$ as a \emph{deterministic function} of $\mathbf{c}$ and a random variable $\bm{\eta}$. In this reformulation, $\mathbf{x}$ is still a random variable, but the randomness comes from the variable $\bm{\eta}$.
They consider $\bm{\eta}$ to be  a random variable having a density proportional to $\exp(- \nu(\bm{\eta}))$ for a twice-differentiable function $\nu$.
Moreover, they propose to multiply the random variable $\bm{\eta}$ with a temperature parameter $\epsilon  >0$, which controls the strength of perturbing $\mathbf{c}$ by the random variable $\bm{\eta}$. 
In summary, first $\mathbf{c}$ is perturbed with random perturbation vector $\epsilon\bm{\eta}$, where $\bm{\eta}$ is sampled from the aforementioned density function, and then the maximizer of the perturbed vector $c + \epsilon \bm{\eta}$ is viewed as a sample from the conditional distribution, i.e., $\mathbf{x}^\star_\epsilon ( \mathbf{c}) = \mathbf{x}^\star (\mathbf{c} + \epsilon \bm{\eta}) $ is considered as a sample drawn from $p( \mathbf{x} |\mathbf{c})$ for given values of $\mathbf{c}$ and $\epsilon$.
They call $\mathbf{x}^\star_\epsilon ( \mathbf{c}) $ a \emph{perturbed optimizer}.
Note that, for $\epsilon \to 0$, $\mathbf{x}^\star_\epsilon (\mathbf{c}) \to \mathbf{x}^\star (\mathbf{c})$.
Like before, $\mathbf{x}^\star_\epsilon (c)$ can be estimated by Monte Carlo simulation by sampling i.i.d. random noise $\bm{\eta}^{(m)}$ from the aforementioned density function.
The advantage is that the Monte Carlo estimate is \emph{continuously} differentiable with respect to $\mathbf{c}$. 
This Monte Carlo estimate $\bar{ \mathbf{x} }^\star_\epsilon (\mathbf{c})$ can be expressed as:
\begin{align}
     \bar{ \mathbf{x} }^\star_\epsilon (\mathbf{c}) &= \frac{1}{M} \sum_{m=1}^M \mathbf{x}^\star \Big( \mathbf{c} + \epsilon \bm{\eta}^{(m)} \Big)
     \label{eq:perturbed}
\end{align}
Moreover, its derivative can be 
estimated by Monte Carlo simulation too
\begin{equation}
     \frac{d \bar{ \mathbf{x} }^\star_\epsilon (\mathbf{c}) }{d \mathbf{c}} =  \frac{1}{\epsilon} \frac{1}{M} \sum_{m=1}^M  \mathbf{x}^\star ( \mathbf{c} + \epsilon \bm{\eta}^{(m)} ) \nu^\prime (\bm{\eta}^{(m)})^\top 
     \label{eq:jacobian_dpo}
\end{equation}
where $\nu^\prime$ is the first order derivative of $\nu$.
They can approximate $\frac{d \mathbf{x}^\star ( \mathbf{c}) }{d \mathbf{c}}$  by $\frac{d \bar{ \mathbf{x} }^\star_\epsilon ( \mathbf{c}) }{d \mathbf{c}}$ to implement the backward pass.
As mentioned before, if $\epsilon \to 0$, the estimation will be an unbiased estimate of $\mathbf{x}^\star ( \mathbf{c})$.
However, in practice, for low values of $\epsilon$, the variance of the Monte-Carlo estimator will increase, leading to unstable and noisy gradients.
This is in line with the smoothing-versus-accuracy trade-off mentioned before.
\citeA{berthet2020learning} use this DPO framework to differentiate any CO problem with linear objective. For a CO problem with discrete feasible space, they consider the convex hull of the discrete feasible region.
Furthermore, \citeA{berthet2020learning} construct the Fenchel-Young loss function and show for Fenchel-Young loss function, the gradient can be approximated in the following way:
\begin{align}
     \nabla  \mathcal{L}^{FY} (\mathbf{x}^\star(\mathbf{\hat{c}}) ) = -\big( \bar{ \mathbf{x} }^\star_\epsilon ( \mathbf{\hat{c}})  - \mathbf{x}^\star (\mathbf{c}) \big)
     \label{eq:fy}
\end{align}

In a later work, \citeA{dalle2022learning} extend the
perturbation approach, where they consider multiplicative perturbation. 
This is useful when the cost parameter vector is restricted to be non-negative, such as in the applications of shortest path problem variants.
The work of \citeA{SSTPaulusCT0M20} can also be viewed as an extension of the DPO framework.
They introduce stochastic softmax tricks (SST), a framework of Gumbel-softmax distributions, where they propose differentiable methods by sampling from more complex categorical distributions. 
\paragraph{Implicit maximum likelihood estimation (I-MLE).}
The \emph{perturb-and-MAP} framework is also used by \citeA{niepert2021implicit}.
However, they do not sample noise from the Gumbel distribution, rather they report better results when the noise $\bm{\eta}^\gamma$ is sampled from a \emph{Sum-of-Gamma} distribution with hyperparameter $\gamma$.
Combining the finite difference approximation \eqref{eq:domke} with the perturb-and-MAP framework, the  gradient takes the following form:
\begin{align}
     \frac{d \mathcal{L}(\mathbf{x}^\star ( \mathbf{ \hat{c}} ) )} { d \mathbf{\hat{c}}} \approx \nabla^{(IMLE)}\mathcal{L}(\mathbf{x}^\star ( \mathbf{ \hat{c}} ) ) =   \Bigg(  \mathbf{x}^\star \Big(\mathbf{\hat{c}} + \delta \frac{d \mathcal{L}(\mathbf{x}^\star ( \mathbf{ \hat{c}} ) )}{d \mathbf{x}^\star ( \mathbf{ \hat{c}} )} + \epsilon \bm{\eta}^\gamma \Big) -  \mathbf{x}^\star \Big( \mathbf{\hat{c}} + \epsilon \bm{\eta}^\gamma \Big)  \Bigg)
     \label{eq:imle}
\end{align}
\noindent where $\delta$ is the step size of
the finite difference approximation and $\epsilon  >0$ is a temperature parameter, which controls the strength of noise perturbation. 
Clearly, \eqref{eq:imle} turns into \eqref{eq:dbb} when there is no noise perturbation, i.e., if $\bm{\eta}^\gamma =0$.
A later work \cite{AIMLE} extends I-MLE by adaptively selecting $\delta$ based on the ratio between the norm of the parameter $\mathbf{\hat{c}}$ and the norm of $\frac{d \mathcal{L}(\mathbf{x}^\star ( \mathbf{ \hat{c}} ) )}{d \mathbf{x}^\star ( \mathbf{ \hat{c}} )}$.
\paragraph{Discussion.}
One major advantage of the techniques explained in this subsection is that for gradient computation, they call the CO solver as a `blackbox oracle' and only use the solution returned by it for gradient computation.
In essence, these techniques are not concerned with \textit{how} the CO problem is solved. The users can utilize any techniques of their choice---constraint programming (CP) \cite{rossi2006handbook}, Boolean satisfiability (SAT) \cite{gomes2008satisfiability} or linear programming (LP) and integer linear programming (ILP) to solve the CO problem. 
\subsubsection{Differentiation of Surrogate Loss Functions}
\label{sect:surrogate}
The techniques explained in the preceding subsections can be viewed as implementations of differentiable optimization layers, which solve the CO problem in the forward pass and return useful approximations of $\frac{ d \mathbf{x}^\star ( \mathbf{ \hat{c}} )}{ d \mathbf{ \hat{c} }}$ in the backward pass. 
Consequently, those techniques can be used to introduce optimization layers \emph{anywhere in a neural network architecture}, and can be combined with arbitrary loss functions.
In contrast, the techniques that will be introduced next can only be used to differentiate regret \eqref{eq:regret}---a specific task loss. Hence, models can only be trained in an end-to-end fashion using these techniques when the CO problem occurs in the \textit{final} stage of the pipeline, as in the case of \pto problems.
Also note that the computation of the regret requires both the ground-truth cost vector $\mathbf{c}$; as well as ground-truth solution $\mathbf{x}^\star (\mathbf{c})$. 
If $\mathbf{c}$ is observed, $\mathbf{x}^\star (\mathbf{c})$ can be computed. However, if only $\mathbf{x}^\star ( \mathbf{c})$ is observed, $\mathbf{c}$ cannot directly be recovered. 
Hence, the techniques that will be discussed next are not suitable when the true cost vectors $\mathbf{c}$ are not observed in the training data.

\paragraph{Smart ``Predict, Then Optimize''.} \citeA{elmachtoub2022smart} develop Smart ``Predict, Then Optimize'' (SPO), a seminal work in DFL.
As the gradient of the regret with respect to cost vector $\mathbf{ \hat{c} }$ is zero almost everywhere, SPO instead uses a surrogate loss function that has subgradients which \textit{are} useful in training.
They start by proposing a convex surrogate upper bound of regret, which they call the SPO+ loss. 
\begin{equation}
    \mathcal{L}_{SPO+} (\mathbf{x}^\star(\mathbf{\hat{c}}) ) = 2 \mathbf{\hat{c}}^\top \mathbf{x}^\star (\mathbf{c}) - \mathbf{c}^\top \mathbf{x}^\star (\mathbf{c}) +  \max_{ \mathbf{x} \in \mathcal{F} } \{ \mathbf{c}^\top \mathbf{x} - 2 \mathbf{\hat{c}}^\top \mathbf{x} \} 
\end{equation}
\citeA{elmachtoub2022smart} directly minimize $\mathcal{L}_{SPO+}$ for a linear predictive model.
However, they note that directly minimizing $\mathcal{L}_{SPO+}$ is prohibitive for complex predictive models such as NNs.
Hence, they derive the following useful subgradient of $ \mathcal{L}_{SPO+} (\mathbf{x}^\star(\mathbf{\hat{c}}))$:
\begin{align}
     \mathbf{x}^\star (\mathbf{c}) - \mathbf{x}^\star(2 \mathbf{\hat{c}} - \mathbf{c}) \in \partial \mathcal{L}_{SPO+}
    \label{eq:spo}
\end{align}
This subgradient is used 
in the backward pass for training NNs. While this technique proposes a surrogate loss function, one could also view the solving of $\mathbf{x}^\star(2 \mathbf{\hat{c}} - \mathbf{c})$ as solving of $\mathbf{\hat{c}}$ perturbed by the deterministic perturbation vector $\mathbf{\hat{c}} - \mathbf{c}$. 

From a theoretical point of view, the SPO+ loss has the Fisher consistency property with respect to the regret under certain distributional assumptions.
A surrogate loss function satisfies the Fisher consistency property if the function that minimizes the surrogate loss also minimizes the true loss in expectation \cite{zou2008new}. 
Concretely, this means that minimizing the SPO+ loss corresponds to minimizing the regret in expectation.
While training ML models with a finite dataset, an important property of considerable interest would be \emph{risk bounds} \cite{massart2006risk}.
\citeA{liu2021risk} develop risk bounds for SPO+ loss and show that low excess SPO+ loss risk translates to low excess regret risk. 
Furthermore, \citeA{BalghitiEGT19} develop worst-case generalization bounds of the SPO loss.  

The SPO framework is applicable not only to LPs, but to \emph{any CO problems where the cost parameters appear linearly in the objective function}. This includes QPs, ILPs and MILPs.  \citeA{mandi2020smart} empirically investigated how the framework performs on ILP problems. However, as these problems are much more computationally expensive to solve than the ones considered by \citeA{elmachtoub2022smart}, they compared the standard SPO framework with a variant in which,
it is significantly cheaper to solve the CO problem during training. To be specific, they consider LP relaxations of the ILPs.
These LP relaxations are obtained by considering the continuous relaxation of the ILPs, i.e., they are variants of the ILPs in which the integrality constraints are dropped. Using the LP relaxations significantly expedite training, without any cost: \citeA{mandi2020smart} did not observe a significant difference in the final achieved regret between these two approaches, with both of them performing better than the prediction-focused approach. However, one should be cautious to generalize this result across different problems, as it might be dependent on the integrality gap between the ILP and its LP relaxation.

Next, within this category, a different type of DFL techniques is being surveyed.
In these DFL techniques, the surrogate loss functions are supposed to reflect the decision quality, but their computations do \emph{not} involve solving the CO problems, thereby avoiding the zero-gradient problem.

\paragraph{Noise contrastive estimation.}
One such approach is introduced by \citeA{mulamba2020discrete}. 
Although their aim is still to minimize regret, computation of $\nabla_{\mathbf{\hat{c}}}\mathit{Regret}\; (\mathbf{x}^\star(\mathbf{\hat{c}}),\mathbf{ c})$
has been avoided by using a surrogate loss function.
In their work, the CO problem is viewed from a probabilistic perspective, as in \eqref{eq:tempered_softmax}. However, instead of maximum likelihood estimation, the noise contrastive estimation (NCE)~\cite{gutmann2010noise} method is adopted.
NCE has been extensively applied in many applications such as language modeling \cite{icml/MnihT12}, information retrieval \cite{cikm/HuangHGDAH13} and entity linking \cite{gillick-etal-2019-learning}. 
Its basic idea is to learn to discriminate between data coming from the true underlying distribution and data coming from a noise distribution. 
In the context of DFL, this involves contrasting the likelihood of ground-truth solution $\mathbf{x}^\star ( \mathbf{c})$ and a set of negative examples $S$. In other words, the following ratio is maximized:
\begin{equation}
    \label{eq:nce_likelihood}
    \max_{ \mathbf{\hat{c}}} \sum_{\mathbf{x}^\prime \in S}\frac{p_\tau (\mathbf{x}^\star ( \mathbf{c})|\mathbf{\hat{c}}) }{p_\tau (\mathbf{x}^\prime | \mathbf{\hat{c}})}
\end{equation}
where $\mathbf{x}^\prime \in S$ is a negative example.
Because the probability $p_\tau (\mathbf{x}^\star ( \mathbf{c})| \mathbf{\hat{c}})$ is defined as in \eqref{eq:tempered_softmax}, when $\tau = 1$, maximizing \eqref{eq:nce_likelihood} corresponds to minimizing the following loss:
\begin{align}
     \mathcal{L}_{NCE}(\mathbf{\hat{c}}, \mathbf{c}) = \sum_{\mathbf{x}^\prime \in S}  f(\mathbf{x}^\star (\mathbf{c}), \mathbf{\hat{c}}) -  f(\mathbf{x}^\prime, \mathbf{\hat{c}})
   \label{eq:NCE_expect}
\end{align}
In other words, this approach learns to predict a $\mathbf{ \hat{c} }$ for which ground-truth solution $\mathbf{x}^\star (\mathbf{c})$ achieves a good objective value, and for which other feasible solutions $\mathbf{x}^\prime$ achieve worse objective values. 
Note that when $ f(\mathbf{x}^\star (\mathbf{c}), \mathbf{\hat{c}}) \leq f(\mathbf{x}^\prime, \mathbf{\hat{c}})$ for all $\mathbf{x}^\prime \in \mathcal{F}$, it holds that $\mathbf{x}^\star ( \mathbf{c}) = \mathbf{x}^\star ( \mathbf{ \hat{c}} )$, and thus the regret is zero.
Also note that computing $\mathcal{L}_{NCE}(\mathbf{\hat{c}}, \mathbf{c})$ does not involve computing $\mathbf{x}^\star( \mathbf{\hat{c}})$, circumventing the zero-gradient problem.

As an alternative to NCE, \citeA{mulamba2020discrete} also introduce a maximum a posteriori (MAP) approximation, in which they only contrast the ground-truth solution with the most probable negative example from $S$ according to the current model:
\begin{align}
     \mathcal{L}_{MAP}( \mathbf{\hat{c}},  \mathbf{ c}) &= \max_{\mathbf{x}^\prime \in S}  f(\mathbf{x}^\star (\mathbf{c}), \mathbf{\hat{c}}) -  f(\mathbf{x}^\prime, \mathbf{\hat{c}}) \nonumber \\ &= f(\mathbf{x}^\star (\mathbf{c}), \mathbf{\hat{c}}) -  f(\mathbf{x}^\prime, \mathbf{ \hat{c}}) \text{ where } \mathbf{x}^\prime = \argmin_{ \mathbf{x} \in S}f(\mathbf{x}, \mathbf{\hat{c}})
   \label{eq:MAP_expect}
\end{align}
Note that whenever $\mathbf{x}^\star(\mathbf{\hat{c}}) \in S$, it holds that $\mathcal{L}_{MAP}( \mathbf{\hat{c}},  \mathbf{ c}) = f(\mathbf{x}^\star (\mathbf{c}), \mathbf{\hat{c}}) -  f(\mathbf{x}^\star( \mathbf{\hat{c}}), \mathbf{\hat{c}})$. This is also known as \textit{self-contrastive estimation} (SCE) \cite{goodfellow2014distinguishability} since the ground-truth is contrasted with the most likely output of the current model itself.

Also note that for optimization problems with a linear objective, the losses are
$\mathcal{L}_{NCE}( \mathbf{\hat{c}},  \mathbf{ c}) = \sum_{\mathbf{x}^\prime \in S}   \mathbf{\hat{c}}^\top (\mathbf{x}^\star (\mathbf{c}) -  \mathbf{x}^\prime)$
and
$\mathcal{L}_{MAP}( \mathbf{\hat{c}},  \mathbf{ c}) = \mathbf{\hat{c}}^\top (\mathbf{x}^\star (\mathbf{c}) - \mathbf{x}^\prime)$, where $\mathbf{x}^\prime = \argmin_{\mathbf{x} \in S}f(\mathbf{x}, \mathbf{\hat{c}})$. 
In order to prevent the model from simply learning to predict $\mathbf{\hat{c}}= \mathbf{0}$, the following alternate loss functions are proposed for these kinds of problems: 
\begin{align} 
\label{eq:NCE_lin_obj}
\mathcal{L}_{NCE}^{(\mathbf{\hat{c}}  - \mathbf{c} )}( \mathbf{\hat{c}},  \mathbf{ c}) = \sum_{\mathbf{x}^\prime \in S}   (\mathbf{\hat{c}}  - \mathbf{c} )^\top (\mathbf{x}^\star (\mathbf{c}) -  \mathbf{x}^\prime)
\end{align}
\begin{align} 
\label{eq:MAP_lin_obj}
\mathcal{L}_{MAP}^{(\mathbf{\hat{c}}  - \mathbf{c} )}( \mathbf{\hat{c}},  \mathbf{ c}) = \max_{\mathbf{x}^\prime \in S}   (\mathbf{\hat{c}}  - \mathbf{c} )^\top (\mathbf{x}^\star (\mathbf{c}) - \mathbf{x}^\prime) 
\end{align}
\raggedbottom
\paragraph{Construction of $S$.} Forming $S$ by sampling points from the feasible region $\mathcal{F}$ is a crucial part of using the contrastive loss functions. To this end, \citeA{mulamba2020discrete} proposes to construct $S$ by caching all the optimal solutions in the training data. That is why they name $S$ as `solution cache'. 
While training, more feasible points are gradually added to $S$ by
solving for some of the predicted cost vectors. 
However, in order to avoid computational cost, the solver call is not made for 
each predicted cost during training. 
Whether to solve for a predicted cost vector is decided by pure random sampling, i.e., is based on a biased coin toss with probability $p_{solve}$.
Intuitively, the $p_{solve}$ hyperparameter determines the proportion of instances for which the CO problem is solved during training.
Experimentally, it has been reported that $p_{solve}=5\%$ of the time is often adequate, which translates to solving for only $5\%$ of the predicted instances. 
This translates to reducing the computational cost by approximately $95\%$, since solving the CO problems represents the major bottleneck in terms of computation time in DFL training.

\paragraph{Approximation of a solver by a solution-cache.}
Furthermore, \citeA{mulamba2020discrete} propose a solver-free training variant for any DFL technique that treats the CO solver as a blackbox oracle. Such techniques include the aforementioned I-MLE, DBB, SPO. 
In this solver-free implementation,
solving the CO problem is substituted with a cache lookup strategy, where the minimizer within the cache $S \subset \mathcal{F}$ is considered as a proxy for the solution to the CO problem (i.e., the minimizer within $\mathcal{F}$). 
This significantly reduces the computational cost as solving an CO problem is replaced by a linear search within a limited cache.
Such an approximation can be useful in case the CO problem takes long to solve.
\begin{table}[]
    \centering
    \resizebox{\linewidth}{!}{
    \begin{tabular}{cccc}
    \rowcolor{black} 
    \color{white} 
    Technique & \color{white} CO Problem Forms  & \color{white} Computation of Gradient & \color{white} \makecell{Differentiable \\Optimization\\ Layer} \\
     \toprule
    \makecell{OptNet\\ \cite{amos2017optnet}} & Convex QPs & \makecell{Implicit differentiation of\\ KKT conditions} & \coloredtickmark{green} \\[3mm]
    \makecell{Cvxpylayers\\ \cite{agrawal2019differentiable}} & \makecell{Convex problems} & \makecell[c]{Implicit differentiation of\\ HSD of conic programs} &\coloredtickmark{green}\\[3mm]
    \makecell{Fold-opt\\ \cite{kotary2023folded}} & \makecell{Convex and nonconvex\\problems}  & \makecell{Implicit differentiation\\ based on unrolling} & \coloredtickmark{green}\\[3mm]
    \makecell{QPTL\\ \cite{aaai/WilderDT19}} & LPs, ILPs & \makecell[c]{Implicit differentiation\\ after transforming into QPs\\ by  adding regularizer} & \coloredtickmark{green}\\[5mm]
    \makecell{Intopt \\ \cite{mandi2020interior}} & LPs, ILPs & \makecell[c]{Implicit differentiation of\\ HSD of (relaxed) LPs by\\ adding log-barrier relaxation } & \coloredtickmark{green}\\[5mm]
    \makecell{Mipaal\\ \cite{aaaiFerberWDT20}} & ILPs & \makecell[c]{Conversion of ILPs into LPs \\by method of cutting planes \\before applying QPTL} & \coloredtickmark{green} \\[5mm]
    \makecell{DBB \\ \cite{PogancicPMMR20}} & \makecell{Optimization problems \\with a linear objective} & \makecell[c]{Differentiation of\\ linear interpolation\\ of optimization mapping} & \coloredtickmark{green} \\[5mm]
    \makecell{Negative identity \\ \cite{sahoo2022gradient}}  & \makecell{Optimization problems \\with a linear objective} & \makecell[c]{Treating the CO solver as\\ negative identity mapping} & \coloredtickmark{green} \\[3mm] 
    \makecell{I-MLE \\ \cite{niepert2021implicit}}  & \makecell{Optimization problems \\with a linear objective} & \makecell[c]{Finite difference approximation\\ with perturb-and-MAP } & \coloredtickmark{green} \\[3mm] 
    \makecell{DPO\\ \cite{berthet2020learning}} & \makecell{Optimization problems \\with a linear objective} & \makecell[c]{Differentiation of\\ perturbed optimizer} & \coloredtickmark{green} \\[3mm]
    \makecell{FY\\ \cite{berthet2020learning}} & \makecell{Optimization problems \\with a linear objective} & \makecell[c]{Differentiation of perturbed\\ Fenchel-Young loss} & \coloredcross{red} \\[3mm]
    \makecell{SPO\\ \cite{elmachtoub2022smart}} & \makecell{Optimization problems \\with a linear objective} & \makecell[c]{Differentiation of\\ surrogate  SPO+ loss} & \coloredcross{red} \\[3mm]
    \makecell{NCE\\ \cite{mulamba2020discrete}} & \makecell{Generic optimization\\ problems} & \makecell[c]{Differentiation of\\ surrogate contrastive loss} & \coloredcross{red} \\[3mm]
    \makecell{LTR\\ \cite{mandi22-pmlr}} & \makecell{Generic optimization\\ problems} & \makecell[c]{Differentiation of\\surrogate LTR loss} & \coloredcross{red} \\[3mm]
    \makecell{LODL\\ \cite{shahdecision}} & \makecell{Generic optimization\\ problems} &
    \makecell[c]{Differentiation of a \\learned convex local surrogate loss} & \coloredcross{red} \\[2mm]
     \bottomrule
    \end{tabular}
    }
    \caption{A concise overview of gradient modeling techniques in key DFL techniques that use gradient-based learning.}
    \label{tab:dfl_formula}
\end{table}

\paragraph{DFL as a learning to rank (LTR) problem.} In a later work, \citeA{mandi22-pmlr} observe that $\mathcal{L}_{NCE}$ \eqref{eq:NCE_expect} can be derived by formulating DFL as a \textit{pairwise learning to rank} task~\cite{Joachims02}. 
The learning to rank task consists of learning the implicit order over the solutions in $S$ invoked by the objective function values achieved by the solutions with respect to $\mathbf{c}$. 
In other words, it involves learning to predict a $\mathbf{ \hat{c} }$ that ranks the solutions in $S$ similarly to how $\mathbf{c}$ ranks them.
In the \textit{pairwise} approach, $\mathbf{x}^\star (\mathbf{c})$ and any $\mathbf{x}^\prime \in S$ are treated as a pair and the model is trained to predict $\mathbf{ \hat{c} }$ such that the ordering of each pair is the same for $\mathbf{c}$ and $\mathbf{ \hat{c} }$. The loss is considered to be zero if $ \mathbf{\hat{c}}^\top \mathbf{x}^\star (\mathbf{c})$ is smaller than $\mathbf{\hat{c}}^\top \mathbf{x}^\prime$ by at least a margin of $\Theta > 0$. The pairwise loss is formally defined in the following form:
\begin{align}
     \mathcal{L}_{\text{Pairwise}}( \mathbf{\hat{c}},  \mathbf{ c})  =  
      \sum_{\mathbf{x}^\prime \in S}   \max \left(0,  \Theta + (f(\mathbf{x}^\star (\mathbf{c}), \mathbf{\hat{c}}) -  f(\mathbf{x}^\prime, \mathbf{\hat{c}}))  \right)
    \label{eq:pairwise}
\end{align}
Another loss function is formulated by considering the difference in differences between the objective values at the true optimal $\mathbf{x}^\star ( \mathbf{c})$ and non-optimal $\mathbf{x}^\prime$ with $\mathbf{c}$ and $\mathbf{ \hat{c} }$ as the parameters.
\begin{align}
      \mathcal{L}_{\text{PairwiseDifference}} ( \mathbf{\hat{c}},  \mathbf{ c}) = \sum_{\mathbf{x}^\prime \in S}  \bigg(  \big( f(\mathbf{x}^\star (\mathbf{c}), \mathbf{\hat{c}}) - f(\mathbf{x}^\prime,  \mathbf{\hat{c}})   \big) - \big(f(\mathbf{x}^\star (\mathbf{c}),\mathbf{ c}) -  f(\mathbf{x}^\prime,\mathbf{ c})\big)  \bigg)^2
    \label{eq:pairwise_diff}
\end{align}

Further, motivated by the \textit{listwise learning to rank task}~\cite{cao2007learning}, a loss function is proposed by \citeA{mandi22-pmlr} where the ordering of all the items in $S$ is considered, rather than the ordering of pairs of items. 
\citeA{cao2007learning} define this listwise loss based on a \emph{top-one probability} measure. The top-one probability of an item is the probability of it being the best of the set. Note that such probabilistic interpretation in the context of DFL is already defined in Section \ref{sect:perturb}. 
\citeA{mandi22-pmlr} make use of the tempered softmax probability defined in \eqref{eq:tempered_softmax}. 
Recall that this $p_\tau(\mathbf{x}|\mathbf{c})$ can be interpreted as a probability measure of $\mathbf{x} \in \mathcal{F}$ being the minimizer of $f(\mathbf{x}, \mathbf{c})$ in $\mathcal{F}$ for a given $\mathbf{c}$.
However, as mentioned before, direct computation of $p_{\tau}( \mathbf{x}| \mathbf{c})$ requires iterating over all feasible points in $\mathcal{F}$, which is intractable. Therefore \citeA{mandi22-pmlr} compute the probability with respect to $S \subset \mathcal{F}$.
This probability measure finally is used to define a listwise loss---the cross-entropy loss between $p_\tau(\mathbf{x}| \mathbf{c})$ and $p_\tau(\mathbf{x}| \mathbf{\hat{c}})$, the distributions obtained for ground-truth $\mathbf{c}$ and predicted $\mathbf{ \hat{c} }$. 
This can be written in the following form:
\begin{align}
     \mathcal{L}_{\text{Listwise}} ( \mathbf{\hat{c}},  \mathbf{ c})=  \bigg( -\frac{1}{|S|} \sum_{\mathbf{x}^\prime \in S} p_{\tau}(\mathbf{x}^\prime|  \mathbf{c}) \log p_{\tau}(\mathbf{x}^\prime|\mathbf{\hat{c}}) \bigg)
    \label{eq:listwise}
\end{align}
\noindent The main advantage of \eqref{eq:NCE_expect}, \eqref{eq:MAP_expect}, \eqref{eq:pairwise}, \eqref{eq:pairwise_diff} and \eqref{eq:listwise} is that they are differentiable and can be computed directly by any neural network library via automatic differentiation. 
Also note that the computation and differentiation of the loss functions are solver-free, i.e., they need not solve the CO problem to compute the loss or its derivative.


\paragraph{Learning efficient surrogate losses.}
Another research direction without optimization in the loop focuses on minimizing the computational burden of solving the CO problems repeatedly. This is achieved by devising efficiently computable and differentiable surrogate losses that approximate and substitute the true task loss.
In this regard, \citeA{shahdecision} propose to learn a surrogate of the regret function using parametric local losses. Due to the difficulty of learning a single convex surrogate function to estimate regret, a convex local surrogate is learned for each data point in training, based on other points sampled around these data points. By design, the surrogate losses are automatically differentiable, and thus they eliminate the need for a differentiable optimization solver. In subsequent work, \citeA{leavingthenest} extended on this, by improving the sample efficiency of learning the surrogate losses, and by introducing a manner of producing surrogate losses that are faithful to the true regret also outside of the local neighbourhoods around the training examples.
\paragraph{DFL as a learning to optimize problem.}
 DFL is viewed as an extension of learning to optimize (LtO) by \citeA{kotary2023predict}.
 In LtO, the goal is to train an ML model as an approximate CO solver, which maps the parameters of a CO problem to its optimal solution. \citeA{kotary2023predict} show that the technique of LtO can be applied to the DFL setting, by treating the feature variables $\mathbf{z}$ as inputs to the LtO model, which predicts $\mathbf{x}^{\star}(\mathbf{c})$. The resulting LtO models function as joint prediction and optimization models, whose training is consistent with regret minimization on the DFL task. 

\subsection*{Discussion}\label{Sect:discussion}
So far, in this subsection, an extensive overview of different gradient-based DFL techniques has been provided.
For the ease of the readers, a summary of some of the key DFL techniques, discussed so far, have been provided in Table \ref{tab:dfl_formula}.
The second column of Table \ref{tab:dfl_formula} highlights the form of the CO problem applicable to the technique. 
Note that although some techniques are generally applicable to any CO problem forms, most techniques have been evaluated so far using CO problems with linear objective functions.
The third column summarizes the gradient computation technique. 
The fourth column indicates whether that particular technique is compatible with any generic task loss. 
Techniques, termed as implementations of \emph{differentiable optimization layers}, can be embedded in any stage of an NN architecture.
The other techniques are applicable where optimization is the final stage of the pipeline (such as in \pto problem formulations) and a particular loss (most often regret) is used as the task loss.

\subsection{Review of Gradient-Free DFL Methodologies}
The techniques reviewed so far implement DFL by gradient descent training, which is the go-to approach for training neural networks.
Next, we review DFL techniques, which do not rely on gradient-based training.
Note that, since these techniques do not utilize gradient-based training, the predictive models in most cases consist of either tree-based methods or explicitly specified models such as linear models.

First of all, as $\mathcal{L}_{SPO+} (\mathbf{x}^\star(\mathbf{\hat{c}}) )$ is a convex loss function, it can be minimzed outside gradient-based training if the predicive model is linear.
Firstly, since $\mathcal{L}_{SPO+} (\mathbf{x}^\star(\mathbf{\hat{c}}) )$ is a convex loss function, it can be minimized using any convex solver without using gradient-based training \emph{if} the predictive model is linear.
\citeA{elmachtoub2020decisiontree} extend SPO considering 
the predictive model to be a decision tree or ensemble of decision trees. Such models can be learned by recursive partitioning with respect to the regret directly, and thus do not require the use of the SPO+ surrogate loss function introduced by \citeA{elmachtoub2022smart}. Alternatively, the tree learning problem can be posed as a MILP and be solved by an off-the-shelf solver. 
\citeA{jeong22a} formulate the problem of minimizing regret  as a mixed-integer linear program (MILP) for a linear predictive model.
They start from the bilevel optimization formulation, introduced in \eqref{eq:bilevel_outer} and  \eqref{eq:bilevel_inner}.
First, the transition points where the solution of the lower level program \eqref{eq:bilevel_inner} changes are identified, and then the solution space is exhaustively partitioned, and for each partition the solution is annotated. This paves the way to construct a MILP formulation of the outer program \eqref{eq:bilevel_outer}. 
This MILP problem is solved to learn the parameters $\bm{\omega}$ of the linear predictive model. The resulting model is guaranteed to be globally optimal, which is not the case for gradient-based methods that might get stuck in a local optimum. However, their method is limited to ML models that are linear and optimization problems that are binary MILPs.

\citeA{Dynamic_2020} consider linear ML models 
and represent the objective function of the CO problem as a piece-wise linear function of the ML parameters. 
In this proposed technique, the ML parameters are updated via a coordinate descent algorithm, where each component of the cost vector is updated at a time to minimize the regret keeping other components fixed. 
This technique requires identifying the \emph{transition points}, where regret changes, as a function of each component of the cost parameter.
\citeA{Dynamic_2020} consider CO problems that can be solved by dynamic programming and identify the transition points using dynamic programming.
In a later work, \citeA{Divide_and_Conquer_Guler} extend this technique by employing a `divide-and-conquer' algorithm to identify the transition points for CO problems whose objective function is a bilinear function of the decision variables and the predicted parameters.
This development generalizes the previous work~\cite{Dynamic_2020} to cover a much broader class of CO problems and offers a substantial speed improvement.
The `branch \& learn' approach \cite{hu2022branch} also extends the approach by \citeA{Dynamic_2020} by developing a recursive algorithm, allowing to tackle CO problems that can be solved by recursion.

\subsection{Other Aspects of Decision-Focused Learning}\label{sect:other_aspect}
In the following, some aspects related to DFL, 
that have not yet been discussed in this survey paper so far, will be highlighted.
To begin with, it should be noted that certain CO problems may have \emph{multiple non-unique optimal solutions} for a given cost vector. This can occur for an LP when the cost vector of an LP is parallel to one of the faces of the feasible polyhedron.
Moreover, problems involving symmetric graphs,
whose solution can be transformed into other solutions through automorphisms \cite{graphmorph},
exhibit multiple optimal solutions.
If the predicted cost vector has multiple non-unique optimal solutions, each of these solutions result in different values of regret.
In such scenarios, \citeA{elmachtoub2022smart} propose to consider the worst-case regret.
If the set of optimal solutions of $\mathbf{ \hat{c} }$ is be represented by $\mathcal{X} ^\star ( \mathbf{\hat{c} })$,
the worst-case regret can be defined in the following form:
\begin{equation}
    \mathit{Regret}\; (\mathbf{x}^\star(\mathbf{\hat{c}}), \mathbf{c} )= \max_{\mathbf{x}^\star ( \mathbf{ \hat{c}} ) \in \mathcal{X} ^\star ( \mathbf{\hat{c}} )} f(\mathbf{x}^\star ( \mathbf{ \hat{c}} ), \mathbf{c}) - f(\mathbf{x}^\star(\mathbf{ c}), \mathbf{c})
    \label{eq:worstregret_lp}
\end{equation}
Having addressed the possibility of the presence of multiple non-unique optimal solutions in a CO problem, the focus now turns to other important facets of DFL.
\subsubsection{Prediction-Focused vs. Decision-Focused Learning}
DFL methodologies are expected to deliver lower regret than a PFL approach in \pto problems, as the ML model is directly trained to achieve low regret. However, as discussed before, the implementation of DFL poses significant challenges.
In fact, practitioners may be tempted to resort to a PFL approach to circumvent the computational costs associated with DFL, when dealing with real-world \pto problems. To motivate adoption of DFL methodologies among practitioners, it is crucial to investigate scenarios where DFL methodologies outperform the PFL approach. To this end, \citeA{ETOvsIEO} conduct a theoretical comparison of the limiting distributions of the optimality gaps between the two approaches in the context of stochastic optimization.
They show the PFL approach that does not consider the CO problem while training the model asymptotically outperforms the integrated prediction and optimization approach, employed by DFL methodologies, \emph{if} the underlying prediction model is well-specified. This is intuitive, as a well-specified model tends to produce highly accurate predictions, which can contribute to the success of the PFL approach. In such cases, the DFL methodologies  might perform worse than PFL since training in DFL involves \emph{approximate} gradients (because the true gradient is zero almost everywhere), whereas the gradient is well-defined for a PFL approach.
On the other hand, they show that if the model is not well-specified, a PFL approach performs suboptimally compared to the DFL approach. 
Hence, it is recommended to use DFL when there exists aleatoric or epistemic uncertainty.
As most real-world settings include various sorts of uncertainty---both aleatoric and epistemic---DFL methodologies are expected to outperform the PFL approach.
In a separate work, \citeA{Cameron_Hartford_Perils} show that the suboptimality of PFL becomes more pronounced in the presence of correlations between the predicted parameters. 

\subsubsection{Multi-task Decision-Focused Learning}
In most DFL works, a single task is considered. For instance,
in the shortest path benchmark considered by \citeA{elmachtoub2022smart}, 
the grid structure and the start and end nodes are the same in all instances. 
However, one often has to deal with multiple tasks at once, in which it would be convenient to make decision-focused predictions, without having to train a separate model for each task.   
A first step in this direction was recently taken by \citeA{multi-task-DFL}. They propose a way of training a model in a decision-focused way with respect to multiple tasks at once by considering two kinds of architectures. The first is a regular multi-layer perceptron that outputs a single vector $\mathbf{ \hat{c} }$ which is used in the different tasks. The different resulting task losses then get aggregated to inform the update to weights $\bm{\omega}$, i.e., the weights $\bm{\omega}$ are trained to produce a $\mathbf{ \hat{c} }$ that generally works well for the different tasks considered. The second architecture is a multi-headed one, consisting of one or more shared first layers, followed by a dedicated head for every task. This produces a different vector $\hat{c}_i$ for every task. Their results show that they can train a model that can make effective decision-focused predictions for multiple tasks at once, and that this is particularly beneficial when not that many training data are available. However, a remaining limitation is that the model can still not be trained with the aim of \textit{generalizing} to new tasks.

A related topic is studied by \citeA{dinh2024end}, which focuses on DFL in the context of fair multiobjective optimization. In the presence of multiple objective functions, many applications call for pareto-optimal solutions which are `fair' with respect to the competing objective functions. In such cases, it is common to optimize the Ordered Weighted Average (OWA) of those functions, which guarantees `equitably efficient' solutions \cite{ogryczak2003solving}. Incorporating OWA optimization in DFL faces a unique challenge, since the OWA objective itself is nondifferentiable.
\citeA{dinh2024end} mitigate by replacing the OWA function by its smooth Moreau envelope approximation. This yields a differentiable approximate OWA optimization, and enables joint DFL learning of each individual objective function's parameters to minimize regret of their fair OWA aggregation.

\subsubsection{Predicting Parameters in the Constraints}\label{sect:df_for_constraints}
The majority of the works in DFL aim to predict parameters in the objective function and assume that the feasible space is known. 
However, in many applications the unknown parameters occur in the constraints as well as in the objectives. 
When the parameters in the constraints are predicted and prescribed decisions are made using the predicted parameters, one major issue is that
the prescribed decisions might turn out to be \emph{infeasible} with respect to the true parameters.
In this case, task loss to minimize the suboptimality of the prescribed decisions is not enough, but it should also penalize if the prescribed decisions become infeasible.
Hence DFL methods for such problems entails a few additional considerations. 
The first consideration deals with quantifying the extent of infeasibility when the prescribed decisions become infeasible with respect to the true parameters. 
In this regard, 
\citeA{garcia2022applicationdriven} propose to add artificial slack variables with high penalty costs in the objective function to penalize infeasible decisions.
\citeA{PredictOptimizeforPacking} introduce the notion of \emph{post-hoc} regret.
In \emph{post-hoc} regret a non-negative penalty is added to regret to account for correcting the infeasible solutions into feasible ones.
This idea of such a corrective actions shares a fundamental resemblance 
 to the concept of recourse actions in stochastic programming \cite{stochasticprogrammingmodels}.

The next consideration is computing the gradients of this task loss with respect to the parameters in the constraints.
Some of the techniques discussed in Section \ref{sect:kkt} can be utilized for this purpose.
For example, the gradient can be obtained by solver unrolling.
\citeA{TanDT19} compute the gradient by unrolling a LP. 
As the parameters in the constraints are also present in the
the KKT conditions \eqref{eq:KKT}, it is possible to compute the gradients for optimization problems, with differentiable constraint functions by differentiating the KKT conditions using the techniques discussed in Section \ref{sect:kkt}.

\citeA{TanNEURIPS2020} provide an ERM formulation for predicting parameters in both the objective function and constraints of an LP.
For parameters in the objective function, the ERM accounts for suboptimality of the decision and for parameters within the constraints, the ERM formulation considers the feasibility of optimal points observed in the training data.
%
%
This ERM formulation takes the form of a non-linear optimization program and they propose to compute the derivative of the predicted parameters by considering its sequential quadratic programming (SQP) approximation.
For packing and covering LPs, \citeA{PredictOptimizeforPacking} compute the derivative of \emph{post-hoc} regret with respect to the predicted parameters in the constraints by considering the Lagrangian relaxation of the LP when the corrective action is linear.
\citeA{nips/HuLL23} extend this approach by treating the correction action also as an LP.

The task of computing the gradients of the task loss with respect to the parameters in the constraints is particularly challenging for combinatorial optimization problems, which often involve discrete feasible space. 
For combinatorial optimization problems, it might happen that no constraints are active at the optimal point.
So, slight changes in the parameters in the constraints do not change the optimal solution, leading to the problem of zero gradients. Hence, coming up with meaningful gradients for back-propagation is more challenging for combinatorial optimization problems.
 If either the CO problem or the corrective action is an ILP, \citeA{nips/HuLL23} consider the corresponding LP relaxation.
\citeA{paulus2021comboptnet} develop a differentiable optimization layer for ILPs that considers the downstream gradient of the solution as an input and returns the directions of updating the parameters in the backward pass. 
They update the parameters along the directions so that the Euclidean distance between the solution of the updated parameter and the updated solution with the downstream gradient is minimized. 
For ILPs, \citeA{nandwani2022a} view the task of constraint learning from the lens of learning hyperplanes, which is common in classification tasks. Such an approach requires negative samples. However, the negative samples in this setting must also include infeasible points, which is not the case for the solution-cache approach reviewed in Section \ref{sect:surrogate}. 
In the realm of gradient-free methods, \citeA{hu2023branch} apply `branch \& learn' \cite{hu2022branch} to minimize \emph{post-hoc} regret considering CO problems, solvable by recursion.

\subsubsection{Model Robustness in Decision-Focused Learning} 
The issue of model robustness arises often in deep learning. As has been shown in many works, it is often possible for malicious actors to craft inputs to a neural network in such a way that the output is manipulated (evasion attacks) \cite{goodfellow2014explaining}, or to generate training data which cause adverse effects on the performance of the trained model (poisoning attacks). As a subset of ML, some adversarial settings also apply in DFL. 

Evasion attacks, despite being the most commonly studied adversarial attacks, do not generalize straightforwardly to DFL since they inherently pertain to classification models with finite output spaces. On the other hand, it is shown by \citeA{kinsey2023exploration} that effective poisoning attacks can be made against DFL models. The paper shows that while such attacks can be effective, they are computationally expensive due to the optimization which must be repeatedly evaluated to form the attacks. On the other hand, it is also demonstrated  that poisoning attacks designed against two-stage models can be transferred to fully integrated DFL models. 

Separately, \citeA{yu2023stackel} study robustness of decision-focused learning under label noise. The paper provides bounds on the degradation of regret when test-set labels are corrupted by noise relative to those of the training set. An adversarial training scheme is also proposed to mitigate this effect. The robust training problem is equivalent to finding the equilibrium solution to a Stackelberg game, in which a figurative adversary applies label noise that is optimized to raise regret, while the main player seeks model parameters that minimize regret. 

\subsubsection{Stochastic Optimization} 
Settings in decision-focused learning based on stochastic optimization models are studied by \citeA{DontiKA17}. In contrast to more typical settings, the downstream decision model is considered to be a stochastic optimization problem. In this formulation, it is only possible to predict parameters of a random distribution that models the parameters of an optimization problem. For instance, the mean and variance of load demands in a power scheduling problem could be modeled as parameters of the optimization problem. Their work shows how such problems can be converted to DFL with deterministic decision models and solved using the techniques described in this survey article.
To this end, it also introduces an effective technique for approximating the derivatives through arbitrary convex optimization problems, by forming and differentiating their quadratic programming approximations, as computed by sequential quadratic programming. 
A more elaborate relation to stochastic optimization is provided by \citeA{sadana2023survey} within the context of contextual stochastic optimization.

\subsubsection{Active Learning Algorithm for DFL} 
Active learning concerns ML problems where labeled data are scarce or expensive to obtain. 
To address the challenge of limited training data, active learning algorithms choose the most informative instances for labeling~
\cite{settles.tr09}. \citeA{liu2023active} study active learning in DFL paradigm. 
To identify the features for which knowing the cost parameter is most beneficial for training,
they propose to use the notion of `distance to degeneracy' \cite{BalghitiEGT19}.
Distance to degeneracy measures how far the predicted cost vector is from the set of cost vectors that have multiple optimal solutions.
They argue that if distance to degeneracy is higher at a datapoint, there is more certainty regarding the solution (of the CO problem); hence they propose to acquire the label of a datapoint if its distance to degeneracy is lower than a threshold. 

We end this section with a remark that DFL can be extended to address optimization problems beyond the `classical' CO problem, as defined in \eqref{eq:opt_generic}.
For instance, \citeA{WilderEDT19} embed $K$-means clustering as a layer in a neural network by differentiating through the clustering layer, using DFL techniques. \citeA{wang2021learning} employ a DFL approach to predict the parameters of Markov decision processes (MDPs).

\section{Applications of Decision-Focused Learning}\label{sect:applications}
The \pto problem occurs in many real-world applications, as optimal decisions can be found by solving CO problems and due to the presence of uncertainty, some parameters of the CO problems must be estimated.
Having seen the development of DFL for \pto problems in the preceding section,
practical uses of DFL in various application domains will be presented below.
As this survey paper reviews DFL techniques, which predict cost parameters in Section \ref{sect:review}, 
 the applications presented below are related specifically to the task of predicting only the cost parameters.
\paragraph{Computer vision.}The DBB framework \cite{PogancicPMMR20} (reviewed in Section \ref{sect:perturb}) has been employed in many interesting computer vision applications.
\citeA{RolinekMPPMM20} use for differentiating rank-based metrics such as precision and recall, whereas \citeA{RolinekSZPMM20} and \citeA{miccai/KainmuellerJRM14} use it for differentiating bipartite matching in deep graph and multi-graph matching problems respectively in the application of semantic keypoint matching of images.
\citeA{NEURIPS2021_83a368f5} use DBB to differentiate through a multiway cut combinatorial optimization problem for graph partitioning allowing them to perform end-to-end panoptic segmentation of images.

\paragraph{Fair Learning to Rank.} In learning to rank (LTR), a machine learning model must produce rankings of documents in response to users' search queries, in which those most relevant to a given query are placed in the highest ranking positions. In this setting, the relevance of documents to queries is often measured empirically by historical user click rates \cite{cao2007learning}.  In fair learning to rank (FLTR), this relevance-based matching must be performed subject to strict constraints on the relative exposure between predefined groups. Due to the difficulty of enforcing such constraints on the outputs of a machine learning model, many FLTR frameworks resort to a two-stage approach in which prediction of query-document relevance scores is learned by a typical LTR model without constraints on fairness of exposure. At test time, the predicted relevance scores inform the objective of a separate fair ranking optimization program \cite{singh2018fairness}. \citeA{kotary2022end} use DFL to unify the prediction of relevance scores with the subsequent optimization of fair rankings, in an end-to-end model trained by SPO. This DFL approach to FLTR requires significant computation to solve a linear program for fair rankings throughout training; it was subsequently proposed in \citeA{dinh2024learning} to employ a more efficient optimization of ordered weighted averages following \citeA{do2022optimizing}, adapted to leverage the SPO training framework. 

\paragraph{Route optimization.}\citeA{ferber2023predicting} present an interesting application, where DFL is used to combat the challenge of wildlife trafficking. They consider the problem of predicting the flight trajectory of traffickers based on a given pair of source and destination airports. It is framed as a shortest path problem in a graph, where each node is an airport.
In the prediction stage, the probability of using a directed edge $(i,j)$ to leave the node $i$ is predicted. In the optimization stage, the most likely path from the source to the destination is found by solving a shortest path problem where the negative log probabilities are used as edge weights. 
In this \pto formulation, the probabilities are predicted via DFL, using the DBB framework for gradient computation.

Solving a shortest path problem by considering the negative log probabilities as edge weights has also been explored by \citeA{CP.2021.42}.
In their work, the objective is to prescribe most preferred routing in a capacitated vehicle routing problem (CVRP) \cite{toth2014vehicle} for last-mile delivery applications.
A high probability value for the edge $(i,j)$ indicates that it is the preferred edge to leave the node $i$. However, they do not observe any advantage of the DFL paradigm over the PFL paradigm and attribute this to the lack of training data instances (fewer than $200$ instances).
DFL is used for last-mile delivery applications by \citeA{chu2021data} too.
However, there the objective is to minimize total travel time.
In the prediction stage, the travel times of all the edges are predicted and in the optimization stage, the CVRP is solved to minimize the total travel time. 
The underlying model is trained using the SPO framework to directly minimize the total travel time.

\paragraph{Maritime transportation.} The inspection of ships by port state control has been framed as a \pto problem by \citeA{yang2022pairwise}.
Due to limited number of available personnel, the objective is to identify non-compliant ships that are more likely to be detained, and then select those ships for inspection.
A ship can be found to be non-compliant by port state control in multiple categories. If a ship is found to be non-compliant for a category, 
a `deficiency number' will be recorded for the ship in that category. 
In the prediction stage, a linear model is built to identify deficiency numbers of the ships in all the categories and in the optimization stage, a CO problem is solved to select ships maximizing the total `deficiency number'. 
Given the scale of the CO problem at hand, training with standard SPO framework becomes prohibitive.
Therefore,
they employ pairwise-comparison based loss function, similar to Eq.~\eqref{eq:pairwise} to implement DFL.
Ship maintenance activities by ship owners have been framed as \pto problems by \citeA{TIAN202332}. 
The ship owners have to schedule regular maintenance activities to remain compliant. However, as maintenance activities are expensive, the objective of identifying categories that may warrant immediate detentions by the port state control, has been considered. To do so, in the prediction stage,
a random forest model is built to predict the deficiency number (likelihood of non-compliance) for each category. 
In the optimization stage, a CO problem is formulated considering maintenance cost and detention cost to determine whether maintenance activity should be scheduled for each category. The random forest models are trained to directly minimize regret using SPOTs~\cite{elmachtoub2020decisiontree}.

\paragraph{Power and energy systems.}
\citeA{WAHDANY2023109384} provide a use-case of DFL in renewable power system application. In their work, the prediction stage involves the task of generating wind power forecasts. As these forecasts are further used in power system energy scheduling, 
the predictive ML model is directly trained with the objective of minimizing power system operating costs using
\textsl{cvxpylayers} \cite{agrawal2019differentiable}.
\citeA{TschoraPriceForecasting} observe that electricity prices are set by regulators by optimizing social welfare while maintaining a constant energy balance. They formulate this optimization as a Mixed-Integer Quadratic Program (MIQP) problem and embed the CO problem into the training loop of an electricity price forecasting framework,
which outperforms the conventional price forecasting approach. By analytically differentiating the CO problem, they derive a step function as the derivative, which is used for training the ML model in DFL paradigm.
%
%
\citeA{sang2022electricity} consider another \pto problem in power system applications, where electricity prices are predicted in the prediction stage and the optimization stage deals with optimal energy storage system scheduling to maximize arbitrage benefits.
Lower values of regret have been reported in their work, when the prices are predicted using the SPO framework.

\paragraph{Communication technology.} 
DFL is applied to mobile wireless communication technology applications by ~\citeA{chai2022port}.
Fluid antenna systems \cite{wong2020fluid} are one of the recent developments in mobile wireless communication technology. However, its effectiveness depends on the position of the radiating element, known as the port. 
\citeA{chai2022port} frame the port selection problem as a \pto problem, where in the prediction stage signal-to-noise ratio for each position of the port is predicted and then the optimal position of the port is decided in the optimization stage. They use LSTM as the predictive model and report the SPO framework is very effective for such port selection applications.
\paragraph{Solving non-linear combinatorial optimization problems.}
\citeA{ferber2022surco} study the problem of learning a linear surrogate optimizer to solve non-linear optimization problems. The objective is to learn a surrogate linear optimizer whose optimal solution is the same as the solution to the non-linear optimization problem. Learning the parameters of the surrogate linear optimizer entails backpropagating through the optimizer, which is implemented using \textsl{cvxpylayers} \cite{agrawal2019differentiable}.

Further, interested readers are referred to the artcle by \citeA{MisicINFORMS} for more applications of \pto problems in various areas within OR.
\section{Experimental Evaluation on Benchmark Problemsets}\label{sect:data}
DFL recently has received increasing attention. The methodologies discussed in Section \ref{sect:review} have been tested so far on several different datasets.  Because a common benchmark for the field has not yet been set up, comparisons among techniques are sometimes inconsistent. In this section, an effort is made to propose several benchmark test problems for evaluating DFL techniques.
Then some of the techniques explained in Section \ref{sect:review} are compared on these test problems. 
\begin{table}[]
    \centering
    \resizebox{\linewidth}{!}{
    \begin{tabular}{lccccc}
    \rowcolor{black} 
    \color{white} 
    Problem & \color{white} Constraint Functions & \color{white} Decision Variables & \color{white}  CO SOlver &  \color{white} Predictive Model \\
     \toprule
\makecell[l]{Shortest path problem \\on a $5 \times 5$ grid} & Linear & Continuous & \textsl{OR-Tools} & Linear\\
Portfolio optimization   & Quadratic& Continuous & \textsl{Gurobi} & Linear\\
Warcraft shortest path  & Linear & Continuous & \makecell{Customized python \\implementation of Dijkstra} & CNN\\
\makecell[l]{Energy-cost aware\\ scheduling}  & Linear & Discrete  & \textsl{Gurobi} & Linear\\
Knapsack problem & Linear & Discrete & \textsl{OR-Tools}  & Linear\\
Diverse bipartite matching  & Linear & Discrete  & \textsl{OR-Tools} &  Multi-layer NN \\
Subset selections  & Linear & Continuous & \textsl{Gurobi} & Linear \\
     \bottomrule
    \end{tabular}
    }
    \caption{Brief overview of the test problems considered for experimental evaluation. The \textbf{objective functions are linear} for all the optimization problem.}
    \label{Tab:ProblemDescri}
\end{table}

\subsection{Problem Descriptions}
All the test problems, which are selected for benchmarking, have been previously used in the DFL literature and their datasets are publicly available.
All these problems are \pto problems, meaning they encompass the two stages---prediction and optimization. 
Moreover, all the optimization problems have linear objective functions.
Table \ref{Tab:ProblemDescri} provides an overview of the experimental setups associated with each test problem, including the specification of the CO problem and the type of predictive model.
Next, these test problems are described in detail.
\subsubsection{Shortest Path Problem on a $5 \times 5$ grid}
This experiment is adopted from the work of \citeA{elmachtoub2022smart}. It is a shortest path problem on a $5 \times 5$ grid, with the objective of going from the southwest corner of the grid to the northeast corner where the edges can go either north or east. This grid consists of $25$ nodes and $40$ edges. 
\paragraph{Formulation of the optimization problem.}
The shortest path problem on a graph with a set  $V$ of vertices and a set $E$ of edges can be formulated as an LP problem in the following form:
\begin{subequations}
    \label{eq:shortestpath}
\begin{align}
    \min_{\mathbf{x}} & \;\ \mathbf{c}^\top \mathbf{x} \\
    \label{eq:shortestpath.eq_constraint}
   \texttt{s.t.} & A \mathbf{x} =\mathbf{ b} \\
   & \mathbf{x } \geq \mathbf{0}
\end{align}
\end{subequations}
Where $A \in  \mathbb{R}^{|V| \times |E|}$ is the incidence matrix of the graph.
The decision variable $\mathbf{x}\in  \mathbb{R}^{|E|}$ is a binary vector whose entries are 1 only if the corresponding edge is selected for traversal. $\mathbf{b} \in  \mathbb{R}^{ |V| }$ is the vector whose entry corresponding to the source and sink nodes are 1 and $-1$ respectively; all other entries are 0. The constraint \eqref{eq:shortestpath.eq_constraint} must be satisfied to ensure the path will go from the source to the sink node.
The objective is to minimize the cost of the path with respect to the (predicted) cost vector $\mathbf{c} \in \mathbb{R}^{|E|}$.

\paragraph{Synthetic data generation process.}
In this problem, the prediction task is to predict the cost vector $\mathbf{c}$ from the feature vector $\mathbf{z}$.
The feature and cost vectors are generated according to the data generation process defined by \citeA{elmachtoub2022smart}. 
For the sake of completeness,
the data generation process is described below.\footnote{The generator in \url{https://github.com/paulgrigas/SmartPredictThenOptimize} is used to generate the dataset.}
Each problem instance has cost vector of dimension $|E| = 40$ and feature vector of dimension $p=5$.
The training data consists of $\{ (\mathbf{z_i},  \mathbf{c_i})\}_{i=1}^N$, which are generated synthetically.
The feature vectors are sampled from a multivariate Gaussian distribution with zero mean and unit variance, i.e., $\mathbf{z_i} \sim \mathbf{N} (0, I_p)$
To generate the cost vector, first a matrix $B \in \mathbb{R}^{|E| \times p} $ is generated, which represents the true underlying model.
The cost vectors are then generated according to the following formula:
\begin{equation}
    c_{ij} = \bigg [ \bigg(\frac{1}{\sqrt{p}} \big(B \mathbf{z_i}  \big) +3  \bigg)^{\text{Deg}  } +1 \bigg]\xi_i^j
\end{equation}
where $c_{ij}$ is the $j^\text{th}$ component of cost vector $\mathbf{c_i}$. 
The \emph{Deg} parameter specifies the extent of model misspecification, because a linear model is used as a predictive model in the experiment. The higher the value of \emph{Deg}, the more the true relation between the features and objective parameters deviates from a linear one and the larger the prediction errors will be. Finally, $\xi_i^j$ is a multiplicative noise term sampled randomly from the uniform distribution $[1- \vartheta, 1+ \vartheta ]$. 
The experimental evaluation involves five values of the parameter \emph{Deg}, which are $ 1, 2, 4, 6 \text{ and } 8$, and  the noise-halfwidth parameter $\vartheta$ being  $0.5$. 
Furthermore, for each setting,
a different training set of of size 1000 is used.
In each case, the final performance of the model is evaluated on a test set of size $10,000$. 
\paragraph{Predictive model.}
In each setting, the underlying predictive model is a one-layer feedforward neural network without any hidden layer, i.e., a linear model.
The input to the model is a $p$ dimensional vector, and output is a $|E|$ dimensional vector.
Note that a multi-layer neural network model can be used to improve the accuracy of the predictive model. The intuition behind using a simple predictive model is to test the efficacy of the DFL techniques when the predictions are not 100\% accurate. 
The DFL techniques are trained to minimize regret, and the prediction-focused model is trained by minimizing the MSE loss between the true and predicted cost vectors.

\subsubsection{Portfolio Optimization Problem}
A classic problem that combines prediction and optimization is the Markowitz portfolio optimization problem, in which asset prices are predicted by a model based on empirical data, and then subsequently, a risk-constrained optimization problem is solved for a portfolio which maximizes expected return. This experiment is also adopted from the work of \citeA{elmachtoub2022smart}.
\paragraph{Formulation of the optimization problem.}
In portfolio optimization problem, the objective is to choose a portfolio of assets having highest return subject to a constraint
on the total risk of the portfolio.
The problem is formulated in the following form:
\begin{subequations}
\label{eq:portfolio}
\begin{align}
    \label{eq:portfolio_obj}
    \max_{\mathbf{x}} &\;\;
    \mathbf{c}^\top \mathbf{x}\\
    \label{eq:portfolio_ineq}
    \texttt{s.t.}\; \;\; &
     \mathbf{x}^\top \Sigma \mathbf{x} \leq \gamma \\
    \label{eq:portfolio_eq}
    & \mathbf{1} ^\top  \mathbf{x} \leq 1 \\
    \label{eq:portfolio_bound}
    &\;\;\;  \mathbf{x} \geq 0
\end{align}
\end{subequations}
\noindent
where $\mathbf{1}$ is the vector of all-ones of same dimension as $\mathbf{x}$, $\mathbf{ c}$ is the vector of asset prices, and $\Sigma$ is a predetermined matrix of covariances between asset returns. The objective  (\ref{eq:portfolio_obj}) is to maximize the portfolio's total value. Eq. \eqref{eq:portfolio_ineq} is a risk constraint, which bounds the overall variance of the portfolio, and \eqref{eq:portfolio_eq}, \eqref{eq:portfolio_bound} model $\mathbf{x}$ as a vector of proportional allocations among assets. 

\paragraph{Synthetic data generation process.}
Synthetic input-target pairs $(\mathbf{z}, \mathbf{c})$ are randomly generated, according to a random function with a specified degree of nonlinearity $\text{Deg} \in \mathbb{N}$. The procedure for generating the random data as follows: 

Given a number of assets $d$ and input features of size $p$, input samples $\mathbf{x_i} \in \mathbb{R}^p$ are sampled element wise from i.i.d. standard normal distributions $\mathbf{N}(0,1)$. A random matrix $B \in \mathbb{R}^{d \times p}$, whose elements $B_{ij} \in \{ 0,1 \}$ are drawn from i.i.d. Bernoulli distributions which take the value $1$ with probability $0.5$, is created. For a chosen noise magnitude $\vartheta$, $L \in \mathbb{R}^{n \times 4}$ whose entries are drawn uniformly over $[-0.0025\vartheta, 0.0025\vartheta]$ is generated. Asset returns are calculated first in terms of their conditional mean $\bar{c}_{ij}$ as 

\begin{equation}
    \label{eq:portfolio_conditional_mean}
    \bar{c}_{ij} \coloneqq ( \frac{0.05}{\sqrt{p}}(B \mathbf{z_i})_j + (0.1)^{\frac{1}{\text{Deg}}} )^{\text{Deg}}
\end{equation}
\noindent
Then the observed return vectors $\mathbf{c_{i}}$ are defined as $c_{ij} \coloneqq \bar{r}_i + Lf + 0.01  \vartheta \xi$, where $f \sim \mathbf{N}(0,I_4) $ and noise $\xi  \sim \mathbf{N}(0,I_d)$
This causes the $c_{ij}$ to obey the covariance matrix $\Sigma \coloneqq L L^\top + (0.01 \zeta)^2 \mathbf{I}$, which is also used to form the constraint (\ref{eq:portfolio_ineq}), along with a bound on risk, defined as $\gamma \coloneqq 2.25 \; e^\top \Sigma e$ where $e$ is the equal-allocation solution (a constant vector).
Four values of the parameter \emph{Deg}---$ 1,  4, 8, 16$ have been used in the experimal evaluation.
The value of noise magnitude parameter $\vartheta$ is set to  $1$. 
It is assumed that the covariance matrix of the asset returns does not depend on the features.
The values of $\Sigma$ and $\gamma$ are constant, and randomly generated for each setting. 

\paragraph{Predictive model.}
Like the previous experiment, 
the underlying predictive model is a linear model,
whose input is a feature vector $\mathbf{z} \in \mathbb{R}^{p} $ and output is the return vector $\mathbf{c} \in  \mathbb{R}^{d}$.






\subsubsection{Warcraft Shortest Path Problem}
This experiment was adopted from the work of \citeA{PogancicPMMR20}.
Each instance in this problem is an image of a terrain map using the Warcraft II tileset \cite{warcraft}. 
Each image represents a grid of dimension $d \times d$.
Each of the $d^2$ pixels has a fixed underlying cost, which is unknown and to be predicted.
The objective is to identify the minimum cost path from the top-left pixel to the bottom-right pixel.
From one pixel, one can go in eight neighboring pixels---up, down, front, back, as well as four diagonal ones.
Hence, it is a shortest path problem on a graph with $d^2$ vertices and $\mathcal{O} (d^2)$ edges.
\paragraph{Formulation of the optimization problem.}
Note that this is a node-weighted shortest path problem, where each node (pixel) in the grid is assigned a cost value; whereas in the previous shortest path problem, each edge is assigned a cost value.
However, this problem can be easily reduced to the more familiar edge weighted shortest path problem by `node splitting'. 
`Node splitting' splits each node into two separate nodes---entry and exit nodes and adds an edge, that has a weight equal to the node weight, from the entry node to the exit node.
For each original edge, an edge, with null weight, from the exit node of the source node to the entry node of the sink node, is constructed. 
\paragraph{Predictive model.}
The prediction task is to predict the cost associated with each pixel. 
The actual cost ranges from 0.8 to 9.2 and is dependent on visible characteristics of the pixel.
For instance, cost changes depending on whether the pixel represents a water-body, land or wood.
To predicts the cost of each node (pixel),
a convolutional neural network (CNN) is employed.
The CNN takes the $d \times d$ image as an input and outputs costs of the $d^2$ pixels.
The ResNet18~\cite{Resnet_CVPR} architecture is slightly modified to form the ML model.
 The first five layers of ResNet18 are followed by a
max-pooling operation to predict the underlying cost of each pixel.
Furthermore, a Relu activation function \cite{relu2019deep} is used to ensure the predicted cost remains positive, thereby avoiding the existence of negative cycles in the shortest path edge weights.
\raggedbottom
 
\subsubsection{Energy-Cost Aware Scheduling Problem}
This experiment setup was adopted from the work of \citeA{mandi2020smart}.
This is a resource-constrained day-ahead job scheduling problem \cite{csplib:prob059} with the objective of minimizing energy cost. Tasks must be assigned to a given number of machines, where each task has a duration, an earliest start, a latest end, a resource requirement and a power usage. Each machine has a resource capacity constraint. 
Also, tasks cannot be interrupted once started, nor migrated to another machine and must be completed before midnight. 
The scheduling is done one day in advance. So, the prediction task is to predict the energy prices of the next day.
\paragraph{Formulation of the optimization problem.}
The scheduling problem is formulated as an ILP.
Let $J$ be the set of tasks to be scheduled on a set of machines $I$ while maintaining resource requirement of $W$ resources. The tasks must be scheduled over $T $ number of time slots.
Each task $j$ is specified by its duration $\zeta_j$, earliest start time $\zeta_j^{(1)}$, latest end time $\zeta_j^{(2)}$, power usage $\phi_j$. 
Let $\rho_{jw}$ be the resource usage of task $j$ for resource $w$ and $q_{iw}$ is the capacity of machine $i$ for resource $w$.
Let $x_{jit}$ be a binary variable that takes the value 1 only if task $j$ starts at time $t$ on machine $i$.
The objective of minimizing energy cost while satisfying the required constraints can be expressed by the following ILP:
\begin{subequations}
\begin{align}
\label{eq:energy_obj}
     \min_{x_{jit}}  \sum_{j \in J} \sum_{i \in I} \sum_{t \in T} & x_{jit} \Big( \sum_{t \leq t\prime < t+ \zeta_j} \phi_j c_{t\prime} \Big) \\
 \label{eq:energy_onlyonceconstraint}
     \texttt{s.t.}\; \;\;  \sum_{i \in I} \sum_{t \in T} x_{jit}  &= 1 \ , \forall_{j \in J} \\
  \label{eq:energy_earlystart}
      x_{jit}   &= 0 \ \ \forall_{j \in J} \forall_{i \in I} \forall_{t < \zeta_j^{(1)}} \\
  \label{eq:energy_latestend}
      x_{jit}  &= 0 \ \ \forall_{j \in J} \forall_{i \in I} \forall_{t + \zeta_j > \zeta_j^{(2)}} \\
  \label{eq:energy_resourceconstraint}
      \sum_{j \in J} \sum_{t - \zeta_{j}  < t^\prime \leq t} x_{jit\prime } \rho_{jw} & \leq q_{iw},  \forall_{i \in I} \forall_{w \in W} \forall_{t \in T}\\
      x_{jit} \in \{0,1\}  & \forall_{j \in J} \forall_{i \in I} \forall_{t \in T}
 \end{align}
 \end{subequations}
\noindent
The \eqref{eq:energy_onlyonceconstraint} constraint ensures each task is scheduled once and only once.
The constraints in \eqref{eq:energy_earlystart} and \eqref{eq:energy_latestend} ensure that the task scheduling abides by earliest start time and latest end time constraints.
\eqref{eq:energy_resourceconstraint} imposes the constraints of resource requirement.

\paragraph{Data description.}
The prediction task is to predict the energy prices one day in advance.
The energy price dataset comes from the Irish Single Electricity Market Operator (SEMO) \cite{ifrim2012properties}. 
This dataset consists of historical energy price data at 30-minute intervals starting from midnight on the 1st of November, 2011 until the 31st of December, 2013.
In this setup, each day forms an optimization instance, which comprises of $48$ time slots, corresponding to $48$ half-hour slots.
Each half-hour instance of the data has calendar attributes, day-ahead estimates of weather characteristics, SEMO day-ahead forecasted energy-load, wind-energy production and prices, actual wind-speed, temperature and $CO_2$ intensity, which are used as features.
So, the dimension of feature vector is 8.
Note that, in this dataset, each $c_t$ in the cost vector is associated with an eight dimensional feature vector, i.e., $\mathbf{c} \in \mathbb{R}^{48}$ 
and $\mathbf{z} \in \mathbb{R}^{48\times 8}$.

\paragraph{Predictive model.}
As energy prices of each half-hour slot is associated with $8$ features,
the input to the predictive model is a feature vector of dimension $8$ and output is a scalar. In this case also, 
the predictive model is a linear model, i.e., a feed forward neural network without any hidden layer.
\subsubsection{Knapsack Problem}
This problem setup was also adopted from the work of \citeA{mandi2020smart}. 
The objective of the knapsack problem is to choose a maximal value subset from a given set of items, subject to a capacity constraint.
In this case, the weights of all items and the knapsack capacity are known. What is unknown are the values of each item.
Hence, the prediction task is to predict the value of each item.
\paragraph{Formulation of the optimization problem.}
The formulation of the knapsack optimization problem with unit weights has already been provided in Eq.~\eqref{eq:unitknapsackformulation}.
However, in general the weights of all items are not equal. 
So, a general knapsack optimization problem can be formulated as follows:
\begin{subequations}
    \label{eq:knapsackformulation}
    \begin{align}
   \max_{\mathbf{x} } \;\;& \mathbf{c}^\top \mathbf{x} \\ 
   \texttt{s.t.}\; & \mathbf{w}^\top \mathbf{x} \leq \text{Capacity}\\
   & \mathbf{x} \in \{0,1\}
   \end{align}
\end{subequations}
\noindent where $\mathbf{w}$, $\mathbf{c}$ are the vector of weights and values respectively.
\paragraph{Data description.}
For this problem again, the dataset is adapted from the Irish Single Electricity Market Operator (SEMO) \cite{ifrim2012properties}. 
In this setup, each day forms an optimization instance and each half-hour corresponds to a knapsack item.
So the cost vector $\mathbf{c}$ and the weights $\mathbf{w}$ are of length 48, corresponding to 48 half-hours.
The motivation can be framed as identifying half-hour slots that yield maximum revenue while adhering to the constraint of booking each slot.
Similar to the energy scheduling problem, each item of the cost vector is associated with a feature vector of dimension 8.
The weight vector is fixed.
The weights are generated synthetically as done by \citeA{mandi2020smart}. The data generation is as follows.
First a weight $w_i$ is assigned to each of the 48 half-hour slots, by sampling from the set $\{3,5,7\} $.
In order to introduce correlation between the item weights and the item values,
the energy price vector is multiplied with the weight vector and then a  
randomness is incorporated by adding Gaussian noise $ \xi \sim \mathbf{N}(0,25)$, which produces the final item values $c_i$.
The motivation behind introducing correlation between the item weights and the item values stems from the fact that solving a knapsack problem with correlated item weights and values is considered to be hard to solve \cite{PISINGER20052271}.
The sum of the weights of each instance is $240$. 
60, 120, and 180 are the three values of capacity with which the experiments are performed.
\paragraph{Predictive model.}
The predictive model is same as the previous problem, i.e., a feed forward neural network without any hidden layer.

\subsubsection{Diverse Bipartite Matching Problem}
This experimental setup is adopted from \citeA{aaaiFerberWDT20}.
In this problem, two disjoint sets of nodes are provided and the objective is to match between the nodes of the two sets. 
The graph topologies are taken from the CORA citation network \cite{cora2008}, where a node represent an published article and an edge represent a citation.
So the goal of the matching problem is to identify the citations between the two sets of articles. 
Furthermore, the matching must obey diversity constraints, as described later.

\paragraph*{Optimization problem formulation.}
Let $S_1$ and $S_2$ denote the two sets.
The matching must satisfy the following diversity constraints: 
a minimum $\rho_1 \% $ and $\rho_2 \%$ of the suggested pairings should belong to same and distinct fields of study respectively. 
In this matching problem, each edge does not have an associated cost in the true sense.
The DFL approaches consider the likelihood of the existence of each edge as the edge weights and then determine which edges should be present while ensuring all the constraints are satisfied.
Let $c_{ij}$ be the likelihood of an edge existing between article $i$ and $j$, $\forall i \in S_1, j \in S_2$. 
With this likelihood value, the matching can be performed by solving the following ILP, which ensures the diversity constraints:
\begin{subequations}
     \begin{align}
          \max_{\mathbf{x}} & \sum_{i, j} c_{i j} x_{i j} & \\
          \texttt{s.t.}\; \;\; & \sum_{j} x_{i j} \leq 1 & \forall i \in S_{1} \\
          & \sum_{i} x_{i j} \leq 1 & \forall j \in S_{2} \\ 
          & \sum_{i , j} \phi_{i, j} x_{i j} \geq \rho_1 \sum_{i, j} x_{i j} & \\
          & \sum_{i , j} (1-\phi_{i j}) x_{i j} \geq \rho_2 \sum_{i, j} x_{i j} \\ 
          & x_{ij} \in \{0,1\} & \forall i \in S_1, j \in S_2 
     \end{align}
\end{subequations}
\noindent
where $\phi_{ij}$ is an indicator, which takes the value $1$ only if article $i$ and $j$ are of same field, and $0$ if they belong to two different fields.
\paragraph{Data description.}
The network is divided into 27 disjoint topologies, each of which forms a distinct optimization problem instance containing 100 nodes. 
In each instance, the $100$ nodes are split into two sets of 50 nodes $S_1$ and $S_2$; so 
each instance forms a bipartite matching problem between two sets of cardinality 50.
Each publication (node) has 1433 bag-of-words features. The feature of an edge is formed by concatenating features of the two corresponding nodes.
The prediction task is to estimate
$c_{ij}$ values.
In this problem, each individual $c_{ij}$ is associated with a feature vector of length 2866.

\paragraph{Predictive model.}
Although the DFL paradigm considers the likelihood of the existence of each edge as the edge weights,
in the prediction-focused approach the model is trained by directly predicting the presence or absence of each edge.
The predictive model is a neural network model. 
The input to the neural network is a 2866 dimensional vector and final output is a scalar between 0 and 1.
The neural network has one hidden layer and uses a sigmoid activation function on the output.

\subsubsection{Subset Selections}
This experiment is a structured prediction task, in which the object is to learn a mapping from feature vectors to binary vectors which represent subset selections. Unlike the other experiments above, the ground-truth data take the form of optimal solutions to an optimization problem, rather than its corresponding problem parameters. Thus, the regret loss is not suitable for training a prediction model. Instead, a task loss based on the error of the predicted solutions with respect to ground-truth solutions is used in this experiment. 

\paragraph{Optimization problem formulation.}

For any $\mathbf{c} \in \mathbb{R}^n$, the objective of the optimization problem is to output a binary vector in $\mathbb{R}^n$, where the non-zero values correspond to the top-$k$ values of $\mathbf{c}$. This can be formulated as an LP problem in the following form:
\begin{subequations}
\label{eq:topk-lp}
\begin{align}
     \argmax_{ \mathbf{x} } &\;\;
    \mathbf{c}^\top \mathbf{x}\\
    \texttt{s.t.} \;\; &
    1^\top \mathbf{x} = k \\
    & 0 \leq \mathbf{x} \leq 1
\end{align}
\end{subequations}
As a totally unimodular linear program with integral parameters, this problem has (binary) integer  optimal solutions. This mapping is known for its ability to represent subset selections in structured prediction, and is useful for multilabel classification \cite{amos2019limited} . 

\paragraph*{Data Description.}

Let $\mathbf{U}(0,1)$ be a uniform random distribution; then a  collection of feature vectors $\mathbf{z}$ are generated by $\mathbf{z} \sim  \mathbf{U}(0,1)^{n}$. For each $\mathbf{z}$, its corresponding target data is a binary vector containing unit values corresponding to the top-$k$ values of $\mathbf{z}$, and zero values elsewhere. Three datasets are generated, each of $1000$ training samples, in which the selection problem takes size $n = 25$, $n = 50$, and $n = 100$ respectively. The subset size $k$ is chosen to be one fifth of $n$, in each case.

\paragraph{Predictive model.}
Like the previous problem, the predictive model here also is a linear model, i.e., a feed forward neural network without any hidden layer.
In this problem, the task loss to train the predicative model is the negated inner product between true selection $\mathbf{ x}$ and prescribed selection $\mathbf{\hat{x} }$, i.e., $\mathcal{L}( \mathbf{\hat{x}}, \mathbf{x}) = \mathbf{\hat{x}}^\top \mathbf{ x}$, which is minimized when $\mathbf{\hat{x} }=\mathbf{ x}$. 
Since the model is not regret-based, and does not assume access to the ground-truth parameters $\mathbf{c}$, techniques which rely on such assumptions are not tested on this problem.

\begin{table}[]
    \centering
    \resizebox{\linewidth}{!}{
    \begin{tabular}{ccc}
    \rowcolor{black} 
    \color{white} 
         Hyperparameter &  \color{white} \makecell{Techniques Utilizing\\ the Hyperparameter} &   \color{white} Range \\
         \toprule
         learning rate & All  & $\{ 5\times 10^{-4}, 1\times 10^{-3}, 5\times 10^{-3}, 0.01, 0.05, 0.1 , 0.5, 1.0 \}$ \\
         $\delta$ & I-MLE, DBB & $\{ 0.1, 1, 10, 100 \}$ \\
         $\epsilon$ & I-MLE, FY & $\{ 0.05, 0.1, 0.5,1,2,5 \}$ \\
         $\kappa$ & I-MLE & $\{ 5,10, 50\}$ \\
           $\tau$ & Listwise & $\{ 0.05, 0.1, 0.5,1,2,5 \}$ \\
         $\Theta$ & Pairwise & $\{ 0.01, 0.05, 0.1, 1., 10., 50.\}$ \\
         $\mu$ & QPTL, HSD & $\{ 0.01, 0.1, 1., 10.,  \}$ \\
         \bottomrule
    \end{tabular}
    }
    \caption{The range of hyperparameters for hyperparameter tuning by grid search.}
    \label{tab:hyperparam_range}
\end{table}

\subsection{Experimental Results and Analysis}
\label{sect:result}
In this subsection, results of comparative evaluations of some of the techniques introduced in Section \ref{sect:review} on the datasets mentioned in Section \ref{sect:data} are presented. 
The following techniques are considered for evaluations:
1. Prediction-focused (PF) approach,
2. Smart ``Predict, Then Optimize'' \eqref{eq:spo} [SPO], 
3. Differentiation of blackbox combinatorial solvers \eqref{eq:dbb} [DBB],
4. Implicit maximum likelihood estimation \eqref{eq:imle} [I-MLE],
5. Fenchel-Young loss \eqref{eq:fy} [FY],
6. Differentiation of homogeneous self-dual  embedding \eqref{eq:intopt_LP} [HSD],
7. Quadratic programming task loss \eqref{eq:qptl} [QPTL],
8. Listwise LTR loss \eqref{eq:listwise} [Listwise],
9. Pairwise LTR loss \eqref{eq:pairwise} [Pairwise],
10. Pairwise difference LTR loss \eqref{eq:pairwise_diff} [Pairwise(diff)],
11. Maximum a posteriori contrastive loss \eqref{eq:MAP_lin_obj} [MAP]. 
The reason behind including prediction-focused approach is that it is considered as a benchmark.
Note that among these techniques, Listwise, Pairwise, Pairwise(diff), and MAP make use of a solution cache. 
The solution cache is implemented 
using the procedure proposed by \citeA{mulamba2020discrete}.
In this approach, the solution cache is initialized by caching all the solutions in the training data and the cache is later expanded by employing a $p_{solve}$ parameter value greater than zero.
As in \cite{mulamba2020discrete,mandi22-pmlr} it is reported that $p_{solve} = 5\%$ is adequate for most applications, the value of $p_{solve}$ is set to $5\%$.
Next, the procedure systematically followed for the empirical evaluations is explained.
\paragraph{Experimental setup and procedures.} 
The performance of a technique is sensitive to the choice of the technique specific hyperparameters as well as some other fundamental hyperparameters, common in any neural network training such as learning rate.
These are called hyperparameters because they cannot be estimated by training the model, rather they must be selected before training begins. 
Tuning hyperparameters is the process of identifying the set of hyperparameter values that are expected to produce the best model outcome.
In the experimental evaluations, hyperparameter tuning is performed via grid search. 
In the grid search, each of the hyperparameters is tried for a set of values. The set of values to be tested for each hyperparameter is predetermined. 
Grid search suffers from the curse of dimensionality in the hyperparameter space, as the number of combinations grows exponentially with the number of hyperparameters. However, it is possible to train the different models for different combinations of hyperparameters in parallel as the combinations are independent. 

The hyperparameter of each model for each experiment is selected based on performance on the validation dataset. For each hyperparameter a range of values as defined in Table \ref{tab:hyperparam_range} is considered. The hyperparameter combination which produces the lowest average regret on the validation dataset is considered to be the `optimal' one. 
 For both validation and testing, $10$ trials are run where in every trial the network weights are initialized with a different seed. 
 To be specific,  values of seed from 0 to 9 have been considered.
 Each model for each setup is trained using \textsl{pytorch} \cite{paszke2019pytorch} and \textsl{pytorch-lightning} \cite{falcon2019pytorch} with the \textsl{adam optimizer} \cite{Adamarxiv.1412.6980} and \textsl{`reduceLROnPlateau'}\cite{ReduceLROnPlateau} learning rate scheduler.
 As mentioned before, the learning rate of \textsl{adam optimizer} is treated as a hyperparameter.
For QPTL, the QP problems are solved using \textsl{cvxpylayers} \cite{agrawal2019differentiable}. 
For other techniques, which treat the CO solver as a blackbox solver,
\textsl{gurobi} \cite{gurobi} or \textsl{ortools} \cite{ortools} is used as the solver. 
\paragraph{Evaluation metric.}
After selecting the `optimal' hyperparameter combination for each test problem, \textbf{10} trials of all the techniques with the `optimal' hyperparameter combination are run on test dataset.
Unless otherwise mentioned the comparative evaluation is made based on relative regret on the test dataset. 
The \textbf{relative regret} is defined as follows:
\begin{equation}
\label{eq:relative_regret}
\frac{1}{N_{test}} \sum_{i=1}^{N_{test}} \frac{\mathbf{c_i}^\top ( \mathbf{x}^\star ( \mathbf{\hat{c_i}}) - \mathbf{x}^\star (\mathbf{c_i}))} { \mathbf{c_i}^\top \mathbf{x}^\star ( \mathbf{c_i})}.
\end{equation}

\noindent In practice, $\mathbf{c}$ (or $\mathbf{ \hat{c} }$) can have non-unique optimal solutions.
However, note that if all the entries in $\mathbf{c}$ are continuous, it is very unlikely that $\mathbf{c}$ will have non-unique solutions. For instance, in the case of an LP,
the only circumstance in which the LP can have multiple solutions is
when $\mathbf{c}$ is parallel to one of the faces of the LP polyhedron.
Nevertheless, if the cost vector is predicted by an ML model, a pathological case might occur, especially at the beginning of model training, when all the cost parameters are zero. This results in all feasible solutions being optimal with zero cost.
However, to avoid this complexity in the experiments, it is assumed that the solution $\mathbf{x}^\star(\mathbf{\hat{c}})$ is obtained by calling
a CO solver as an oracle and that if there exist non-unique solutions, the oracle returns a single optimal solution by breaking ties in a pre-specified manner.
This is true if a commercial solver such as  Gurobi is used to solve the CO problem.

\subsubsection{Comparative Evaluations}
\begin{figure}
    \centering
    \begin{subfigure}[b]{\textwidth}
    \centering
    \includegraphics[width=\textwidth]{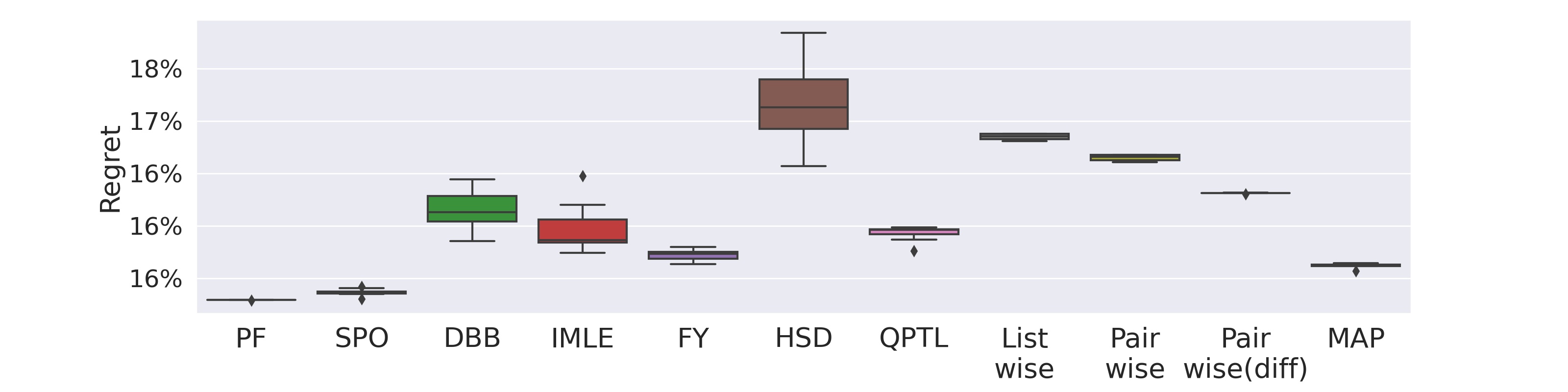}
    \caption{$\text{Deg}=1$}
    \label{fig:shortest_path_noise0.5_deg1}
  \end{subfigure}
  \begin{subfigure}[b]{\textwidth}
    \centering
    \includegraphics[width=\textwidth]{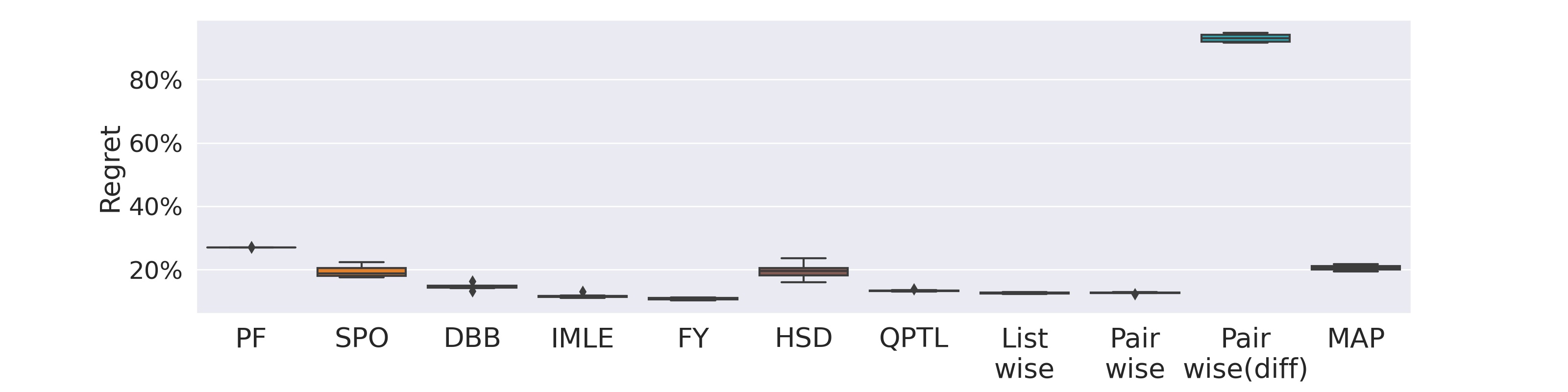}
    \caption{$\text{Deg}=8$}
    \label{fig:shortest_path_noise0.5_deg8}
  \end{subfigure}
    \caption{Comparative evaluations on the synthetic shortest path problem with noise-halfwidth parameter $\vartheta$ = 0.5. The boxplots show the distributions of relative regrets.}
    \label{fig:shortest_path_noise0.5}
\end{figure}
Next, the performances of the 11 DFL techniques across the 7 problems are presented for comparative evaluation.
\paragraph{Shortest path problem on a $5 \times 5$ grid.}
The comparative evaluation for the synthetic 
shortest path problem is shown in Figure~\ref{fig:shortest_path_noise0.5}  with the aid of box plots. 
To conserve space, boxplots for two values of Deg are shown in Figure~\ref{fig:shortest_path_noise0.5}. The boxplots for all the five degrees are shown in Figure~\ref{fig:shortest_path_noise0.5_append} in the Appendix.
In Figure \ref{fig:shortest_path_noise0.5}, the value of  $\vartheta$, the noise-halfwidth parameter is 0.5 for all the experiments and the training set for each Deg contains 1000 instances. 
The predictive model is a simple linear model implemented as a neural network model with no hidden layers. 

For Deg 1, the linear predictive model perfectly captures the data generation process. Consequently, the PF approach is very accurate and it results in the lowest regret. SPO has slightly higher regret than the PF approach. All the other DFL techniques have considerably higher regrets compared to PF.
In contrast, the relative regret worsens for the PF approach, as the value of Deg parameter is increased.
For Deg 8, PF is not the best as the predictive model is misspecified.
In this case, FY has the lowest regret; while  DFL techniques, other than Pairwise(diff) results in lower regret than the PF approach. 
For Deg 6, the best one FY, although its test regret is not very different from SPO, I-MLE and QPTL.
For Deg 4, I-MLE has the lowest regret, closely followed by FY and SPO. 
For Deg 2, both PF and SPO  have lowest regret.

Among the DFL techniques, HSD exhibits higher regret and higher variances than the other DFL approaches. 
It performs better than the PF approach only for Deg 6 and 8.

\begin{figure}
    \centering
    \begin{subfigure}[b]{\textwidth}
    \centering
    \includegraphics[width=\textwidth]{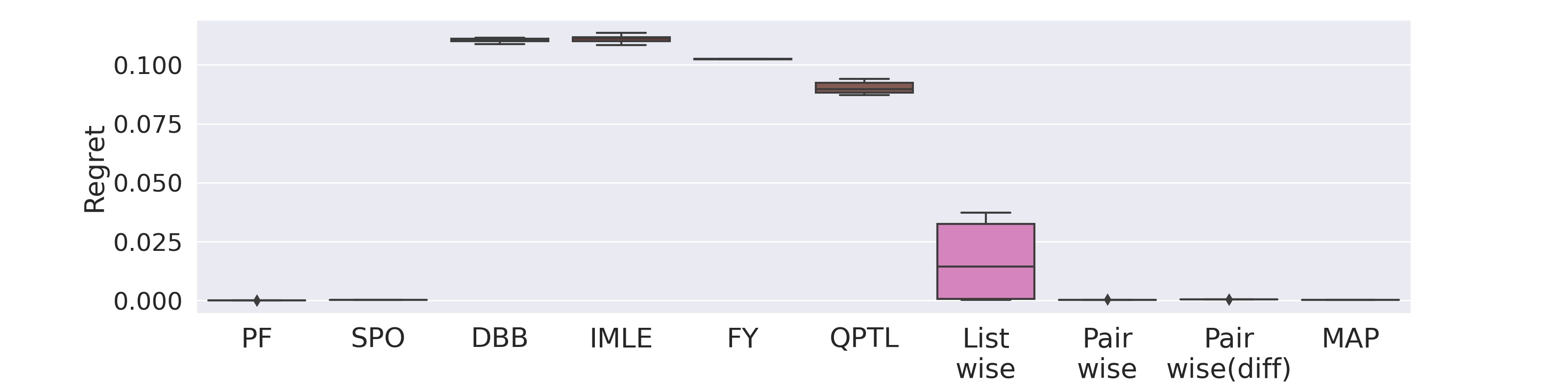}
    \caption{$\text{Deg}=1$}
    \label{fig:portfolio_noise1_deg1}
  \end{subfigure}
  \begin{subfigure}[b]{\textwidth}
    \centering
    \includegraphics[width=\textwidth]{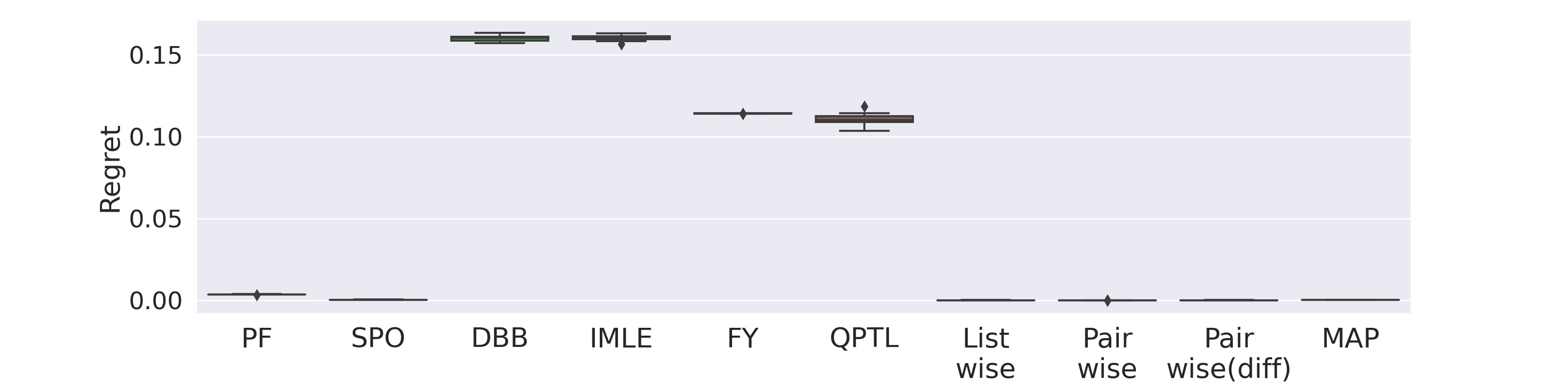}
    \caption{$\text{Deg}=16$}
    \label{fig:portfolio_noise1_deg16}
  \end{subfigure}
    \caption{Comparative evaluations on the synthetic portfolio optimization problem with noise magnitude $\vartheta = 1$. The boxplots show the distributions of \textbf{absolute} regrets.}
    \label{fig:portfolio_noise1}
\end{figure}
\paragraph{Portfolio optimization problem.} 
Note that this is an optimization problem with continuous decision variables having \textbf{quadratic constraints and a linear objective function.} Hence, the HSD approach is not applicable for this problem, as it cannot handle non-linear constraints.
However, as \textsl{cvxpylayers} is used to implement QPTL, it can be applied in this problem.
The boxplots of test regrets for noise magnitude
parameter $\vartheta$ being 1 are shown in Figure 
\ref{fig:portfolio_noise1}. 

In this problem, in some problem instances, all the return values are negative, which makes a portfolio with zero return to be the optimal portfolio. In such cases, relative regret turns infinite as the denominator is zero in Eq.~\eqref{eq:relative_regret}. Hence, for this problem set, the \textbf{absolute regret} instead of relative regret is reported in Figure \ref{fig:portfolio_noise1}. 
The boxplots for Deg values of 1 and 16 are shown in 
The boxplots for all the four degrees are shown in Figure \ref{fig:portfolio_noise1_append} in the Appendix.

Apparently, the PF approach performs very well in this problem; but SPO manages to outperform PF slightly in all cases except for Deg 1. It is evident in Figure \ref{fig:portfolio_noise1} that  DBB, I-MLE, FY and QPTL perform miserably as they generate regret even higher than the PF approach. 
All these techniques were proposed considering problems with linear constraints.
Hence concerns arise that these techniques might not be suitable in the presence of quadratic constraints. 
On the other hand, LTR losses---Pairwise and Pairwise(diff) and  contrastive loss function, MAP, perform even better than SPO for Deg 16.
The Listwise LTR loss exhibits high variance for Deg 1 and for Deg values of 4 and 8.
For Deg 16, it generates average test regret lower than SPO. 
In general, Figure \ref{fig:portfolio_noise1} reveals 
DBB, I-MLE, FY and QPTL perform poorly in this problem, whereas, SPO, MAP Pairwise  and Pairwise(diff) seem to be the best performing DFL techniques for this problem.
\begin{figure}
    \centering
    \begin{subfigure}[b]{\textwidth}
    \centering
    \includegraphics[width=\textwidth]{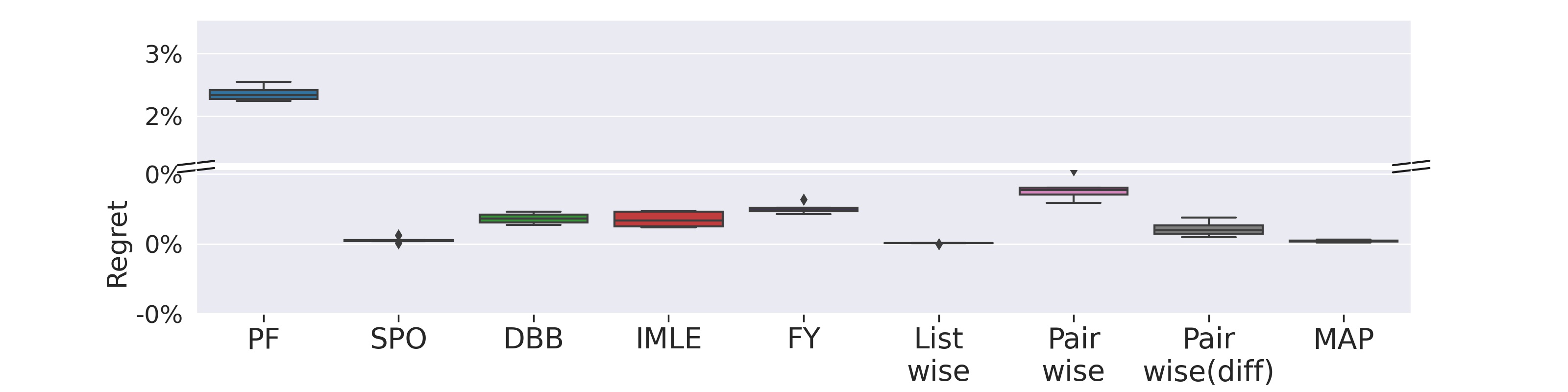}
    \caption{$\text{Image Size } 12 \times 12$}
    \label{fig:warcraft_size12}
  \end{subfigure}
  \begin{subfigure}[b]{\textwidth}
    \centering
    \includegraphics[width=\textwidth]{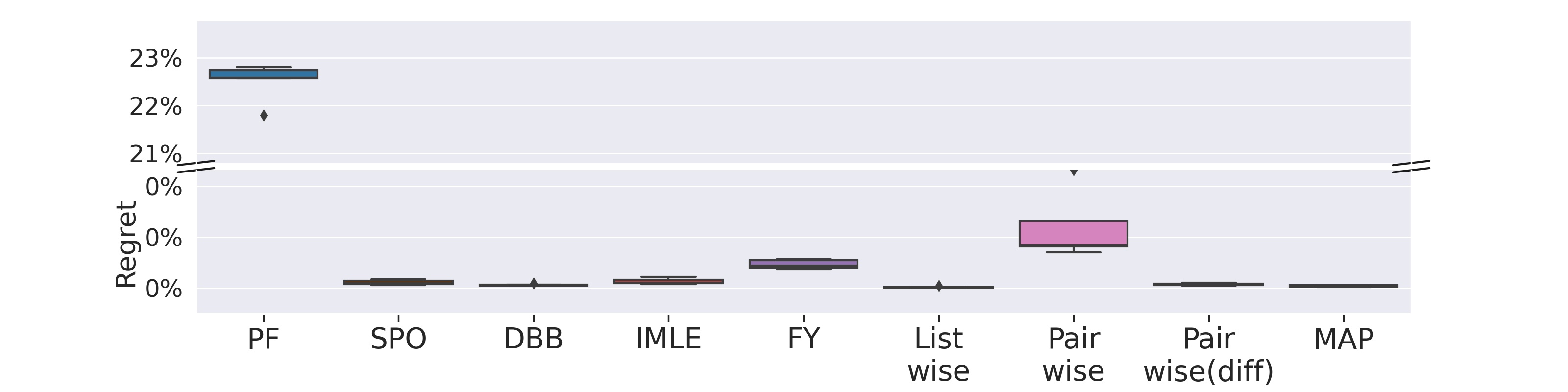}
    \caption{$\text{Image Size } 30 \times 30$}
    \label{fig:warcraft_size30}
  \end{subfigure}
    \caption{Comparative evaluations on the Warcraft shortest path problem instances. The boxplots show the distributions of relative regrets.}
    \label{fig:warcraft}
\end{figure}
\paragraph{Warcraft shortest path problem.} 
 Recall that this a shortest path problem in an image with dimension $d \times d$. 
 The optimization problem can be efficiently solved using Dijkstra's algorithm \cite{dijkstra1959note}, as underlying costs of all the pixel values are non-negative. 
  Hence the shortest path problem is solved using Dijkstra's algorithm for the techniques which view the CO solver as a blackbox oracle. However, HSD and QPTL require the problem to be formulated as an LP and require a primal-dual solver.
Note that in this experiment, the predictive ML model is a CNN, which predicts the cost of each pixel. 
In this case, training the ML model is challenging due to the large number of parameters. 
Hence, combining this ML model with computation-intensive modules such as a primal-dual solver poses significant challenges. 
 We could not run the experiments with HSD and QPTL because of this computational burden.
 
 The dataset contains four values of $d$: $12, 18, 24, 30$. 
 Clearly, as the value of $d$ increases, 
 the number of parameters of the CO problem increases. 
 The boxplots of comparative evaluations are summarized in Figure~\ref{fig:warcraft}. 
 The boxplots of the other two values of $d$ can be found in Figure~\ref{fig:warcraft_append} in the Appendix.
First, note that the PF approach, which is trained by minimizing MSE loss between the predicted cost and the true cost, performs significantly worse than the DFL techniques. 
In fact, the performance of the PF approach deteriorates as the image size increases. 
 As the size of the image increases, the same level of prediction error induces greater inaccuracies in the solution.
  This is because an increase in the area of the image involves dealing with a greater number of decision variables in the CO problem. When the level of prediction error remains constant, the probability of 
  the error in prediction changing at least one of the decision variables also increases. Consequently, there is a higher likelihood of error in the final solution.
  As the regret of the PF approach is significantly higher, note that the scale of the y-axis is changed to fit it into the plot.
  
  Among the DFL techniques, Listwise performs best for sizes 12, 18, and 30 and SPO performs best for size 30. In fact, for sizes 12, 18, and 24, there are not many variations between SPO, Listwise, and MAP.
  After them, the next three best-performing techniques  are Pairwise (diff), I-MLE and DBB.
  However, for size 30, DBB comes third after Listwise and MAP, followed by Pairwise (diff), SPO, and I-MLE in that order. FY and Pairwise perform slightly worse than the other DFL techniques.
  In general, this set of experiments shows the advantage of the DFL approaches as all of them outperform the PF approach.
\begin{figure}
    \centering
    \begin{subfigure}[b]{\textwidth}
    \centering
    \includegraphics[width=\textwidth]{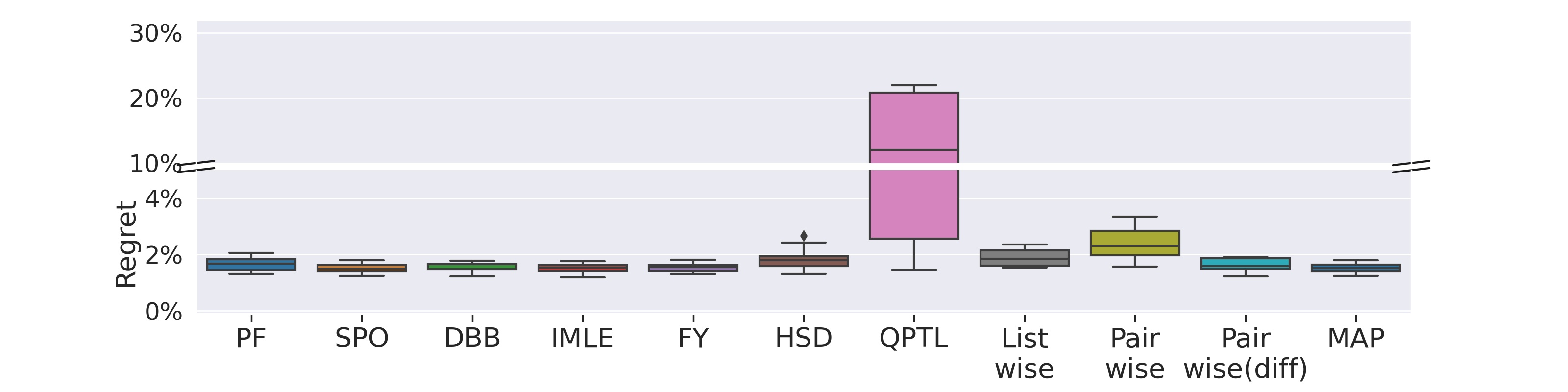}
    \caption{Instance 1}
    \label{fig:energy_scheduling_1}
  \end{subfigure}
    \caption{Comparative evaluations on the energy-cost aware scheduling problem instances. This boxplot shows the distributions of relative regrets.}
    \label{fig:energy_scheduling}
\end{figure}
\paragraph{Energy-cost aware scheduling.}
There are three instances of this scheduling problem. 
All the instances have 3 machines. The first, second, and third instances contain 10, 15, and 20 tasks, respectively. 
 In this problem, the underlying ML model is a simple linear model implemented as a neural network model with no hidden layers. 
 The boxplot of comparative evaluations for the first instance is presented in Figure \ref{fig:energy_scheduling}.
 The boxplots of the other instances can be found in Figure~\ref{fig:energy_scheduling_append} in the Appendix.
 
Note that the scheduling problem is an ILP problem. For HSD and QPTL, the LPs obtained by relaxing the integrality constraints have been considered.
QPTL and HSD also perform poorly in all three instances.
Listwise and Pairwise LTR losses also result to higher regret than the PF approach.
On the other hand, FY, SPO, I-MLE, Pairwise(diff) and MAP come out as the best performing ones closely followed by DBB.

In general, across the three problem instances, it is possible to identify some common patterns. The first one is relaxing the integrality constraints fails to capture the essence of the combinatorial nature of the LP. Consequently, HSD and QPTL perform poorly.  Secondly, Listwise and Pairwise ranking performances are significantly worse than the PF approaches. The learning curve suggests (refer to Appendix \ref{sect:energylearning curve}), these methods fail to converge in these problem instances, although in some epochs, they are able to perform significantly better than the PF approach, their performances never plateau. Lastly, SPO, MAP, FY, and I-MLE perform consistently better than the rest. 

\begin{figure}
    \centering
    \begin{subfigure}[b]{\textwidth}
    \centering
    \includegraphics[width=\textwidth]{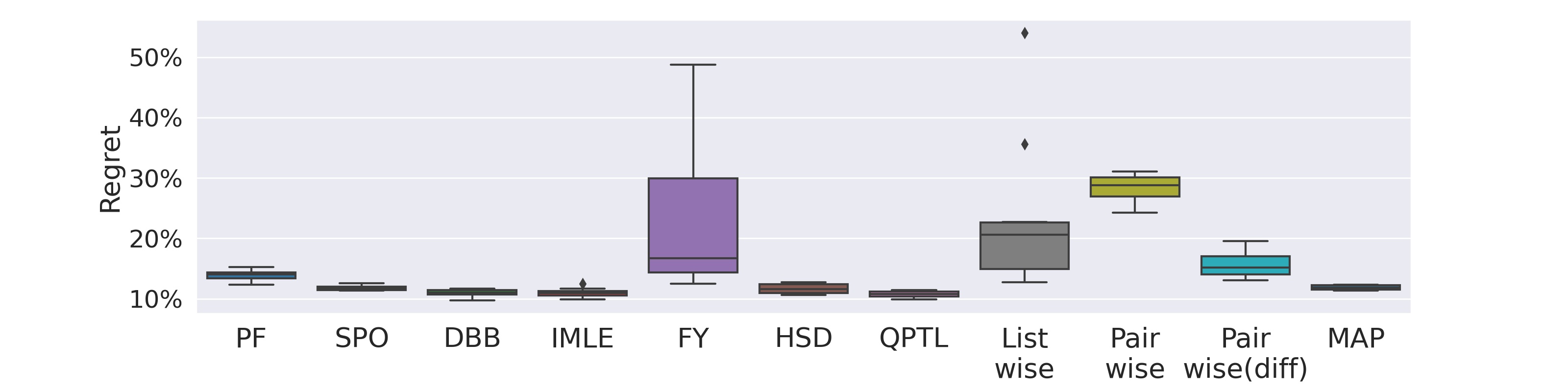}
    \caption{Capacity= 60}
    \label{fig:knapsack_capa60}
  \end{subfigure}
    \caption{Comparative evaluations on the knapsack problem instances. This boxplot shows the distributions of relative regrets}
    \label{fig:knapsack}
\end{figure}
\paragraph{Knapsack problem.} Three instantiations of the knapsack problem are considered for the experiment---each instantiation with a different capacity. 
The three capacity values are---60, 120 and 180.
The boxplot corresponding to capacity value 60 is presented in Figure \ref{fig:knapsack}. 
The boxplots of the other two capacities can be found in Figure \ref{fig:knapsack_append} in the Appendix.
With a capacity of 60, QPTL performs the best, whereas
DBB, and I-MLE, HSD, SPO, and MAP also results in lower regret than the PF approach.
With capacity values of 120 and 180, DBB has the lowest regret. 
Again I-MLE and SPO, HSD and QPTL result in lower regret than PF.
However, MAP, LTR losses, and FY perform poorly especially for high capacity values.
\paragraph{Diverse bipartite matching.} Three instantiations of the diverse bipartite matching problem are formed by changing the values of $\rho_1$ and $\rho_2$. The values of $(\rho_1, \rho_2)$ for the three instantiations  are $(10\%, 10\%)$, $(25\%, 25\%)$, $(50\%, 50\%)$ respectively.
As mentioned before, in this problem, each edge is not associated with an edge weight in the true sense.
Hence, the PF approach is trained by directly learning to predict whether an edge exists. So the loss used for supervised learning for the PF approach is \textbf{BCE} loss.
The DFL approaches consider the predicted probability of each edge as the edge weight and then aim to minimize regret.

\begin{center}
\begin{figure}
    \centering
    \begin{subfigure}[b]{\textwidth}
    \centering
    \includegraphics[width=\textwidth]{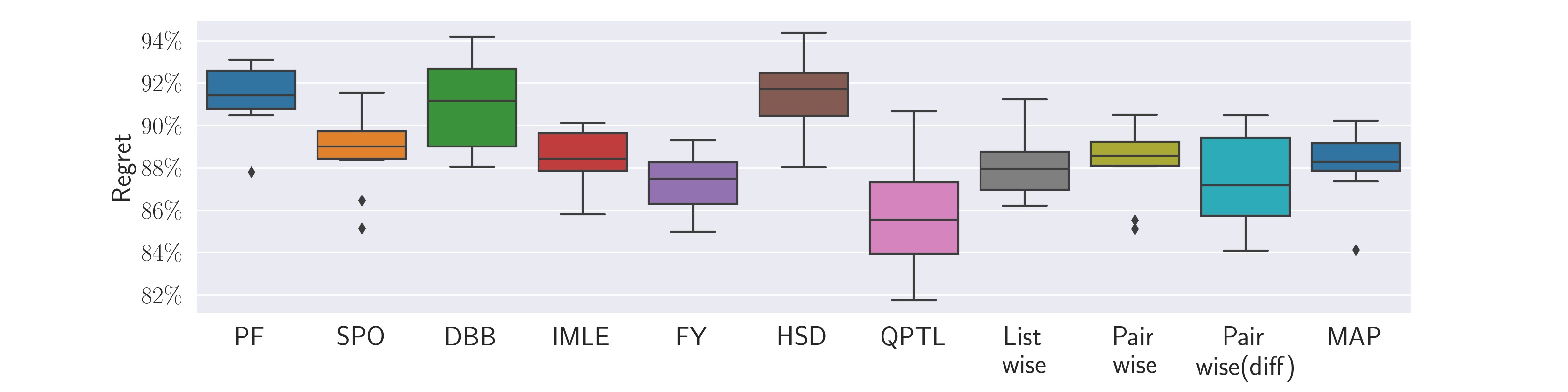}
    \caption{$\rho_1 = \rho_2 = 50\%$}
    \label{fig:matching_50}
  \end{subfigure}
    \caption{Comparative evaluations on the diverse bipartite matching problem instances. This boxplot shows the distributions of relative regrets.}
    \label{fig:matching}
\end{figure}
\end{center}

The boxplot of comparative evaluations for $(\rho_1, \rho_2)$ being $(50\%, 50\%)$, is presented in Figure \ref{fig:matching}.
The boxplots of the other two instances  can be found in Figure~\ref{fig:matching_append} in the Appendix.
Firstly note that relative regrets of all the techniques are very high (higher than 80\%) for all three instances.
With $\rho_1$ and $\rho_2$ being $10\%$, I-MLE performs considerably better than all the other DFL techniques. 
When $\rho_1$ and $\rho_2$ take the value of $25\%$, QPTL, I-MLE and HSD are the top there techniques, with significantly lower regret than the rest.
In the first two instances, other than I-MLE and QPTL, other DFL techniques do not significantly better than the PF approach.
DFL techniques such as SPO, FY, Listwise, Pairwise and MAP perform considerably better than the PF approach only in the third instances.

\paragraph{Learning subset selections.} 

Subset selection problems of three dimensions: $n = 25$, $n = 50$, and $n = 100$ are considered for evaluation. In each case, the subset size $k$ is chosen to be $\frac{n}{5}$. The error of any predicted subset $\hat{x}$, with respect to ground truth $x$, is considered to be the fraction of items which are selected in $x$ but not in $\hat{x}$.  Such occurrences are referred to as mismatches. 

\begin{figure}
    \centering
    \includegraphics[width=1.0\textwidth]{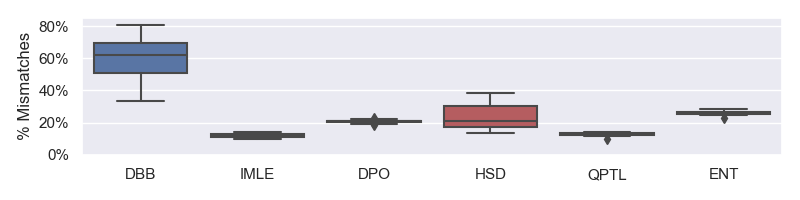}
    \caption{Comparative evaluations on the subset selection problem instances of Size $25$. This boxplot shows the distributions of mismatch rates.}
    \label{fig:subsets_size25}
\end{figure}

Figure \ref{fig:subsets_size25} shows the average mismatch rates over the size  $n = 25$ instances that were achieved by each DFL technique listed in Table \ref{tab:dfl_formula}, excluding those which assume ground-truth data in the form of problem parameters. Here, the ground-truth data are optimal solutions of \eqref{eq:topk-lp} representing subset selections. For each assessed technique, a distribution of results is shown, corresponding to $10$ different randomly generated training datsets. Figure \ref{fig:subsets_append} shows similar results over the larger problem instances.

Note that it is suggested in \cite{amos2019limited} that the entropy function $H(\mathbf{x}) = \sum_i x_i \log x_i$ is particularly well-suited as a regularizer of the objective in \eqref{eq:topk-lp}, for the purpose of multilabel classification, which is identical to the task in terms of its optimization component and the form of its target data. Hence a \textsl{Cvxpylayers} implementation of this model is included and referred to as ENT.

Figure \ref{fig:subsets_append} shows that most of the assessed techniques perform similarly, with DBB performing worst regardless of the problem's dimension. HSD is most sensitive with respect to the randomly generated training set; the rest show consistent performance across datasets. QPTL and IMLE each show a marginal advantage over the other techniques, but DPO and ENT are also competitive. Across all techniques, variation in performance over the randomly generated datasets tends to diminish as problem size increases.

\subsubsection{Comparison on Runtime}
While training the ML model to minimize the task loss is the primary challenge of DFL,
the computational cost associated with repeatedly solving CO problems gives rise to the second challenge.
DFL techniques with low computational cost are essential for scalability for implementing DFL for real-world large-scale \pto problems. 
The importance of scalability and low computational cost becomes significant while dealing with large-scale CO problems, especially NP-hard combinatorial optimization problems. 
Note that while the shortest path and the knapsack problems are relatively easy to solve; the energy-cost aware scheduling problem is much more challenging.
That is why the scheduling problem is considered to compare the computational costs of the DFL techniques. 

\begin{figure}
    \centering
    \begin{subfigure}[b]{\textwidth}
    \centering
    \includegraphics[width=\textwidth]{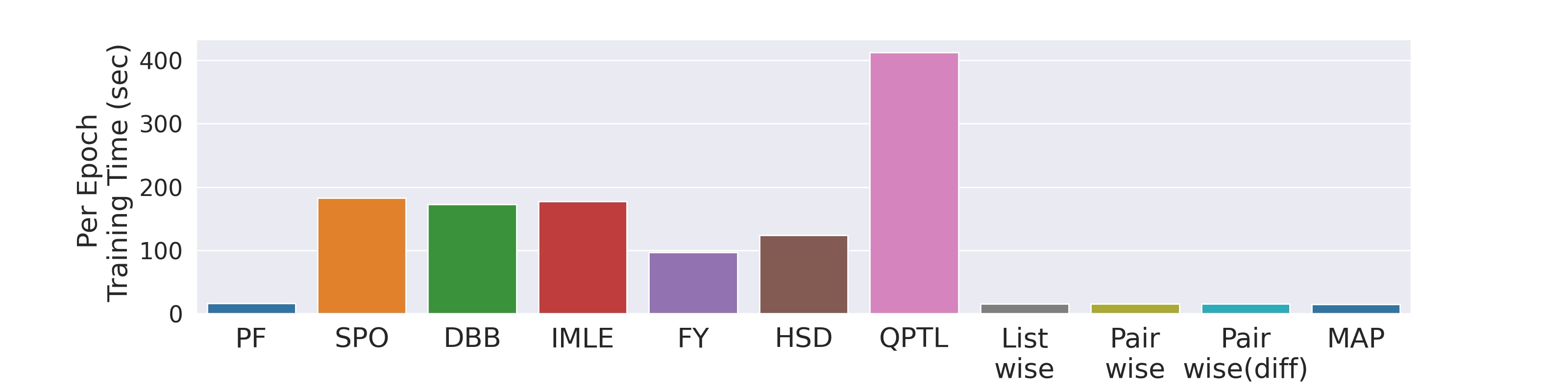}
    \caption{Instance 2}
    \label{fig:energy_runtime_2}
  \end{subfigure}
  \begin{subfigure}[b]{\textwidth}
    \centering
    \includegraphics[width=\textwidth]{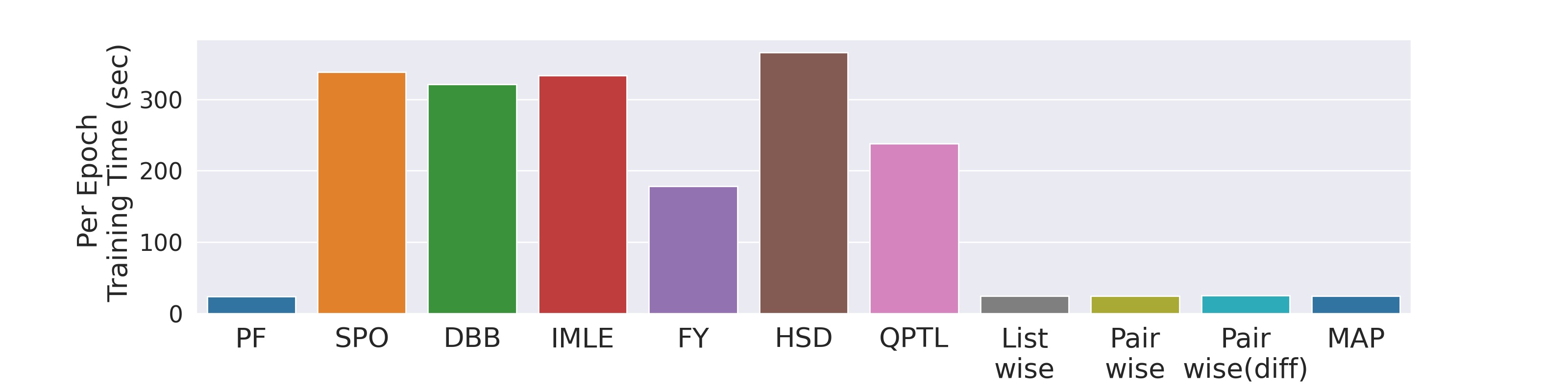}
    \caption{Instance 3}
    \label{fig:energy_runtime_3}
  \end{subfigure}
    \caption{Comparative evaluations of per epoch training time of different DFL techniques on the energy-cost aware scheduling problem.}
    \label{fig:energy_runtime}
\end{figure}
The median training time of an epoch during training of each technique for two instances of the scheduling problem is shown Figure \ref{fig:energy_runtime}.
Recall that the first, second and third instances contain 10, 15 and 20 tasks respectively. So, the first one is the easiest of the three and the third one is the hardest one.
The complexity of the scheduling problem is evident from the fact that a single instance of
the knapsack problem takes $0.001 $ seconds to solve, while solving the most difficult instance of the scheduling problem takes $0.1 $ seconds, both using Gurobi MIP solver.
The readers are cautioned against placing excessive emphasis on the absolute values of training times in Figure \ref{fig:energy_runtime},
as they are subject to system overhead. However, some general conclusions can be drawn from the relative ordering of the training times.
The training time of the PF approach is the lowest, as it does not require solving the CO problem for training. 
Training times of SPO, DBB, I-MLE and FY are almost 100 times higher than the PF approach. 
Although QPTL and HSD consider the relaxed LP problem, 
it is not always the case that they have lower training times.
Recall that QPTL and HSD  solve and differentiate the optimization problem using a primal-dual solver, which involves matrix factorization. 
On the other hand, SPO, DBB, I-MLE and FY can leverage faster commercial CO solvers, as they only require the optimal solution.
However, for Instance 3, it seems solving the ILP problem by calling a CO solver, is more computationally expensive than solving and differentiating the underlying QP problem using \textsl{cvxpylayers}. 

On the other hand, Listwise, Pairwise, Pairwise(diff) and MAP, all of which are run with $p_{solve}=5\%$, exhibit significantly lower training time than the other DFL techniques. 
From this perspective, 
these techniques lie between PF and DFL approaches, balancing the trade-off between scalability and quality.
The same conclusion generally holds true for other experiments as well. However, for relatively easier CO problems, the system overhead time may dominate over training time, which might disrupt the ordering of the training time. 
\subsubsection{Discussion}
The experimental evaluations reveal that no single DFL technique performs the best across all experiments.
Some techniques work well on particular test problems, while others work better on other test problems.
Nevertheless, the following interesting characteristics can be observed in the experimental evaluations:
\begin{enumerate}[label=(\roman*)]
\item \textbf{The performance of SPO is consistently robust across the test problems}, even though it may not outperform other techniques in every experiment. 
\item I-MLE, FY, DBB and QPTL perform worse than the PF approach for the \textbf{portfolio optimization problem, where a quadratic constraint is present.}

\item \textbf{QPTL demonstrates robust performance when the relaxed LP is a good approximation of the ILP}. QPTL outperforms other techniques by a substantial margin in the bipartite matching problem. However, QPTL performs poorly compared to others in the scheduling problem, which is an ILP. 
In this case, the poor performance may be attributed to the fact that QPTL considers a relaxation of the ILP.  In this problem, the LP solution might differ significantly from the true ILP solution. This is not the case for the knapsack problem, because 
the solution of the relaxed LP does not deviate significantly from the ILP solution for the knapsack problem.

\item \textbf{MAP also demonstrates consistent performance across most test problems}; it exhibits higher regret specifically in the knapsack problem for Capacity=180 and in the bipartite matching problem when $\rho_1$ and $\rho_2$ are $25\%$. 
It results in low regret in the portfolio optimization problem; where some DFL techniques perform poorly.

\item  Among the LTR losses, \textbf{Listwise and Pairwise often exhibit high variances, especially in the scheduling and the knapsack problems}. The performance of Pairwise(diff) stands out among the LTR losses due to its lower variance. Its performance is comparable to or slightly worse than MAP for most problems other than the synthetic shortest path problem with high values of Deg, i.e., when the underlying predictive model is completely misspecified.  

\item Due to the limitation of computational resources, it was not possible to run QPTL and HSD on the Warcraft shortest path problem. This highlights the \textbf{advantage of DFL techniques which can make use of any blackbox combinatorial solver (Dijkstra's shortest path solver for instance) to solve the CO problem}. 

\item Continuing on this topic of computational cost, \textbf{MAP and the LTR losses are considerably faster and less computationally intensive when they are run with low values of $p_{solve}$}. 
For large-scale real-world CO problems, the repeated solving of which might be prohibitive,
MAP could be used to implement DFL, as
MAP tends to have regret as low as SPO in most test problems.
\end{enumerate}

\section{Future Research Directions}\label{sect:future}
While there is increasing interest in decision-focused learning research, more work is required to enable its application in a wider range of real-world problems. 
This section aims to summarize some challenges in DFL research that remain open.

\paragraph{Generalizing DFL across related tasks.} 
Existing DFL methods typically tailor the machine learning model to a specific combinatorial optimization task. However, in many real-world applications, the CO problem can vary across different instances. For instance, in the MIT-Amazon Last Mile Routing Challenge \cite{AmazonLastMile}, a Traveling Salesman Problem is solved daily to determine the routing of last-mile package delivery. The nodes in these Traveling Salesman Problem instances change daily as delivery locations vary. A promising research direction would be to investigate the performance of a model trained to minimize regret on one CO problem instance when evaluated on similar but different CO problem instances. Other research directions could involve the integration of multi-task learning with decision focused learning. This exploration could lead to the development of more versatile and robust DFL models capable of generalizing across a range of related tasks, thereby greatly enhancing their practical applicability.

\paragraph{Non-linear objective function.} 
Most DFL research has focused on optimization problems with linear objective functions, as those emphasized in the experimental evaluations in this article. However, many real-world combinatorial optimization problems feature non-linear objective functions and discrete decision variables. For instance, optimally locating substations in an electrical network to minimize distribution costs is a problem often formulated as non-linear programming \cite{lakhera2011approximating}. Similarly, minimizing the makespan in flowshop scheduling is another classic operations research problem that lacks a linear objective function. While \textsl{cvxpylayers} \cite{agrawal2019differentiable} allows for differentiation through convex optimization problems with non-linear objective functions, most techniques discussed in this paper are not equipped to handle such scenarios. Future research should focus on developing and evaluating DFL techniques that can effectively address combinatorial optimization problems with non-linear objective functions. This advancement would significantly expand the applicability and impact of DFL in solving complex real-world problems.

\paragraph{Robust risk-sensitive DFL.}
Most DFL techniques formulate empirical risk minimization problems to minimize expected regret. However, in many real-world applications, particularly those that are risk-sensitive, safety concerns renders the focus from expected regret less relevant. A more viable alternative in these cases would be to focus on regret in distributional worst-case scenarios. This necessitates minimizing risk-sensitive losses, such as value-at-risk or min-max regret. To address these cases, future research could draw inspiration from robust optimization \cite{ben2009robust}, a well-established field within operations research that addresses worst-case regret within an uncertainty set. The prediction-focused approach could be adapted to design an ``estimate-then-optimize'' strategy \cite{qi2023integrated}, incorporating risk-sensitive loss functions by creating a weight-based empirical distribution to represent the uncertainty set \cite{NEURIPS2023_PredictthenCalibrate}. However, the DFL methodologies reviewed in this article are not designed to minimize risk-sensitive losses. While a recent work \cite{ChenreddyNEURIPS2022} proposes a data-driven approach for constructing uncertainty sets using contextual features, the literature on robust DFL remains sparse, offering ample opportunities for future research.

\paragraph{Decision-focused learning by zeroth-order gradient.}
An alternative approach to gradient estimation is to adopt zeroth-order gradient estimation, particularly through the score function gradient estimation technique \cite{williams1992score}. This method has been recently adapted to DFL, where it assumes that the predicted parameter adheres to a specific distribution, and the model is trained accordingly to predict this distribution's parameters \cite{silvestri2023score}. In principle, this method can also be used to predict parameters in the constraints of CO problems.
However, despite providing an unbiased gradient, a significant drawback of the score function gradient estimation is its susceptibility to high variances, which can destabilize the learning process. Addressing this challenge could lead to substantial improvements. Enhancements could focus on variance reduction techniques or the development of more robust estimation algorithms.



\paragraph{Bilevel Optimization Techniques for DFL.}
As mentioned in Section \ref{sect:learningparadigm}, the empirical regret minimization problem can be cast as a pessimistic bilevel optimization problem. A deeper understanding of the mathematical foundations behind this learning process can lead to the development of more effective algorithms for DFL. Recent efforts by \citeA{bucarey2023decision} have reformulated the problem as a quadratic non-convex optimization problem. However, the scalability of this approach remains a significant challenge. This presents a valuable opportunity for the bilevel optimization community to develop scalable and efficient solutions, advancing the field of DFL and enhancing its practical applications.

\paragraph{Scalable DFL.}
While surrogate solvers and solution caching aim to address the scalability challenges in DFL, the trade-off between quality and scalability of these methods is not yet understood. In the context of solution caching, there is ample space for research to investigate the trade-off between the solution cache size and the resulting solution quality. Determining the optimal cache size that balances computational efficiency with high-quality solutions could lead to significant improvements in DFL performance.
Additionally, the effectiveness of using pre-trained surrogate models as proxy CO solvers remains an open question. Specifically, it is unclear whether these surrogates can achieve regret levels comparable to those obtained using traditional CO solvers. Future research should explore the conditions under which surrogate solvers can approximate the performance of CO solvers, and identify strategies to enhance their accuracy and reliability. This line of research could reveal new insights into the design of more scalable DFL systems, which is one of the most important challenges to broaden their applicability to a wide range of complex problems.

\paragraph{Theoretical guarantees.}
While there has been extensive work on providing theoretical guarantees for the SPO+ loss, not all DFL methods offer theoretical guarantees on minimizing regret. Empirical results presented in this article suggest that the performance of some DFL methods may be suboptimal when dealing with combinatorial optimization problems that contain non-linear constraints. This gap underscores the need for further theoretical developments to establish the reliability and effectiveness of DFL methods across a broader range of optimization scenarios.
Ensuring robust theoretical guarantees for DFL methods, particularly in the presence of non-linear constraints, would significantly enhance their trustworthiness especially important for safety-critical applications. This involves developing new analytical tools and techniques to prove the regret-minimizing properties of DFL methods under various conditions. Moreover, research should also focus on extending existing theoretical frameworks to encompass a wider array of CO problems.

\paragraph{Uncertainty in the constraints.}
As reviewed in Section \ref{sect:df_for_constraints}, some recent works have focused on DFL methods for predicting parameters within the constraints of optimization problems. However, these methods have not been explored to the same extent as those predicting objective function parameters, leaving several open research directions for future investigation. For instance, future studies could extend the Neural Combinatorial Optimization \cite{BelloPL0B17} and NCE \cite{mulamba2020discrete} approaches  by considering the prediction of parameters within the constraints. A significant challenge to implement the NCE approach in this setting is determining how to form the solution cache.
Moreover, when predicting parameters in the constraints, the incorporation of risk-sensitive losses becomes particularly important. In these scenarios, the prescribed decision might not be feasible with respect to the true parameters. An intriguing research direction is to develop methods that recommend solutions feasible under extreme distributional variations of the parameters. This would involve creating a framework that jointly optimizes average performance while minimizing worst-case constraint violations. Such a framework could reveal new tracks for both theoretical research and practical applications, ultimately enhancing the practical utility of DFL in problems where constraint parameters are often uncertain and variable.

\paragraph{Extending DFL to multistage settings.}
DFL primarily aims to predict parameters for single combinatorial optimization problems. However, in many practical settings, decision-making occurs over multiple time periods, resulting in multistage optimization problems \cite{pflug2014multistage}. 
This setting is prevalent in production and inventory management \cite{multigoyal1990}, where optimal decisions are made over an extended time horizon in successive stages. At each stage, a subset of uncertain parameters is revealed sequentially, and the decision-maker adjusts their decisions to account for this new information.  To date, no studies have investigated whether applying DFL to predict parameters in multistage optimization leads to improved final objectives.

Furthermore, in numerous AI and ML applications, optimization tasks often serve as intermediate steps between two machine learning stages. For example, consider the task of selecting relevant patches in high-resolution images for a downstream image recognition task. The patch selection task can be modeled as a Top-$k$ selection problem \cite{Cordonnier_2021_CVPR} and embedded as an intermediate layer between two neural networks. Here, the upstream neural network assigns scores to each patch, and the downstream neural network performs the recognition task using the Top-$k$ patches. Techniques that implement differentiable optimization layers, such as I-MLE \cite{niepert2021implicit}, DBB \cite{PogancicPMMR20}, QPTL \cite{aaai/WilderDT19}, and DPO \cite{berthet2020learning} can be applied embedded between the two neural networks. However, their effectiveness in these contexts needs to be empirically evaluated. Exploring DFL in multistage settings and its integration in intermediate ML tasks presents a promising research direction that could lead to the development of methods capable of handling sequential decision-making processes and complex optimization tasks embedded within broader ML frameworks.


\paragraph{DFL with Real-World Multimodal Datasets.}

In real-world applications, uncertainty can arise from various sources. Multimodal ML aims to train models that process data from diverse sources \cite{MultimodalML}. In a ``multimodal''  CO problem, different parameters may be unknown and associated with distinct sets of features. For instance, in a facility location problem, both the transport cost and customer demands might be unknown \cite{Cameron_Hartford_Perils}. Each of these parameters could be predicted by different ML models tailored to their respective feature sets.
Integrating DFL with multimodal datasets introduces several new challenges and opportunities. Firstly, coordinating the predictions from multiple ML models to inform a single CO problem requires new methods to ensure coherence and compatibility between different predicted parameters. This might involve developing novel fusion techniques to combine information from different modalities effectively. 
Secondly, the interdependencies between different parameters need to be accurately captured. In multimodal settings, the prediction errors from different models might interact in complex ways, affecting the overall optimization outcome. Research should explore ways to model these interdependencies and mitigate the compounded impact of prediction errors.
Finally, empirical evaluation of DFL methodologies on real-world multimodal datasets would be a crucial avenue to understand their practical utility and limitations. Such studies have the potential to significantly enhance the applicability of DFL in solving complex, real-world optimization problems characterized by diverse and uncertain data sources.

\section{Conclusion}
The survey article began by highligting the significance of \pto problem formulations, wherein an ML model is followed by a CO problem.  The \pto problem has emerged as a powerful driving force in numerous real-world applications of AI, OR and business analytics. The key challenge in \pto problems is predicting the unknown CO problem parameters in a manner that yields high-quality \textit{solutions}, in comparison to the retrospective solutions obtained when using the groundtruth parameters.
To address this challenge, the DFL paradigm has been proposed, wherein the ML models are directly trained considering the CO problems using task losses that capture the error encountered after the CO problems.

This survey article provides a comprehensive overview of DFL, highlighting recent technological advancements, applications and identifying potential future research directions.
Section \ref{sect:prelim}, laid out the problem description with examples and then presented the two fundamental challenges of DFL: differentiation through the solutions of combinatorial optimization mapping and the computational cost associated with solving CO problems in the ML training loop.
Section \ref{sect:review} then distinguished gradient-based DFL techniques from gradient-free DFL techniques, as well as categorizing gradient-based DFL into four distinct classes, thoroughly reviewing the trade-offs among them.
Section \ref{sect:applications} provided examples of DFL applications addressing real-world \pto problems across various domains.
Furthermore, Section \ref{sect:data} offered extensive comparative evaluations on different problem sets, covering 11 DFL techniques.  

Finally, Section \ref{sect:future} discussed some of the open challenges in DFL and outlined potential research directions.
%
Addressing these challenges can pave the way for more robust and effective DFL methodologies, further enhancing their practical utility in solving complex optimization problems.
We hope this survey and corresponding datasets and codes will serve as a catalyst, inspiring the application of decision-focused learning in diverse domains and contexts as well as stimulating further methodological research and advancements.

\acks{
This research is partly funded by the European Research Council (ERC) under the EU Horizon 2020 research and innovation programme (Grant No 101002802, CHAT-Opt)
and the European Union’s Horizon 2020 research and innovation programme under grant agreement
No 101070149, project Tuples.
This research is also partially supported by NSF grants 2242931, 2232054, 2007164, and NSF CAREER award 2143706. 
Jayanta Mandi is supported by the Research Foundation-Flanders (FWO) project G0G3220N.
Senne Berden is a fellow of the Research Foundation-Flanders (FWO-Vlaanderen, 11PQ024N).
Victor Bucarey was funded by the ANID Fondecyt Iniciacion Grant no. 11220864.
}



\appendix

\setcounter{table}{0}
\renewcommand{\thetable}{A\arabic{table}}
\setcounter{figure}{0}
\renewcommand{\thefigure}{A\arabic{figure}}
\section{Results on All Problem Instances}
\begin{figure}
    \centering
    \includegraphics[width=0.7\textwidth]{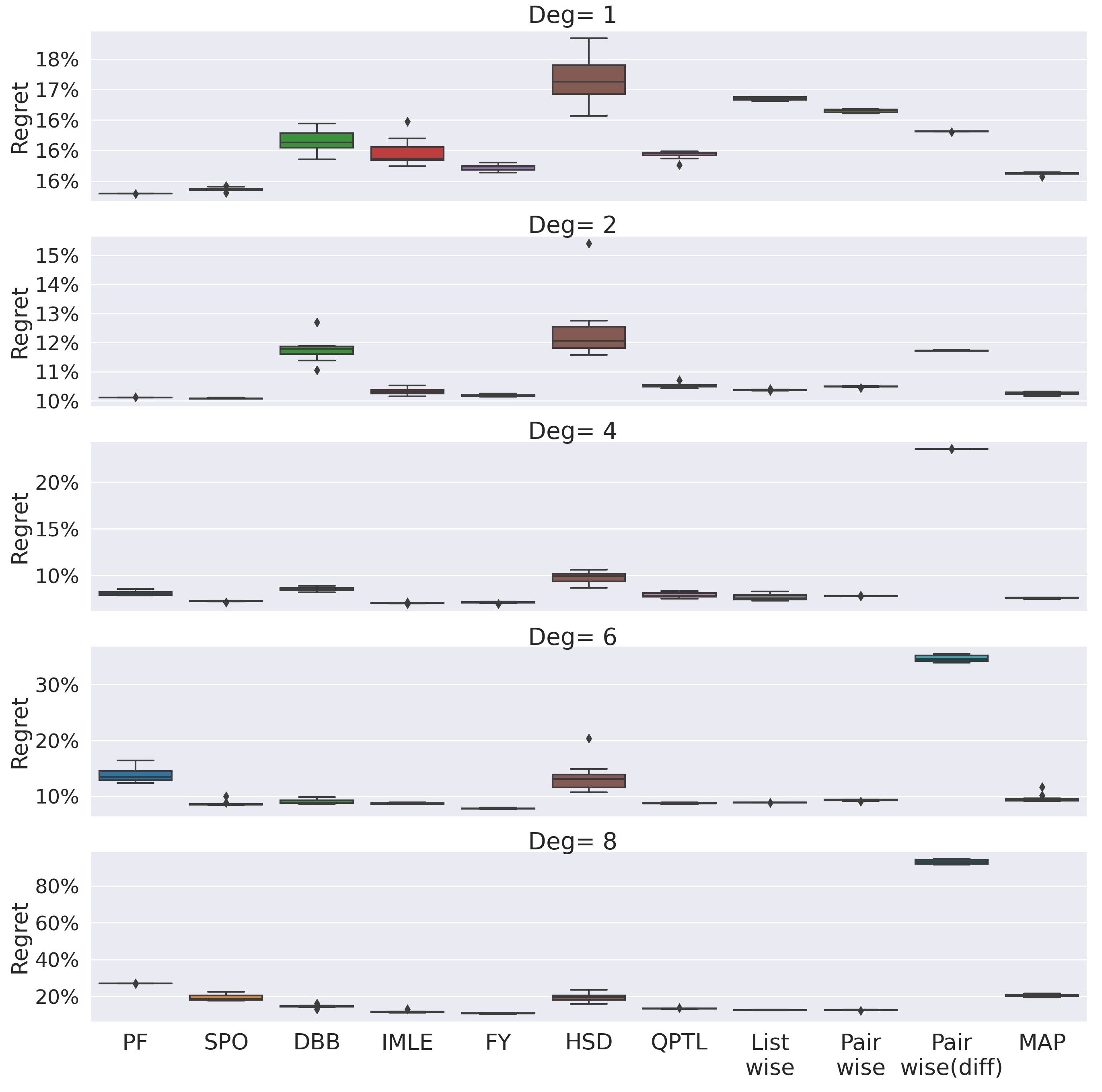}
    \caption{Comparative evaluations on the synthetic shortest path problem with noise-halfwidth parameter $\vartheta$ = 0.5. These boxplots show the distributions of relative regrets.}
    \label{fig:shortest_path_noise0.5_append}
\end{figure}

\begin{figure}
    \centering
    \includegraphics[width=0.7\textwidth]{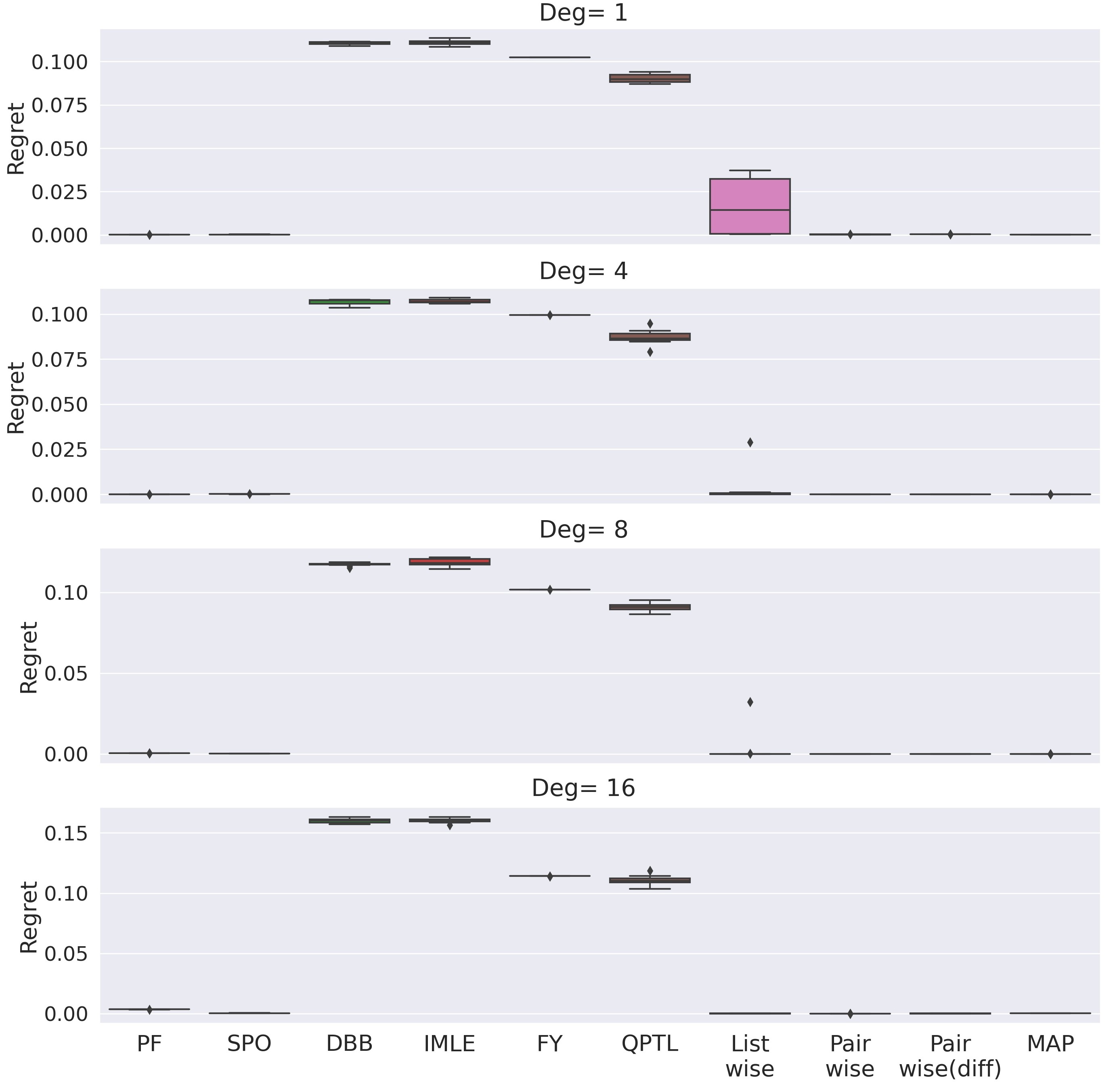}
    \caption{Comparative evaluations on the synthetic portfolio optimization problem with noise magnitude $\vartheta = 1$. These boxplots show the distributions of \textbf{absolute} regrets.}
    \label{fig:portfolio_noise1_append}
\end{figure}

\begin{figure}
    \centering
    \includegraphics[width=0.7\textwidth]{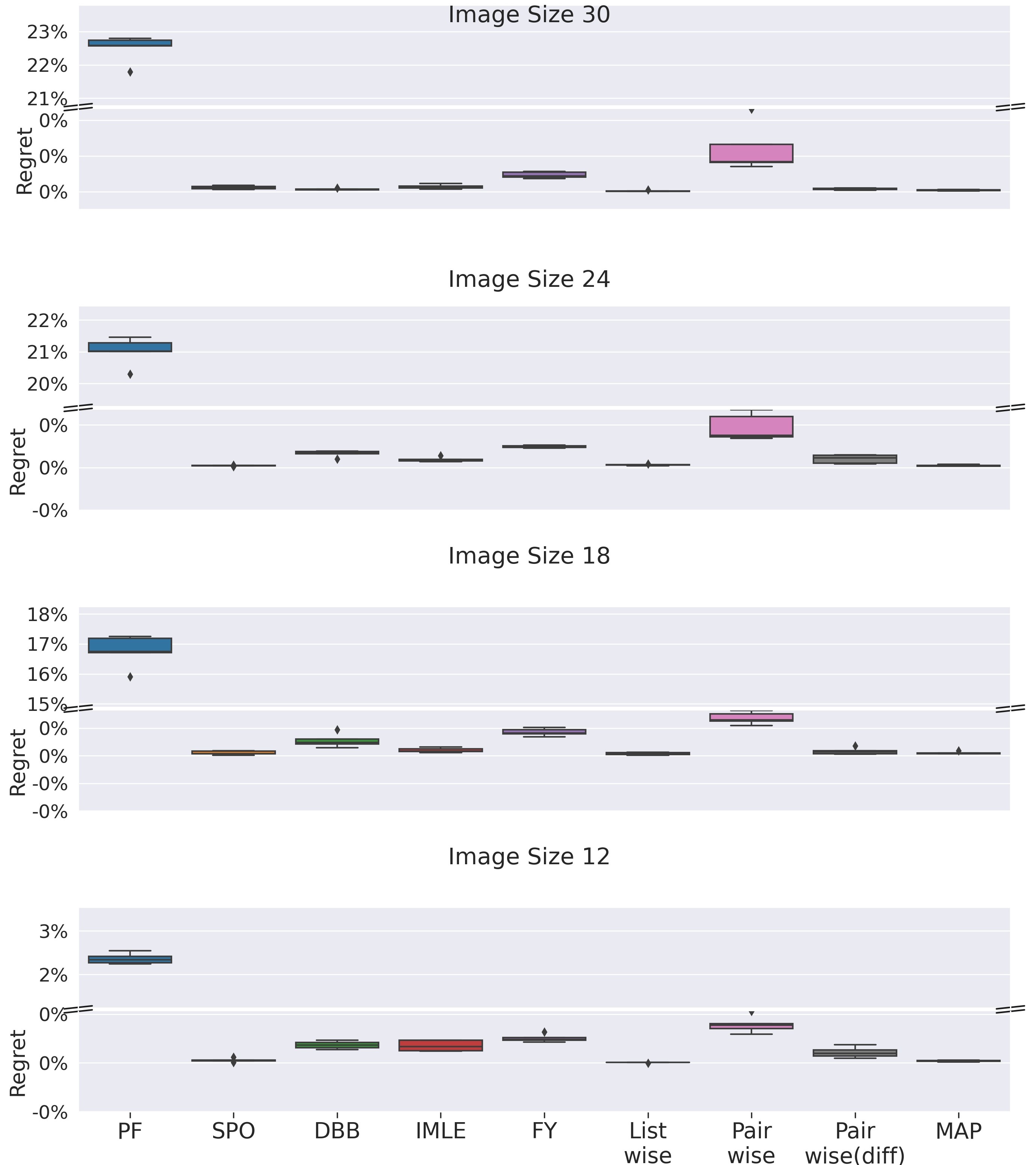}
    \caption{Comparative evaluations on the Warcraft shortest path problem instances. These boxplots show the distributions of relative regrets.}
    \label{fig:warcraft_append}
\end{figure}

\begin{figure}
    \centering
    \includegraphics[width=0.7\textwidth]{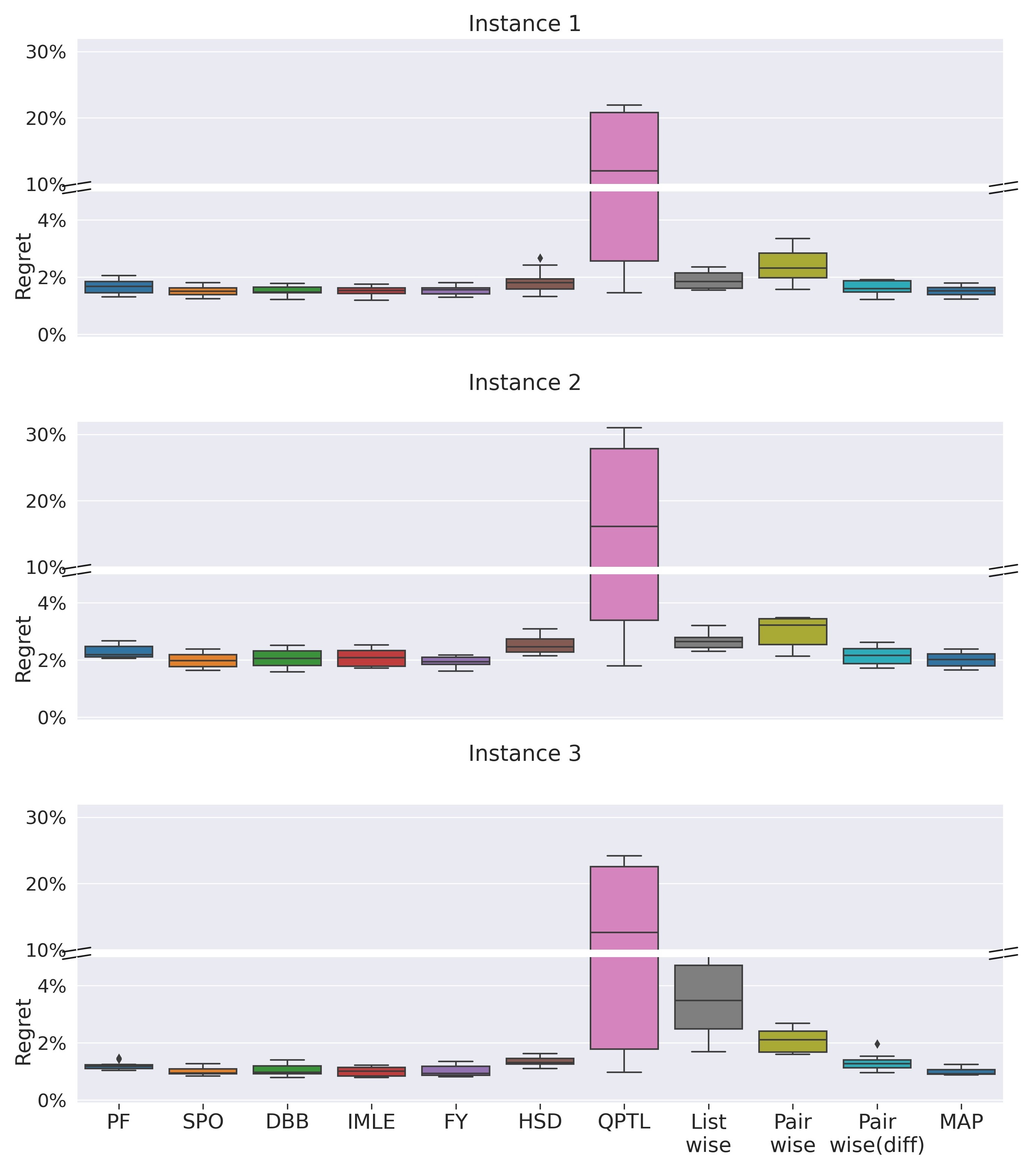}
    \caption{Comparative evaluations on the energy-cost aware scheduling problem instances. These boxplots show the distributions of relative regrets.}
    \label{fig:energy_scheduling_append}
\end{figure}

\begin{figure}
    \centering
    \includegraphics[width=0.7\textwidth]{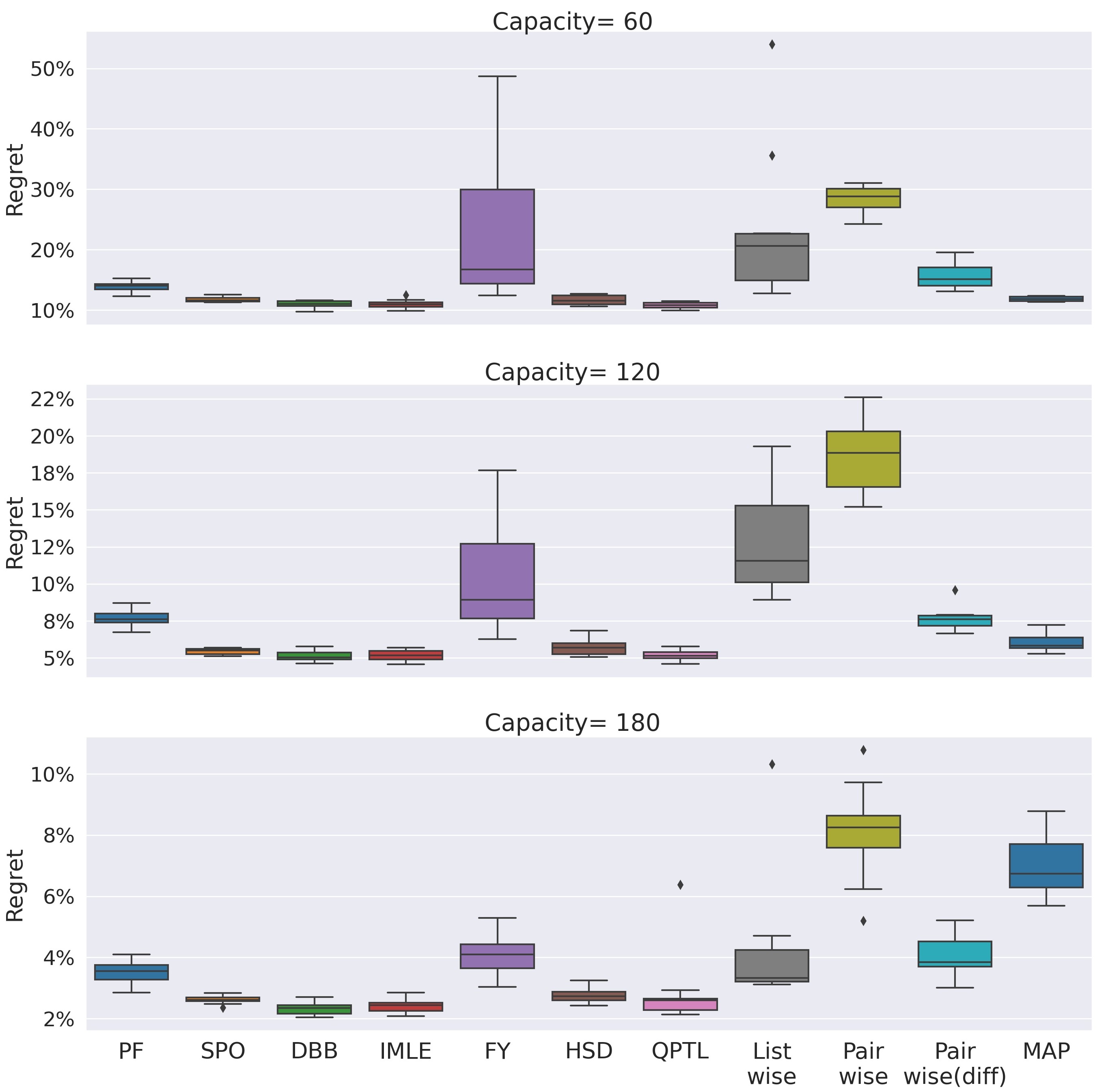}
    \caption{Comparative evaluations on the knapsack problem instances. These boxplots show the distributions of relative regrets.}
    \label{fig:knapsack_append}
\end{figure}

\begin{figure}
    \centering
    \includegraphics[width=0.7\textwidth]{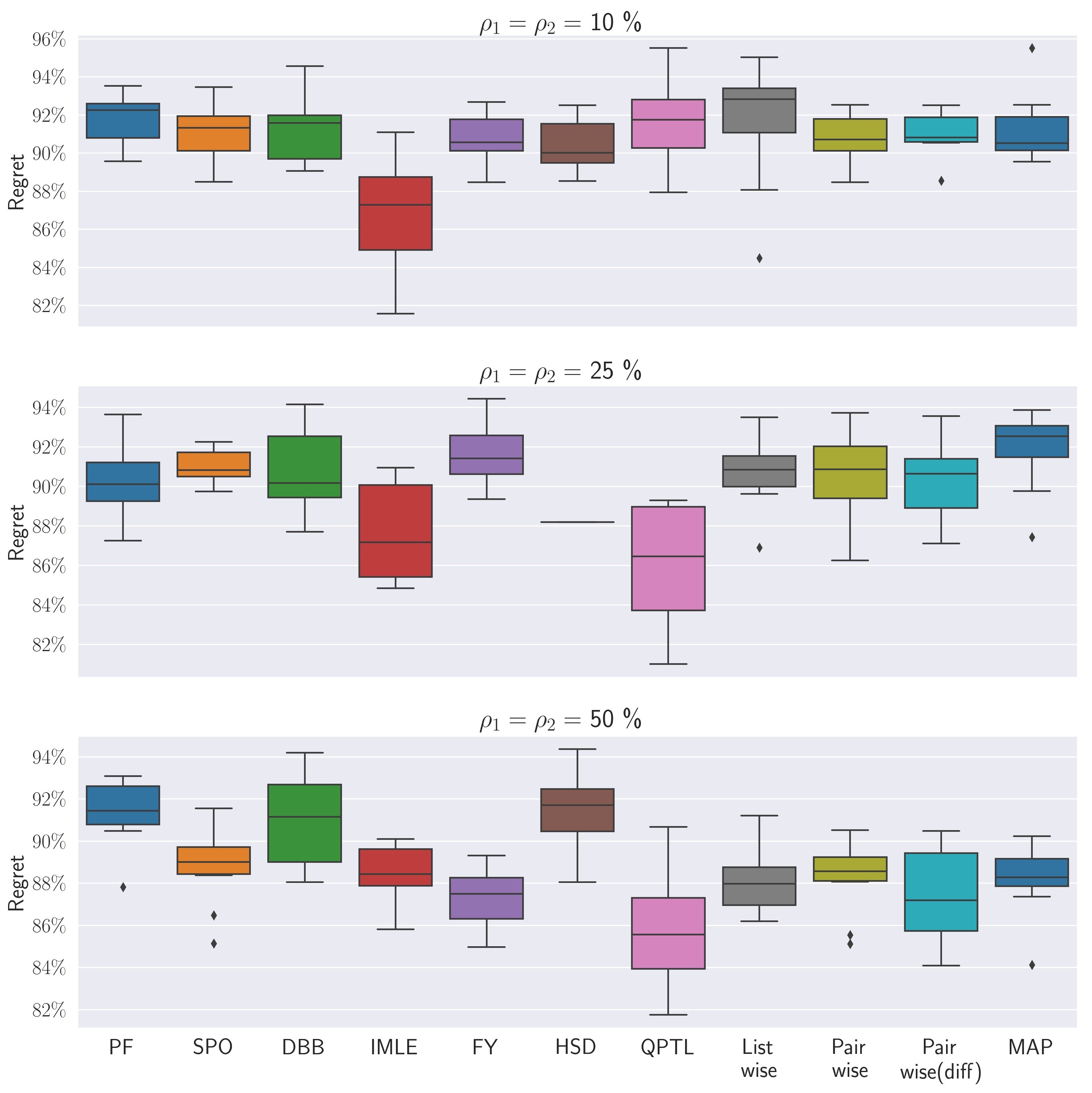}
    \caption{Comparative evaluations on the diverse bipartite matching problem instances. These boxplots show the distributions of relative regrets.}
    \label{fig:matching_append}
\end{figure}

\begin{figure}
    \centering
    \includegraphics[width=0.8\textwidth]{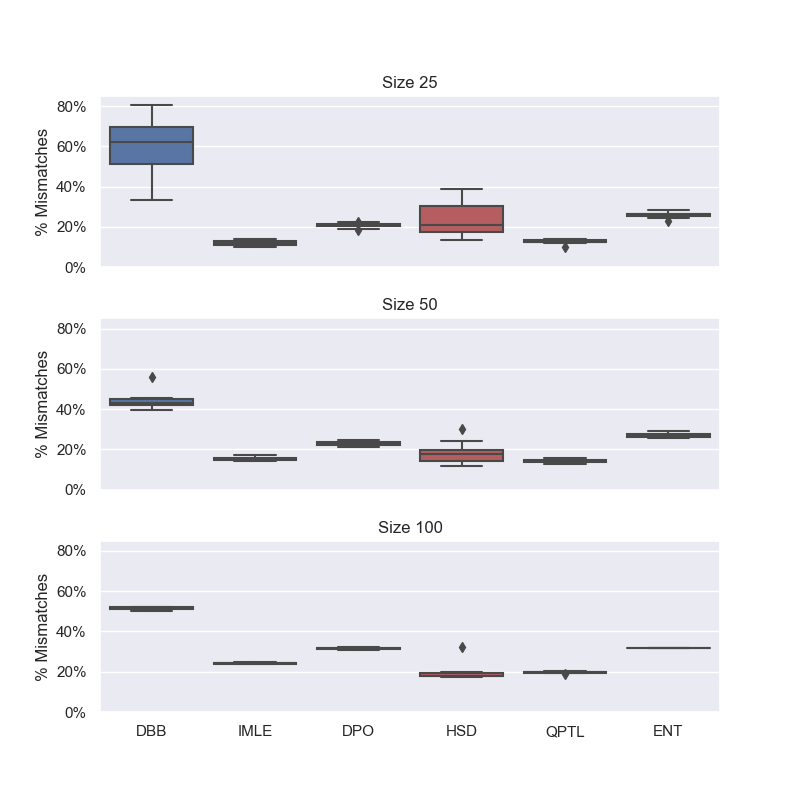}
    \caption{Comparative evaluations on the subset selection problem instances. This boxplot shows the distributions of mismatch rates.}
    \label{fig:subsets_append}
\end{figure}

In the main text, boxplots of only selected problem instances are presented to save space in the main text. 
In this section of Appendix, the boxplots of all the instances are provided.
Figure~\ref{fig:shortest_path_noise0.5_append}, Figure \ref{fig:portfolio_noise1_append}, Figure \ref{fig:warcraft_append}, Figure \ref{fig:energy_scheduling_append}, Figure \ref{fig:knapsack_append}, Figure \ref{fig:matching_append}, and Figure \ref{fig:subsets_append} display the boxplots for all the instances of the grid shortest path problem,
the portfolio optimization problem, the Warcraft shortest path problem, the energy-cost aware scheduling problem, the knapsack problem, the diverse bipartite matching problem, and the subset selection problem respectively.

\section{Learning Curves}
\label{sect:energylearning curve}
\begin{figure}
    \centering
    \includegraphics[width=0.7\textwidth]{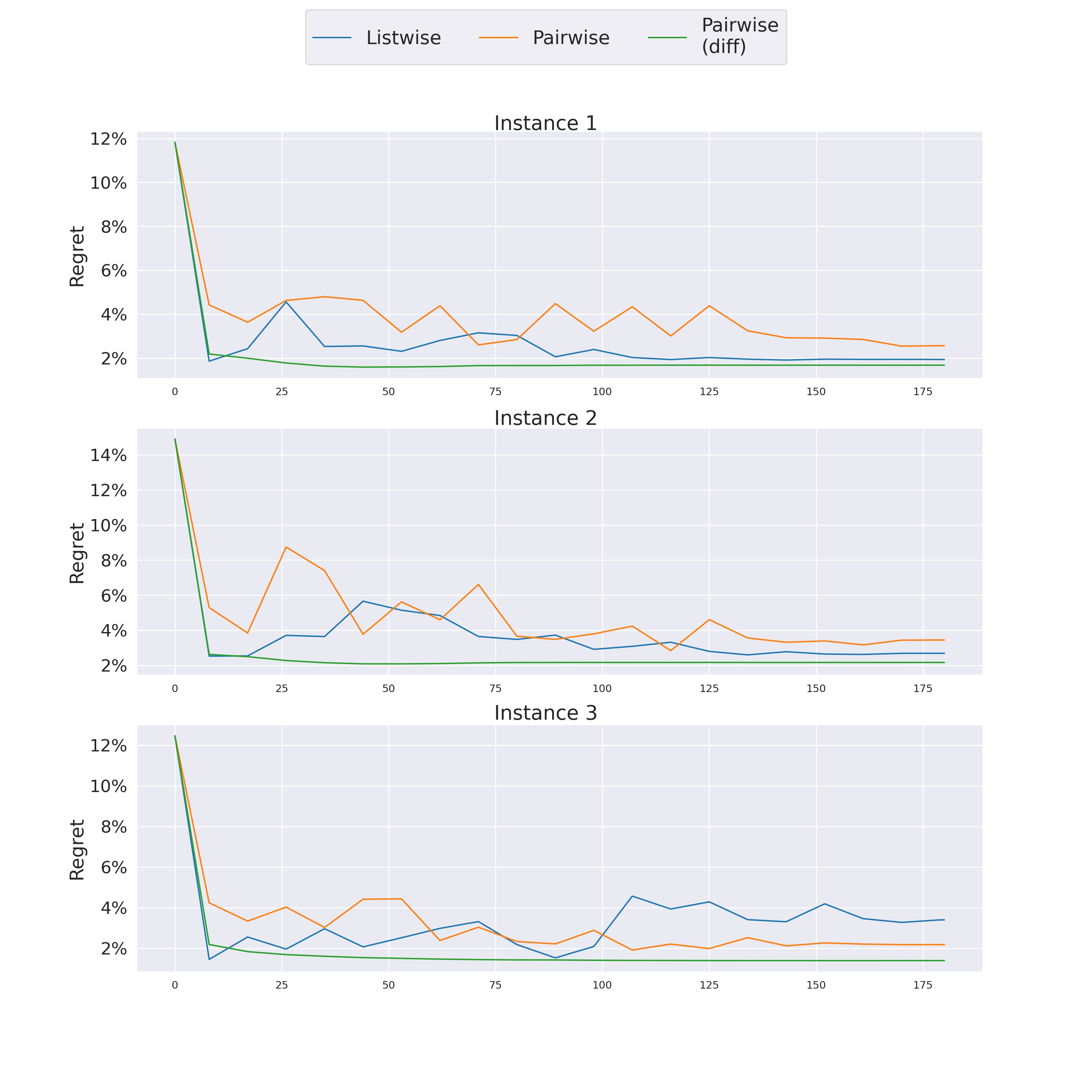}
    \caption{Learning Curves on the energy scheduling problem instances of for the LTR losses.}
    \label{fig:energy_scheduling_lrcurve}
\end{figure}

In this section, the learning curves of the LTR loss functions on the three instances of the scheduling problem are presented. Figure~\ref{fig:energy_scheduling_lrcurve} reveals that learning with Pairwise (diff) ranking loss is stable whereas learning with Listwise and Pairwise ranking losses do not stabilize.

\begin{figure}
    \centering
    \begin{subfigure}[b]{0.32\textwidth}
    \centering
    \includegraphics[width=\textwidth]{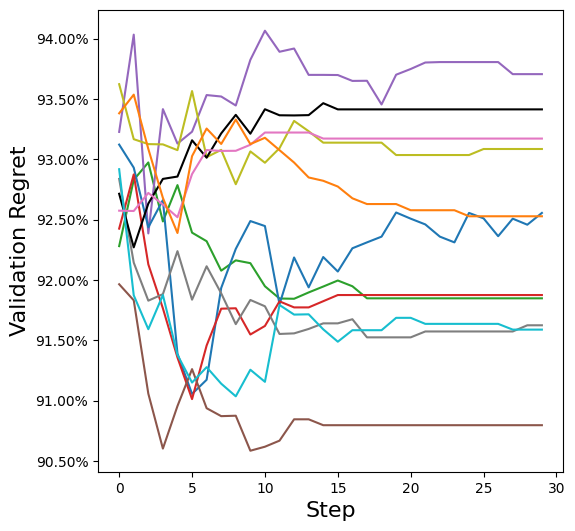}
    \caption{$\rho_1 = \rho_2 = 10\%$}
    \label{fig:matching_lr1}
  \end{subfigure}
  \begin{subfigure}[b]{0.32\textwidth}
    \centering
    \includegraphics[width=\textwidth]{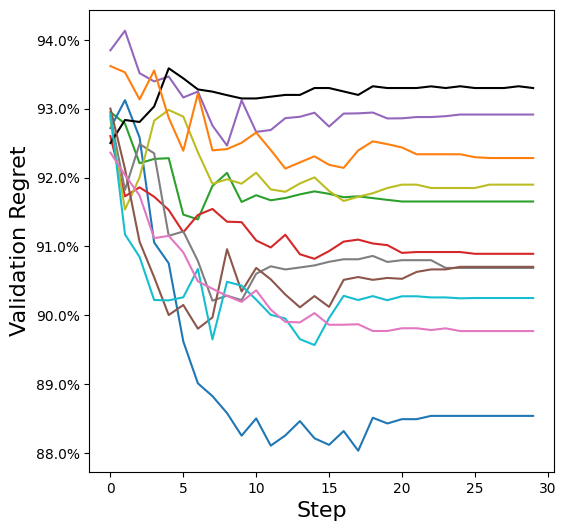}
    \caption{$\rho_1 = \rho_2 = 25\%$}
    \label{fig:matching_lr2}
  \end{subfigure}
    \begin{subfigure}[b]{0.32\textwidth}
    \centering
    \includegraphics[width=\textwidth]{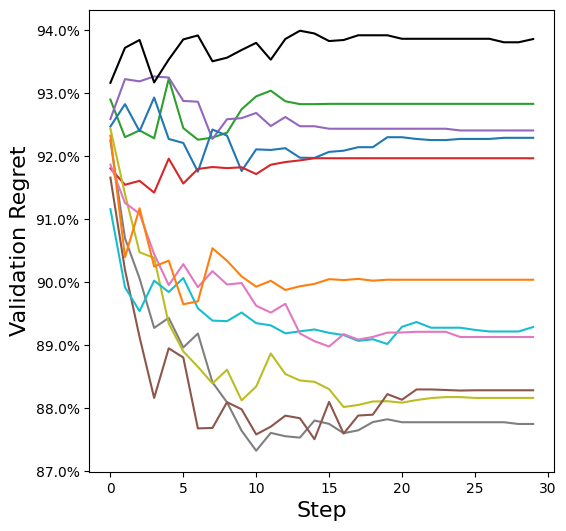}
    \caption{$\rho_1 = \rho_2 = 50\%$}
    \label{fig:matching_lr3}
  \end{subfigure}

      \begin{subfigure}[b]{0.99\textwidth}
    \centering
    \includegraphics[width=\textwidth]{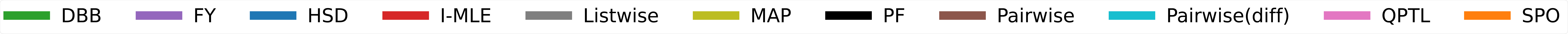}
    \caption{}
  \end{subfigure}
    \caption{Learning Curves on the diverse bipartite matching problem.}
    \label{fig:lr_matching}
\end{figure}

\section{Details about Hyperparameter Configuration}
The hyperparameters for each methodology in each experiment are selected through grid search, as described in Section \ref{sect:result}. For the sake of reproducibility lists of optimal hyperparameter combinations found by grid search and used to evaluate all the methodologies has been provided in the repository: \url{https://github.com/PredOpt/predopt-benchmarks}.

\bibliography{sample}

\begin{thebibliography}{}

\bibitem[\protect\BCAY{Abbas\ \BBA\ Swoboda}{Abbas\ \BBA\
  Swoboda}{2021}]{NEURIPS2021_83a368f5}
Abbas, A.\BBACOMMA\  \BBA\ Swoboda, P. \BBOP2021\BBCP.
\newblock \BBOQ Combinatorial optimization for panoptic segmentation: A fully
  differentiable approach\BBCQ\
\newblock In Ranzato, M., Beygelzimer, A., Dauphin, Y., Liang, P., \BBA\
  Vaughan, J.~W.\BEDS, {\Bem Advances in Neural Information Processing
  Systems}, \lowercase{\BVOL}~34, \BPGS\ 15635--15649. Curran Associates, Inc.

\bibitem[\protect\BCAY{Abernethy, Lee,\ \BBA\ Tewari}{Abernethy
  et~al.}{2016}]{abernethy2016perturbation}
Abernethy, J., Lee, C., \BBA\ Tewari, A. \BBOP2016\BBCP.
\newblock \BBOQ Perturbation techniques in online learning and
  optimization\BBCQ\
\newblock {\Bem Perturbations, Optimization, and Statistics}, {\Bem 233}.

\bibitem[\protect\BCAY{Agarap}{Agarap}{2019}]{relu2019deep}
Agarap, A.~F. \BBOP2019\BBCP.
\newblock \BBOQ Deep learning using rectified linear units (relu)\BBCQ.

\bibitem[\protect\BCAY{Agrawal, Amos, Barratt, Boyd, Diamond,\ \BBA\
  Kolter}{Agrawal et~al.}{2019a}]{agrawal2019differentiable}
Agrawal, A., Amos, B., Barratt, S., Boyd, S., Diamond, S., \BBA\ Kolter, J.~Z.
  \BBOP2019a\BBCP.
\newblock \BBOQ Differentiable convex optimization layers\BBCQ\
\newblock {\Bem Advances in neural information processing systems}, {\Bem 32}.

\bibitem[\protect\BCAY{Agrawal, Barratt, Boyd, Busseti,\ \BBA\ Moursi}{Agrawal
  et~al.}{2019b}]{agrawal2019coneprogram}
Agrawal, A., Barratt, S., Boyd, S., Busseti, E., \BBA\ Moursi, W.~M.
  \BBOP2019b\BBCP.
\newblock \BBOQ Differentiating through a cone program\BBCQ\
\newblock {\Bem Journal of Applied and Numerical Optimization}, {\Bem 1\/}(2),
  107 – 115.

\bibitem[\protect\BCAY{Amos\ \BBA\ Kolter}{Amos\ \BBA\
  Kolter}{2017}]{amos2017optnet}
Amos, B.\BBACOMMA\  \BBA\ Kolter, J.~Z. \BBOP2017\BBCP.
\newblock \BBOQ Optnet: Differentiable optimization as a layer in neural
  networks\BBCQ\
\newblock In {\Bem International Conference on Machine Learning}, \BPGS\
  136--145. PMLR.

\bibitem[\protect\BCAY{Amos, Koltun,\ \BBA\ Kolter}{Amos
  et~al.}{2019}]{amos2019limited}
Amos, B., Koltun, V., \BBA\ Kolter, J.~Z. \BBOP2019\BBCP.
\newblock \BBOQ The limited multi-label projection layer\BBCQ.

\bibitem[\protect\BCAY{Baltrušaitis, Ahuja,\ \BBA\ Morency}{Baltrušaitis
  et~al.}{2019}]{MultimodalML}
Baltrušaitis, T., Ahuja, C., \BBA\ Morency, L.-P. \BBOP2019\BBCP.
\newblock \BBOQ Multimodal machine learning: A survey and taxonomy\BBCQ\
\newblock {\Bem IEEE Transactions on Pattern Analysis and Machine
  Intelligence}, {\Bem 41\/}(2), 423--443.

\bibitem[\protect\BCAY{Barndorff-Nielsen}{Barndorff-Nielsen}{1978}]{exponentialbarndorff}
Barndorff-Nielsen, O. \BBOP1978\BBCP.
\newblock {\Bem Information and exponential families: in statistical theory}.
\newblock John Wiley \& Sons.

\bibitem[\protect\BCAY{Baydin, Pearlmutter, Radul,\ \BBA\ Siskind}{Baydin
  et~al.}{2018}]{JMLR:autodiff}
Baydin, A.~G., Pearlmutter, B.~A., Radul, A.~A., \BBA\ Siskind, J.~M.
  \BBOP2018\BBCP.
\newblock \BBOQ Automatic differentiation in machine learning: a survey\BBCQ\
\newblock {\Bem Journal of Machine Learning Research}, {\Bem 18\/}(153), 1--43.

\bibitem[\protect\BCAY{Bazaraa, Jarvis,\ \BBA\ Sherali}{Bazaraa
  et~al.}{2008}]{bazaraa2008linear}
Bazaraa, M.~S., Jarvis, J.~J., \BBA\ Sherali, H.~D. \BBOP2008\BBCP.
\newblock {\Bem Linear programming and network flows}.
\newblock John Wiley \& Sons.

\bibitem[\protect\BCAY{Bello, Pham, Le, Norouzi,\ \BBA\ Bengio}{Bello
  et~al.}{2017}]{BelloPL0B17}
Bello, I., Pham, H., Le, Q.~V., Norouzi, M., \BBA\ Bengio, S. \BBOP2017\BBCP.
\newblock \BBOQ Neural combinatorial optimization with reinforcement
  learning\BBCQ\
\newblock In {\Bem 5th International Conference on Learning Representations,
  {ICLR} 2017, Toulon, France, April 24-26, 2017, Workshop Track Proceedings}.
  OpenReview.net.

\bibitem[\protect\BCAY{Ben-Tal, El~Ghaoui,\ \BBA\ Nemirovski}{Ben-Tal
  et~al.}{2009}]{ben2009robust}
Ben-Tal, A., El~Ghaoui, L., \BBA\ Nemirovski, A. \BBOP2009\BBCP.
\newblock {\Bem Robust optimization}, \lowercase{\BVOL}~28.
\newblock Princeton university press.

\bibitem[\protect\BCAY{Berthet, Blondel, Teboul, Cuturi, Vert,\ \BBA\
  Bach}{Berthet et~al.}{2020}]{berthet2020learning}
Berthet, Q., Blondel, M., Teboul, O., Cuturi, M., Vert, J.-P., \BBA\ Bach, F.
  \BBOP2020\BBCP.
\newblock \BBOQ Learning with differentiable perturbed optimizers\BBCQ\
\newblock In Larochelle, H., Ranzato, M., Hadsell, R., Balcan, M.~F., \BBA\
  Lin, H.\BEDS, {\Bem Advances in Neural Information Processing Systems},
  \lowercase{\BVOL}~33, \BPGS\ 9508--9519.

\bibitem[\protect\BCAY{Bertsimas\ \BBA\ Kallus}{Bertsimas\ \BBA\
  Kallus}{2020}]{bertsimas2020predictive}
Bertsimas, D.\BBACOMMA\  \BBA\ Kallus, N. \BBOP2020\BBCP.
\newblock \BBOQ From predictive to prescriptive analytics\BBCQ\
\newblock {\Bem Management Science}, {\Bem 66\/}(3), 1025--1044.

\bibitem[\protect\BCAY{Blondel, Teboul, Berthet,\ \BBA\ Djolonga}{Blondel
  et~al.}{2020}]{blondel2020fast}
Blondel, M., Teboul, O., Berthet, Q., \BBA\ Djolonga, J. \BBOP2020\BBCP.
\newblock \BBOQ Fast differentiable sorting and ranking\BBCQ\
\newblock In {\Bem International Conference on Machine Learning}, \BPGS\
  950--959. PMLR.

\bibitem[\protect\BCAY{Bollob{\'a}s}{Bollob{\'a}s}{2013}]{graphmorph}
Bollob{\'a}s, B. \BBOP2013\BBCP.
\newblock {\Bem Modern graph theory}, \lowercase{\BVOL}\ 184.
\newblock Springer Science \& Business Media.

\bibitem[\protect\BCAY{Boyd\ \BBA\ Vandenberghe}{Boyd\ \BBA\
  Vandenberghe}{2014}]{boyd2004convex}
Boyd, S.~P.\BBACOMMA\  \BBA\ Vandenberghe, L. \BBOP2014\BBCP.
\newblock {\Bem Convex Optimization}.
\newblock Cambridge University Press.

\bibitem[\protect\BCAY{Bucarey, Calder{\'o}n, Mu{\~n}oz,\ \BBA\ Semet}{Bucarey
  et~al.}{2023}]{bucarey2023decision}
Bucarey, V., Calder{\'o}n, S., Mu{\~n}oz, G., \BBA\ Semet, F. \BBOP2023\BBCP.
\newblock \BBOQ Decision-focused predictions via pessimistic bilevel
  optimization: a computational study\BBCQ.

\bibitem[\protect\BCAY{Busseti, Moursi,\ \BBA\ Boyd}{Busseti
  et~al.}{2019}]{busseti2019solution}
Busseti, E., Moursi, W.~M., \BBA\ Boyd, S. \BBOP2019\BBCP.
\newblock \BBOQ Solution refinement at regular points of conic problems\BBCQ\
\newblock {\Bem Computational Optimization and Applications}, {\Bem 74\/}(3),
  627--643.

\bibitem[\protect\BCAY{Cameron, Hartford, Lundy,\ \BBA\ Leyton-Brown}{Cameron
  et~al.}{2022}]{Cameron_Hartford_Perils}
Cameron, C., Hartford, J., Lundy, T., \BBA\ Leyton-Brown, K. \BBOP2022\BBCP.
\newblock \BBOQ The perils of learning before optimizing\BBCQ\
\newblock {\Bem Proceedings of the AAAI Conference on Artificial Intelligence},
  {\Bem 36\/}(4), 3708--3715.

\bibitem[\protect\BCAY{Cao, Qin, Liu, Tsai,\ \BBA\ Li}{Cao
  et~al.}{2007}]{cao2007learning}
Cao, Z., Qin, T., Liu, T.-Y., Tsai, M.-F., \BBA\ Li, H. \BBOP2007\BBCP.
\newblock \BBOQ Learning to rank: from pairwise approach to listwise
  approach\BBCQ\
\newblock In {\Bem Proceedings of the 24th international conference on Machine
  learning}, \BPGS\ 129--136.

\bibitem[\protect\BCAY{Chai, Wong, Tong, Chen,\ \BBA\ Zhang}{Chai
  et~al.}{2022}]{chai2022port}
Chai, Z., Wong, K.-K., Tong, K.-F., Chen, Y., \BBA\ Zhang, Y. \BBOP2022\BBCP.
\newblock \BBOQ Port selection for fluid antenna systems\BBCQ\
\newblock {\Bem IEEE Communications Letters}, {\Bem 26\/}(5), 1180--1184.

\bibitem[\protect\BCAY{Chenreddy, Bandi,\ \BBA\ Delage}{Chenreddy
  et~al.}{2022}]{ChenreddyNEURIPS2022}
Chenreddy, A.~R., Bandi, N., \BBA\ Delage, E. \BBOP2022\BBCP.
\newblock \BBOQ Data-driven conditional robust optimization\BBCQ\
\newblock In Koyejo, S., Mohamed, S., Agarwal, A., Belgrave, D., Cho, K., \BBA\
  Oh, A.\BEDS, {\Bem Advances in Neural Information Processing Systems},
  \lowercase{\BVOL}~35, \BPGS\ 9525--9537. Curran Associates, Inc.

\bibitem[\protect\BCAY{Chu, Zhang, Bai,\ \BBA\ Chen}{Chu
  et~al.}{2023}]{chu2021data}
Chu, H., Zhang, W., Bai, P., \BBA\ Chen, Y. \BBOP2023\BBCP.
\newblock \BBOQ Data-driven optimization for last-mile delivery\BBCQ\
\newblock {\Bem Complex \& Intelligent Systems}, {\Bem 9\/}(3), 2271--2284.

\bibitem[\protect\BCAY{Cordonnier, Mahendran, Dosovitskiy, Weissenborn,
  Uszkoreit,\ \BBA\ Unterthiner}{Cordonnier
  et~al.}{2021}]{Cordonnier_2021_CVPR}
Cordonnier, J.-B., Mahendran, A., Dosovitskiy, A., Weissenborn, D., Uszkoreit,
  J., \BBA\ Unterthiner, T. \BBOP2021\BBCP.
\newblock \BBOQ Differentiable patch selection for image recognition\BBCQ\
\newblock In {\Bem Proceedings of the IEEE/CVF Conference on Computer Vision
  and Pattern Recognition (CVPR)}, \BPGS\ 2351--2360.

\bibitem[\protect\BCAY{Dalle, Baty, Bouvier,\ \BBA\ Parmentier}{Dalle
  et~al.}{2022}]{dalle2022learning}
Dalle, G., Baty, L., Bouvier, L., \BBA\ Parmentier, A. \BBOP2022\BBCP.
\newblock \BBOQ Learning with combinatorial optimization layers: a
  probabilistic approach\BBCQ.

\bibitem[\protect\BCAY{Delage\ \BBA\ Ye}{Delage\ \BBA\
  Ye}{2010}]{delage2010distributionally}
Delage, E.\BBACOMMA\  \BBA\ Ye, Y. \BBOP2010\BBCP.
\newblock \BBOQ Distributionally robust optimization under moment uncertainty
  with application to data-driven problems\BBCQ\
\newblock {\Bem Operations research}, {\Bem 58\/}(3), 595--612.

\bibitem[\protect\BCAY{Demirovi{\'c}, Stuckey, Bailey, Chan, Leckie,
  Ramamohanarao,\ \BBA\ Guns}{Demirovi{\'c}
  et~al.}{2019}]{demirovic2019investigation}
Demirovi{\'c}, E., Stuckey, P.~J., Bailey, J., Chan, J., Leckie, C.,
  Ramamohanarao, K., \BBA\ Guns, T. \BBOP2019\BBCP.
\newblock \BBOQ An investigation into prediction+ optimisation for the knapsack
  problem\BBCQ\
\newblock In {\Bem Integration of Constraint Programming, Artificial
  Intelligence, and Operations Research: 16th International Conference, CPAIOR
  2019, Thessaloniki, Greece, June 4--7, 2019, Proceedings 16}, \BPGS\
  241--257. Springer.

\bibitem[\protect\BCAY{Demirovi{\'c}, Stuckey, Guns, Bailey, Leckie,
  Ramamohanarao,\ \BBA\ Chan}{Demirovi{\'c} et~al.}{2020}]{Dynamic_2020}
Demirovi{\'c}, E., Stuckey, P.~J., Guns, T., Bailey, J., Leckie, C.,
  Ramamohanarao, K., \BBA\ Chan, J. \BBOP2020\BBCP.
\newblock \BBOQ Dynamic programming for predict+optimise\BBCQ\
\newblock {\Bem Proceedings of the AAAI Conference on Artificial Intelligence},
  {\Bem 34\/}(02), 1444--1451.

\bibitem[\protect\BCAY{den Hertog\ \BBA\ Postek}{den Hertog\ \BBA\
  Postek}{2016}]{den2016bridging}
den Hertog, D.\BBACOMMA\  \BBA\ Postek, K. \BBOP2016\BBCP.
\newblock \BBOQ Bridging the gap between predictive and prescriptive
  analytics-new optimization methodology needed\BBCQ\
\newblock In {\Bem Tilburg Univ, Tilburg, The Netherlands}.

\bibitem[\protect\BCAY{Dijkstra}{Dijkstra}{1959}]{dijkstra1959note}
Dijkstra, E. \BBOP1959\BBCP.
\newblock \BBOQ A note on two problems in connexion with graphs\BBCQ\
\newblock {\Bem Numerische Mathematik}, {\Bem 1\/}(1), 269--271.

\bibitem[\protect\BCAY{Dinh, Kotary,\ \BBA\ Fioretto}{Dinh
  et~al.}{2024}]{dinh2024end}
Dinh, M.~H., Kotary, J., \BBA\ Fioretto, F. \BBOP2024\BBCP.
\newblock \BBOQ End-to-end learning for fair multiobjective optimization under
  uncertainty\BBCQ.

\bibitem[\protect\BCAY{Djolonga\ \BBA\ Krause}{Djolonga\ \BBA\
  Krause}{2017}]{NIPS2017Djolonga}
Djolonga, J.\BBACOMMA\  \BBA\ Krause, A. \BBOP2017\BBCP.
\newblock \BBOQ Differentiable learning of submodular models\BBCQ\
\newblock In Guyon, I., Luxburg, U.~V., Bengio, S., Wallach, H., Fergus, R.,
  Vishwanathan, S., \BBA\ Garnett, R.\BEDS, {\Bem Advances in Neural
  Information Processing Systems}, \lowercase{\BVOL}~30. Curran Associates,
  Inc.

\bibitem[\protect\BCAY{Domke}{Domke}{2010}]{Domke10}
Domke, J. \BBOP2010\BBCP.
\newblock \BBOQ Implicit differentiation by perturbation\BBCQ\
\newblock In Lafferty, J., Williams, C., Shawe-Taylor, J., Zemel, R., \BBA\
  Culotta, A.\BEDS, {\Bem Advances in Neural Information Processing Systems},
  \lowercase{\BVOL}~23. Curran Associates, Inc.

\bibitem[\protect\BCAY{Domke}{Domke}{2012}]{domke2012generic}
Domke, J. \BBOP2012\BBCP.
\newblock \BBOQ Generic methods for optimization-based modeling\BBCQ\
\newblock In {\Bem Artificial Intelligence and Statistics}, \BPGS\ 318--326.
  PMLR.

\bibitem[\protect\BCAY{Donti, Kolter,\ \BBA\ Amos}{Donti
  et~al.}{2017}]{DontiKA17}
Donti, P.~L., Kolter, J.~Z., \BBA\ Amos, B. \BBOP2017\BBCP.
\newblock \BBOQ Task-based end-to-end model learning in stochastic
  optimization\BBCQ\
\newblock In Guyon, I., von Luxburg, U., Bengio, S., Wallach, H.~M., Fergus,
  R., Vishwanathan, S. V.~N., \BBA\ Garnett, R.\BEDS, {\Bem Advances in Neural
  Information Processing Systems 30: Annual Conference on Neural Information
  Processing Systems 2017, December 4-9, 2017, Long Beach, CA, {USA}}, \BPGS\
  5484--5494.

\bibitem[\protect\BCAY{El~Balghiti, Elmachtoub, Grigas,\ \BBA\
  Tewari}{El~Balghiti et~al.}{2019}]{BalghitiEGT19}
El~Balghiti, O., Elmachtoub, A.~N., Grigas, P., \BBA\ Tewari, A.
  \BBOP2019\BBCP.
\newblock \BBOQ Generalization bounds in the predict-then-optimize
  framework\BBCQ\
\newblock In Wallach, H., Larochelle, H., Beygelzimer, A., d\textquotesingle
  Alch\'{e}-Buc, F., Fox, E., \BBA\ Garnett, R.\BEDS, {\Bem Advances in Neural
  Information Processing Systems}, \lowercase{\BVOL}~32. Curran Associates,
  Inc.

\bibitem[\protect\BCAY{Elmachtoub, Liang,\ \BBA\ McNellis}{Elmachtoub
  et~al.}{2020}]{elmachtoub2020decisiontree}
Elmachtoub, A., Liang, J. C.~N., \BBA\ McNellis, R. \BBOP2020\BBCP.
\newblock \BBOQ Decision trees for decision-making under the
  predict-then-optimize framework\BBCQ\
\newblock In {\Bem International Conference on Machine Learning}, \BPGS\
  2858--2867. PMLR.

\bibitem[\protect\BCAY{Elmachtoub\ \BBA\ Grigas}{Elmachtoub\ \BBA\
  Grigas}{2022}]{elmachtoub2022smart}
Elmachtoub, A.~N.\BBACOMMA\  \BBA\ Grigas, P. \BBOP2022\BBCP.
\newblock \BBOQ Smart “predict, then optimize”\BBCQ\
\newblock {\Bem Management Science}, {\Bem 68\/}(1), 9--26.

\bibitem[\protect\BCAY{Elmachtoub, Lam, Zhang,\ \BBA\ Zhao}{Elmachtoub
  et~al.}{2023}]{ETOvsIEO}
Elmachtoub, A.~N., Lam, H., Zhang, H., \BBA\ Zhao, Y. \BBOP2023\BBCP.
\newblock \BBOQ Estimate-then-optimize versus
  integrated-estimation-optimization: A stochastic dominance perspective\BBCQ.

\bibitem[\protect\BCAY{Falcon et~al.}{Falcon et~al.}{2019}]{falcon2019pytorch}
Falcon, W.\BBACOMMA\  et~al. \BBOP2019\BBCP.
\newblock \BBOQ Pytorch lightning\BBCQ\
\newblock {\Bem GitHub. Note: https://github.
  com/PyTorchLightning/pytorch-lightning}, {\Bem 3\/}(6), 11.

\bibitem[\protect\BCAY{Ferber, Griffin, Dilkina, Keskin,\ \BBA\ Gore}{Ferber
  et~al.}{2023a}]{ferber2023predicting}
Ferber, A., Griffin, E., Dilkina, B., Keskin, B., \BBA\ Gore, M.
  \BBOP2023a\BBCP.
\newblock \BBOQ Predicting wildlife trafficking routes with differentiable
  shortest paths\BBCQ\
\newblock In {\Bem Proceedings of the Integration of Constraint Programming,
  Artificial Intelligence, and Operations Research: 20th International
  Conference, CPAIOR 2023}.

\bibitem[\protect\BCAY{Ferber, Huang, Zha, Schubert, Steiner, Dilkina,\ \BBA\
  Tian}{Ferber et~al.}{2023b}]{ferber2022surco}
Ferber, A.~M., Huang, T., Zha, D., Schubert, M., Steiner, B., Dilkina, B.,
  \BBA\ Tian, Y. \BBOP2023b\BBCP.
\newblock \BBOQ Surco: Learning linear surrogates for combinatorial nonlinear
  optimization problems\BBCQ\
\newblock In Krause, A., Brunskill, E., Cho, K., Engelhardt, B., Sabato, S.,
  \BBA\ Scarlett, J.\BEDS, {\Bem International Conference on Machine Learning,
  {ICML} 2023, 23-29 July 2023, Honolulu, Hawaii, {USA}}, \lowercase{\BVOL}\
  202 of {\Bem Proceedings of Machine Learning Research}, \BPGS\ 10034--10052.
  {PMLR}.

\bibitem[\protect\BCAY{Ferber, Wilder, Dilkina,\ \BBA\ Tambe}{Ferber
  et~al.}{2020}]{aaaiFerberWDT20}
Ferber, A.~M., Wilder, B., Dilkina, B., \BBA\ Tambe, M. \BBOP2020\BBCP.
\newblock \BBOQ Mipaal: Mixed integer program as a layer\BBCQ\
\newblock In {\Bem The Thirty-Fourth {AAAI} Conference on Artificial
  Intelligence, {AAAI} 2020, The Thirty-Second Innovative Applications of
  Artificial Intelligence Conference, {IAAI} 2020, The Tenth {AAAI} Symposium
  on Educational Advances in Artificial Intelligence, {EAAI} 2020, New York,
  NY, USA, February 7-12, 2020}, \BPGS\ 1504--1511. {AAAI} Press.

\bibitem[\protect\BCAY{Garcia, Street, Homem-de Mello,\ \BBA\ Mu{\~n}oz}{Garcia
  et~al.}{2021}]{garcia2022applicationdriven}
Garcia, J.~D., Street, A., Homem-de Mello, T., \BBA\ Mu{\~n}oz, F.~D.
  \BBOP2021\BBCP.
\newblock \BBOQ Application-driven learning: A closed-loop prediction and
  optimization approach applied to dynamic reserves and demand
  forecasting\BBCQ.

\bibitem[\protect\BCAY{Garlappi, Uppal,\ \BBA\ Wang}{Garlappi
  et~al.}{2006}]{hhl003}
Garlappi, L., Uppal, R., \BBA\ Wang, T. \BBOP2006\BBCP.
\newblock \BBOQ {Portfolio Selection with Parameter and Model Uncertainty: A
  Multi-Prior Approach}\BBCQ\
\newblock {\Bem The Review of Financial Studies}, {\Bem 20\/}(1), 41--81.

\bibitem[\protect\BCAY{Gillick, Kulkarni, Lansing, Presta, Baldridge, Ie,\
  \BBA\ Garcia-Olano}{Gillick et~al.}{2019}]{gillick-etal-2019-learning}
Gillick, D., Kulkarni, S., Lansing, L., Presta, A., Baldridge, J., Ie, E.,
  \BBA\ Garcia-Olano, D. \BBOP2019\BBCP.
\newblock \BBOQ Learning dense representations for entity retrieval\BBCQ\
\newblock In {\Bem Proceedings of the 23rd Conference on Computational Natural
  Language Learning (CoNLL)}, \BPGS\ 528--537, Hong Kong, China. Association
  for Computational Linguistics.

\bibitem[\protect\BCAY{Gomes, Kautz, Sabharwal,\ \BBA\ Selman}{Gomes
  et~al.}{2008}]{gomes2008satisfiability}
Gomes, C.~P., Kautz, H., Sabharwal, A., \BBA\ Selman, B. \BBOP2008\BBCP.
\newblock \BBOQ Satisfiability solvers\BBCQ\
\newblock {\Bem Foundations of Artificial Intelligence}, {\Bem 3}, 89--134.

\bibitem[\protect\BCAY{Goodfellow}{Goodfellow}{2015}]{goodfellow2014distinguishability}
Goodfellow, I.~J. \BBOP2015\BBCP.
\newblock \BBOQ On distinguishability criteria for estimating generative
  models\BBCQ\
\newblock In {\Bem Proceedings of ICLR}.

\bibitem[\protect\BCAY{Goodfellow, Shlens,\ \BBA\ Szegedy}{Goodfellow
  et~al.}{2015}]{goodfellow2014explaining}
Goodfellow, I.~J., Shlens, J., \BBA\ Szegedy, C. \BBOP2015\BBCP.
\newblock \BBOQ Explaining and harnessing adversarial examples\BBCQ\
\newblock In Bengio, Y.\BBACOMMA\  \BBA\ LeCun, Y.\BEDS, {\Bem 3rd
  International Conference on Learning Representations, {ICLR} 2015, San Diego,
  CA, USA, May 7-9, 2015, Conference Track Proceedings}.

\bibitem[\protect\BCAY{Gould, Fernando, Cherian, Anderson, Cruz,\ \BBA\
  Guo}{Gould et~al.}{2016}]{gould2016differentiating}
Gould, S., Fernando, B., Cherian, A., Anderson, P., Cruz, R.~S., \BBA\ Guo, E.
  \BBOP2016\BBCP.
\newblock \BBOQ On differentiating parameterized argmin and argmax problems
  with application to bi-level optimization\BBCQ.

\bibitem[\protect\BCAY{Goyal\ \BBA\ Gunasekaran}{Goyal\ \BBA\
  Gunasekaran}{1990}]{multigoyal1990}
Goyal, S.~K.\BBACOMMA\  \BBA\ Gunasekaran, A. \BBOP1990\BBCP.
\newblock \BBOQ Multi-stage production-inventory systems\BBCQ\
\newblock {\Bem European Journal of Operational Research}, {\Bem 46\/}(1),
  1--20.

\bibitem[\protect\BCAY{Grant\ \BBA\ Boyd}{Grant\ \BBA\
  Boyd}{2008}]{grant2008graph}
Grant, M.~C.\BBACOMMA\  \BBA\ Boyd, S.~P. \BBOP2008\BBCP.
\newblock \BBOQ Graph implementations for nonsmooth convex programs\BBCQ\
\newblock In {\Bem Recent advances in learning and control}, \BPGS\ 95--110.
  Springer.

\bibitem[\protect\BCAY{Guler, Demirovi{\'c}, Chan, Bailey, Leckie,\ \BBA\
  Stuckey}{Guler et~al.}{2022}]{Divide_and_Conquer_Guler}
Guler, A.~U., Demirovi{\'c}, E., Chan, J., Bailey, J., Leckie, C., \BBA\
  Stuckey, P.~J. \BBOP2022\BBCP.
\newblock \BBOQ A divide and conquer algorithm for predict+optimize with
  non-convex problems\BBCQ\
\newblock {\Bem Proceedings of the AAAI Conference on Artificial Intelligence},
  {\Bem 36\/}(4), 3749--3757.

\bibitem[\protect\BCAY{Gurobi~Optimization}{Gurobi~Optimization}{2021}]{gurobi}
Gurobi~Optimization, L. \BBOP2021\BBCP.
\newblock \BBOQ Gurobi optimizer reference manual\BBCQ\
\newblock \url{http://www.gurobi.com}.

\bibitem[\protect\BCAY{Gutmann\ \BBA\ Hyv{\"a}rinen}{Gutmann\ \BBA\
  Hyv{\"a}rinen}{2010}]{gutmann2010noise}
Gutmann, M.\BBACOMMA\  \BBA\ Hyv{\"a}rinen, A. \BBOP2010\BBCP.
\newblock \BBOQ Noise-contrastive estimation: A new estimation principle for
  unnormalized statistical models\BBCQ\
\newblock In {\Bem Proceedings of the Thirteenth International Conference on
  Artificial Intelligence and Statistics}, \BPGS\ 297--304.

\bibitem[\protect\BCAY{Guyomarch}{Guyomarch}{2017}]{warcraft}
Guyomarch, J. \BBOP2017\BBCP.
\newblock \BBOQ Warcraft ii open-source map editor\BBCQ\
\newblock http://github.com/war2/war2edit.

\bibitem[\protect\BCAY{He, Zhang, Ren,\ \BBA\ Sun}{He
  et~al.}{2016}]{Resnet_CVPR}
He, K., Zhang, X., Ren, S., \BBA\ Sun, J. \BBOP2016\BBCP.
\newblock \BBOQ Deep residual learning for image recognition\BBCQ\
\newblock In {\Bem Proceedings of the IEEE Conference on Computer Vision and
  Pattern Recognition (CVPR)}.

\bibitem[\protect\BCAY{Hinton, Vinyals,\ \BBA\ Dean}{Hinton
  et~al.}{2015}]{hinton2015distilling}
Hinton, G., Vinyals, O., \BBA\ Dean, J. \BBOP2015\BBCP.
\newblock \BBOQ Distilling the knowledge in a neural network\BBCQ.

\bibitem[\protect\BCAY{Hu, Wang,\ \BBA\ Gooi}{Hu et~al.}{2016}]{hu2016toward}
Hu, W., Wang, P., \BBA\ Gooi, H.~B. \BBOP2016\BBCP.
\newblock \BBOQ Toward optimal energy management of microgrids via robust
  two-stage optimization\BBCQ\
\newblock {\Bem IEEE Transactions on smart grid}, {\Bem 9\/}(2), 1161--1174.

\bibitem[\protect\BCAY{Hu, Lee,\ \BBA\ Lee}{Hu et~al.}{2023a}]{hu2023branch}
Hu, X., Lee, J. C.~H., \BBA\ Lee, J. H.~M. \BBOP2023a\BBCP.
\newblock \BBOQ Branch {\&} learn with post-hoc correction for predict+optimize
  with unknown parameters in constraints\BBCQ\
\newblock In {\Bem Integration of Constraint Programming, Artificial
  Intelligence, and Operations Research}, \BPGS\ 264--280. Springer Nature
  Switzerland.

\bibitem[\protect\BCAY{Hu, Lee,\ \BBA\ Lee}{Hu et~al.}{2023b}]{nips/HuLL23}
Hu, X., Lee, J. C.~H., \BBA\ Lee, J.~H. \BBOP2023b\BBCP.
\newblock \BBOQ Two-stage predict+optimize for milps with unknown parameters in
  constraints\BBCQ\
\newblock In Oh, A., Naumann, T., Globerson, A., Saenko, K., Hardt, M., \BBA\
  Levine, S.\BEDS, {\Bem Advances in Neural Information Processing Systems 36:
  Annual Conference on Neural Information Processing Systems 2023, NeurIPS
  2023, New Orleans, LA, USA, December 10 - 16, 2023}.

\bibitem[\protect\BCAY{Hu, Lee,\ \BBA\ Lee}{Hu
  et~al.}{2023c}]{PredictOptimizeforPacking}
Hu, X., Lee, J.~C., \BBA\ Lee, J.~H. \BBOP2023c\BBCP.
\newblock \BBOQ Predict+optimize for packing and covering lps with unknown
  parameters in constraints\BBCQ\
\newblock {\Bem Proceedings of the AAAI Conference on Artificial Intelligence},
  {\Bem 37\/}(4), 3987--3995.

\bibitem[\protect\BCAY{Hu, Lee, Lee,\ \BBA\ Zhong}{Hu
  et~al.}{2022}]{hu2022branch}
Hu, X., Lee, J.~C., Lee, J.~H., \BBA\ Zhong, A.~Z. \BBOP2022\BBCP.
\newblock \BBOQ Branch \& learn for recursively and iteratively solvable
  problems in predict+optimize\BBCQ\
\newblock In Oh, A.~H., Agarwal, A., Belgrave, D., \BBA\ Cho, K.\BEDS, {\Bem
  Advances in Neural Information Processing Systems}.

\bibitem[\protect\BCAY{Huang, He, Gao, Deng, Acero,\ \BBA\ Heck}{Huang
  et~al.}{2013}]{cikm/HuangHGDAH13}
Huang, P., He, X., Gao, J., Deng, L., Acero, A., \BBA\ Heck, L.~P.
  \BBOP2013\BBCP.
\newblock \BBOQ Learning deep structured semantic models for web search using
  clickthrough data\BBCQ\
\newblock In He, Q., Iyengar, A., Nejdl, W., Pei, J., \BBA\ Rastogi, R.\BEDS,
  {\Bem 22nd {ACM} International Conference on Information and Knowledge
  Management, CIKM'13, San Francisco, CA, USA, October 27 - November 1, 2013},
  \BPGS\ 2333--2338. {ACM}.

\bibitem[\protect\BCAY{Ifrim, O’Sullivan,\ \BBA\ Simonis}{Ifrim
  et~al.}{2012}]{ifrim2012properties}
Ifrim, G., O’Sullivan, B., \BBA\ Simonis, H. \BBOP2012\BBCP.
\newblock \BBOQ Properties of energy-price forecasts for scheduling\BBCQ\
\newblock In {\Bem International Conference on Principles and Practice of
  Constraint Programming}, \BPGS\ 957--972. Springer.

\bibitem[\protect\BCAY{Ignizio\ \BBA\ Cavalier}{Ignizio\ \BBA\
  Cavalier}{1994}]{ignizio1994linear}
Ignizio, J.~P.\BBACOMMA\  \BBA\ Cavalier, T.~M. \BBOP1994\BBCP.
\newblock {\Bem Linear programming}.
\newblock Prentice-Hall, Inc.

\bibitem[\protect\BCAY{Jeong, Jaggi, Butler,\ \BBA\ Sanner}{Jeong
  et~al.}{2022}]{jeong22a}
Jeong, J., Jaggi, P., Butler, A., \BBA\ Sanner, S. \BBOP2022\BBCP.
\newblock \BBOQ An exact symbolic reduction of linear smart
  {P}redict+{O}ptimize to mixed integer linear programming\BBCQ\
\newblock In {\Bem Proceedings of the 39th International Conference on Machine
  Learning}, \lowercase{\BVOL}\ 162 of {\Bem Proceedings of Machine Learning
  Research}, \BPGS\ 10053--10067. PMLR.

\bibitem[\protect\BCAY{Joachims}{Joachims}{2002}]{Joachims02}
Joachims, T. \BBOP2002\BBCP.
\newblock \BBOQ Optimizing search engines using clickthrough data\BBCQ\
\newblock In {\Bem Proceedings of the Eighth {ACM} {SIGKDD} International
  Conference on Knowledge Discovery and Data Mining, July 23-26, 2002,
  Edmonton, Alberta, Canada}, \BPGS\ 133--142. {ACM}.

\bibitem[\protect\BCAY{Johnson-Yu, Wang, Finocchiaro, Taneja,\ \BBA\
  Tambe}{Johnson-Yu et~al.}{2023}]{yu2023stackel}
Johnson-Yu, S., Wang, K., Finocchiaro, J., Taneja, A., \BBA\ Tambe, M.
  \BBOP2023\BBCP.
\newblock \BBOQ Modeling robustness in decision-focused learning as a
  stackelberg game\BBCQ\
\newblock In {\Bem The 22nd International Conference on Autonomous Agents and
  Multiagent Systems}.

\bibitem[\protect\BCAY{Kainmueller, Jug, Rother,\ \BBA\ Myers}{Kainmueller
  et~al.}{2014}]{miccai/KainmuellerJRM14}
Kainmueller, D., Jug, F., Rother, C., \BBA\ Myers, G. \BBOP2014\BBCP.
\newblock \BBOQ Active graph matching for automatic joint segmentation and
  annotation of c. elegans\BBCQ\
\newblock In Golland, P., Hata, N., Barillot, C., Hornegger, J., \BBA\ Howe,
  R.~D.\BEDS, {\Bem Medical Image Computing and Computer-Assisted Intervention
  - {MICCAI} 2014 - 17th International Conference, Boston, MA, USA, September
  14-18, 2014, Proceedings, Part {I}}, \lowercase{\BVOL}\ 8673 of {\Bem Lecture
  Notes in Computer Science}, \BPGS\ 81--88. Springer.

\bibitem[\protect\BCAY{Kim, Lewis,\ \BBA\ White}{Kim
  et~al.}{2005}]{kim2005optimal}
Kim, S., Lewis, M.~E., \BBA\ White, C.~C. \BBOP2005\BBCP.
\newblock \BBOQ Optimal vehicle routing with real-time traffic
  information\BBCQ\
\newblock {\Bem IEEE Transactions on Intelligent Transportation Systems}, {\Bem
  6\/}(2), 178--188.

\bibitem[\protect\BCAY{Kingma\ \BBA\ Ba}{Kingma\ \BBA\
  Ba}{2015}]{Adamarxiv.1412.6980}
Kingma, D.~P.\BBACOMMA\  \BBA\ Ba, J. \BBOP2015\BBCP.
\newblock \BBOQ Adam: {A} method for stochastic optimization\BBCQ\
\newblock In Bengio, Y.\BBACOMMA\  \BBA\ LeCun, Y.\BEDS, {\Bem 3rd
  International Conference on Learning Representations, {ICLR} 2015, San Diego,
  CA, USA, May 7-9, 2015, Conference Track Proceedings}.

\bibitem[\protect\BCAY{Kingma\ \BBA\ Welling}{Kingma\ \BBA\
  Welling}{2014}]{KingmaW13}
Kingma, D.~P.\BBACOMMA\  \BBA\ Welling, M. \BBOP2014\BBCP.
\newblock \BBOQ Auto-encoding variational bayes\BBCQ\
\newblock In Bengio, Y.\BBACOMMA\  \BBA\ LeCun, Y.\BEDS, {\Bem 2nd
  International Conference on Learning Representations, {ICLR} 2014, Banff, AB,
  Canada, April 14-16, 2014, Conference Track Proceedings}.

\bibitem[\protect\BCAY{Kinsey, Tuck, Sinha,\ \BBA\ Nguyen}{Kinsey
  et~al.}{2023}]{kinsey2023exploration}
Kinsey, S.~E., Tuck, W.~W., Sinha, A., \BBA\ Nguyen, T.~H. \BBOP2023\BBCP.
\newblock \BBOQ An exploration of poisoning attacks on data-based decision
  making\BBCQ\
\newblock In {\Bem Decision and Game Theory for Security: 13th International
  Conference, GameSec 2022, Pittsburgh, PA, USA, October 26--28, 2022,
  Proceedings}, \BPGS\ 231--252. Springer.

\bibitem[\protect\BCAY{Koller\ \BBA\ Friedman}{Koller\ \BBA\
  Friedman}{2009}]{koller2009probabilistic}
Koller, D.\BBACOMMA\  \BBA\ Friedman, N. \BBOP2009\BBCP.
\newblock {\Bem Probabilistic graphical models: principles and techniques}.
\newblock MIT press.

\bibitem[\protect\BCAY{Konishi\ \BBA\ Fukunaga}{Konishi\ \BBA\
  Fukunaga}{2021}]{TakuyaGradientBoosting}
Konishi, T.\BBACOMMA\  \BBA\ Fukunaga, T. \BBOP2021\BBCP.
\newblock \BBOQ End-to-end learning for prediction and optimization with
  gradient boosting\BBCQ\
\newblock In Hutter, F., Kersting, K., Lijffijt, J., \BBA\ Valera, I.\BEDS,
  {\Bem Machine Learning and Knowledge Discovery in Databases}, \BPGS\
  191--207, Cham. Springer International Publishing.

\bibitem[\protect\BCAY{Kotary, Di~Vito, Christopher, Van~Hentenryck,\ \BBA\
  Fioretto}{Kotary et~al.}{2023a}]{kotary2023predict}
Kotary, J., Di~Vito, V., Christopher, J., Van~Hentenryck, P., \BBA\ Fioretto,
  F. \BBOP2023a\BBCP.
\newblock \BBOQ Predict-then-optimize by proxy: Learning joint models of
  prediction and optimization\BBCQ.

\bibitem[\protect\BCAY{Kotary, Dinh,\ \BBA\ Fioretto}{Kotary
  et~al.}{2023b}]{kotary2023folded}
Kotary, J., Dinh, M.~H., \BBA\ Fioretto, F. \BBOP2023b\BBCP.
\newblock \BBOQ Backpropagation of unrolled solvers with folded
  optimization\BBCQ\
\newblock In Elkind, E.\BED, {\Bem Proceedings of the Thirty-Second
  International Joint Conference on Artificial Intelligence, {IJCAI-23}},
  \BPGS\ 1963--1970. International Joint Conferences on Artificial Intelligence
  Organization.

\bibitem[\protect\BCAY{Kotary, Fioretto, Van~Hentenryck,\ \BBA\ Wilder}{Kotary
  et~al.}{2021}]{kotary2021end}
Kotary, J., Fioretto, F., Van~Hentenryck, P., \BBA\ Wilder, B. \BBOP2021\BBCP.
\newblock \BBOQ End-to-end constrained optimization learning: A survey\BBCQ\
\newblock In Zhou, Z.-H.\BED, {\Bem Proceedings of the Thirtieth International
  Joint Conference on Artificial Intelligence, {IJCAI-21}}, \BPGS\ 4475--4482.
  International Joint Conferences on Artificial Intelligence Organization.
\newblock Survey Track.

\bibitem[\protect\BCAY{Kotary, Fioretto, Van~Hentenryck,\ \BBA\ Zhu}{Kotary
  et~al.}{2022}]{kotary2022end}
Kotary, J., Fioretto, F., Van~Hentenryck, P., \BBA\ Zhu, Z. \BBOP2022\BBCP.
\newblock \BBOQ End-to-end learning for fair ranking systems\BBCQ\
\newblock In {\Bem Proceedings of the ACM Web Conference 2022}, \BPGS\
  3520--3530.

\bibitem[\protect\BCAY{Lakhera, Shanbhag,\ \BBA\ McInerney}{Lakhera
  et~al.}{2011}]{lakhera2011approximating}
Lakhera, S., Shanbhag, U.~V., \BBA\ McInerney, M.~K. \BBOP2011\BBCP.
\newblock \BBOQ Approximating electrical distribution networks via
  mixed-integer nonlinear programming\BBCQ\
\newblock {\Bem International Journal of Electrical Power \& Energy Systems},
  {\Bem 33\/}(2), 245--257.

\bibitem[\protect\BCAY{Levy, Carmon, Duchi,\ \BBA\ Sidford}{Levy
  et~al.}{2020}]{levy2020dro}
Levy, D., Carmon, Y., Duchi, J.~C., \BBA\ Sidford, A. \BBOP2020\BBCP.
\newblock \BBOQ Large-scale methods for distributionally robust
  optimization\BBCQ\
\newblock {\Bem Advances in Neural Information Processing Systems}, {\Bem 33},
  8847--8860.

\bibitem[\protect\BCAY{Liu\ \BBA\ Liu}{Liu\ \BBA\ Liu}{2009}]{liu2009theory}
Liu, B.\BBACOMMA\  \BBA\ Liu, B. \BBOP2009\BBCP.
\newblock {\Bem Theory and practice of uncertain programming},
  \lowercase{\BVOL}\ 239.
\newblock Springer.

\bibitem[\protect\BCAY{Liu\ \BBA\ Grigas}{Liu\ \BBA\
  Grigas}{2021}]{liu2021risk}
Liu, H.\BBACOMMA\  \BBA\ Grigas, P. \BBOP2021\BBCP.
\newblock \BBOQ Risk bounds and calibration for a smart predict-then-optimize
  method\BBCQ\
\newblock In Ranzato, M., Beygelzimer, A., Dauphin, Y., Liang, P., \BBA\
  Vaughan, J.~W.\BEDS, {\Bem Advances in Neural Information Processing
  Systems}, \lowercase{\BVOL}~34, \BPGS\ 22083--22094. Curran Associates, Inc.

\bibitem[\protect\BCAY{Liu, Grigas, Liu,\ \BBA\ Shen}{Liu
  et~al.}{2023}]{liu2023active}
Liu, M., Grigas, P., Liu, H., \BBA\ Shen, Z.-J.~M. \BBOP2023\BBCP.
\newblock \BBOQ Active learning in the predict-then-optimize framework: A
  margin-based approach\BBCQ.

\bibitem[\protect\BCAY{Mandi, Bucarey, Tchomba,\ \BBA\ Guns}{Mandi
  et~al.}{2022}]{mandi22-pmlr}
Mandi, J., Bucarey, V., Tchomba, M. M.~K., \BBA\ Guns, T. \BBOP2022\BBCP.
\newblock \BBOQ Decision-focused learning: Through the lens of learning to
  rank\BBCQ\
\newblock In {\Bem Proceedings of the 39th International Conference on Machine
  Learning}, \lowercase{\BVOL}\ 162 of {\Bem Proceedings of Machine Learning
  Research}, \BPGS\ 14935--14947. PMLR.

\bibitem[\protect\BCAY{Mandi, Canoy, Bucarey,\ \BBA\ Guns}{Mandi
  et~al.}{2021}]{CP.2021.42}
Mandi, J., Canoy, R., Bucarey, V., \BBA\ Guns, T. \BBOP2021\BBCP.
\newblock \BBOQ {Data Driven VRP: A Neural Network Model to Learn Hidden
  Preferences for VRP}\BBCQ\
\newblock In Michel, L.~D.\BED, {\Bem 27th International Conference on
  Principles and Practice of Constraint Programming (CP 2021)},
  \lowercase{\BVOL}\ 210 of {\Bem Leibniz International Proceedings in
  Informatics (LIPIcs)}, \BPGS\ 42:1--42:17.

\bibitem[\protect\BCAY{Mandi, Demirovi{\'c}, Stuckey,\ \BBA\ Guns}{Mandi
  et~al.}{2020}]{mandi2020smart}
Mandi, J., Demirovi{\'c}, E., Stuckey, P.~J., \BBA\ Guns, T. \BBOP2020\BBCP.
\newblock \BBOQ Smart predict-and-optimize for hard combinatorial optimization
  problems\BBCQ\
\newblock {\Bem Proceedings of the AAAI Conference on Artificial Intelligence},
  {\Bem 34\/}(02), 1603--1610.

\bibitem[\protect\BCAY{Mandi\ \BBA\ Guns}{Mandi\ \BBA\
  Guns}{2020}]{mandi2020interior}
Mandi, J.\BBACOMMA\  \BBA\ Guns, T. \BBOP2020\BBCP.
\newblock \BBOQ Interior point solving for lp-based
  prediction+optimisation\BBCQ\
\newblock In Larochelle, H., Ranzato, M., Hadsell, R., Balcan, M.~F., \BBA\
  Lin, H.\BEDS, {\Bem Advances in Neural Information Processing Systems},
  \lowercase{\BVOL}~33, \BPGS\ 7272--7282.

\bibitem[\protect\BCAY{Martins\ \BBA\ Astudillo}{Martins\ \BBA\
  Astudillo}{2016}]{martins2016softmax}
Martins, A.\BBACOMMA\  \BBA\ Astudillo, R. \BBOP2016\BBCP.
\newblock \BBOQ From softmax to sparsemax: A sparse model of attention and
  multi-label classification\BBCQ\
\newblock In {\Bem International conference on machine learning}, \BPGS\
  1614--1623. PMLR.

\bibitem[\protect\BCAY{Massart\ \BBA\ N{\'e}d{\'e}lec}{Massart\ \BBA\
  N{\'e}d{\'e}lec}{2006}]{massart2006risk}
Massart, P.\BBACOMMA\  \BBA\ N{\'e}d{\'e}lec, {\'E}. \BBOP2006\BBCP.
\newblock \BBOQ Risk bounds for statistical learning\BBCQ\
\newblock {\Bem The Annals of Statistics}, {\Bem 34\/}(5), 2326--2366.

\bibitem[\protect\BCAY{Merch\'{a}n, Arora, Pachon, Konduri, Winkenbach, Parks,\
  \BBA\ Noszek}{Merch\'{a}n et~al.}{2024}]{AmazonLastMile}
Merch\'{a}n, D., Arora, J., Pachon, J., Konduri, K., Winkenbach, M., Parks, S.,
  \BBA\ Noszek, J. \BBOP2024\BBCP.
\newblock \BBOQ 2021 amazon last mile routing research challenge: Data
  set\BBCQ\
\newblock {\Bem Transportation Science}, {\Bem 58\/}(1), 8--11.

\bibitem[\protect\BCAY{Minervini, Franceschi,\ \BBA\ Niepert}{Minervini
  et~al.}{2023}]{AIMLE}
Minervini, P., Franceschi, L., \BBA\ Niepert, M. \BBOP2023\BBCP.
\newblock \BBOQ Adaptive perturbation-based gradient estimation for discrete
  latent variable models\BBCQ\
\newblock {\Bem Proceedings of the AAAI Conference on Artificial Intelligence},
  {\Bem 37\/}(8), 9200--9208.

\bibitem[\protect\BCAY{Mi\v{s}i\'{c}\ \BBA\ Perakis}{Mi\v{s}i\'{c}\ \BBA\
  Perakis}{2020}]{MisicINFORMS}
Mi\v{s}i\'{c}, V.~V.\BBACOMMA\  \BBA\ Perakis, G. \BBOP2020\BBCP.
\newblock \BBOQ Data analytics in operations management: A review\BBCQ\
\newblock {\Bem Manufacturing \& Service Operations Management}, {\Bem
  22\/}(1), 158--169.

\bibitem[\protect\BCAY{Mnih\ \BBA\ Teh}{Mnih\ \BBA\ Teh}{2012}]{icml/MnihT12}
Mnih, A.\BBACOMMA\  \BBA\ Teh, Y.~W. \BBOP2012\BBCP.
\newblock \BBOQ A fast and simple algorithm for training neural probabilistic
  language models\BBCQ\
\newblock In {\Bem Proceedings of the 29th International Conference on Machine
  Learning, {ICML} 2012, Edinburgh, Scotland, UK, June 26 - July 1, 2012}.
  icml.cc / Omnipress.

\bibitem[\protect\BCAY{Monga, Li,\ \BBA\ Eldar}{Monga
  et~al.}{2021}]{monga2021algorithm}
Monga, V., Li, Y., \BBA\ Eldar, Y.~C. \BBOP2021\BBCP.
\newblock \BBOQ Algorithm unrolling: Interpretable, efficient deep learning for
  signal and image processing\BBCQ\
\newblock {\Bem IEEE Signal Processing Magazine}, {\Bem 38\/}(2), 18--44.

\bibitem[\protect\BCAY{Mulamba, Mandi, Diligenti, Lombardi, Bucarey,\ \BBA\
  Guns}{Mulamba et~al.}{2021}]{mulamba2020discrete}
Mulamba, M., Mandi, J., Diligenti, M., Lombardi, M., Bucarey, V., \BBA\ Guns,
  T. \BBOP2021\BBCP.
\newblock \BBOQ Contrastive losses and solution caching for
  predict-and-optimize\BBCQ\
\newblock In Zhou, Z.-H.\BED, {\Bem Proceedings of the Thirtieth International
  Joint Conference on Artificial Intelligence, {IJCAI-21}}, \BPGS\ 2833--2840.
  International Joint Conferences on Artificial Intelligence Organization.

\bibitem[\protect\BCAY{Nandwani, Ranjan, Mausam,\ \BBA\ Singla}{Nandwani
  et~al.}{2022}]{nandwani2022a}
Nandwani, Y., Ranjan, R., Mausam, \BBA\ Singla, P. \BBOP2022\BBCP.
\newblock \BBOQ A solver-free framework for scalable learning in neural ilp
  architectures\BBCQ\
\newblock In Koyejo, S., Mohamed, S., Agarwal, A., Belgrave, D., Cho, K., \BBA\
  Oh, A.\BEDS, {\Bem Advances in Neural Information Processing Systems},
  \lowercase{\BVOL}~35, \BPGS\ 7972--7986. Curran Associates, Inc.

\bibitem[\protect\BCAY{Nemirovski}{Nemirovski}{2007}]{nemirovski2007advances}
Nemirovski, A. \BBOP2007\BBCP.
\newblock \BBOQ Advances in convex optimization: conic programming\BBCQ\
\newblock In {\Bem International Congress of Mathematicians},
  \lowercase{\BVOL}~1, \BPGS\ 413--444.

\bibitem[\protect\BCAY{Niepert, Minervini,\ \BBA\ Franceschi}{Niepert
  et~al.}{2021}]{niepert2021implicit}
Niepert, M., Minervini, P., \BBA\ Franceschi, L. \BBOP2021\BBCP.
\newblock \BBOQ Implicit mle: Backpropagating through discrete exponential
  family distributions\BBCQ\
\newblock In Ranzato, M., Beygelzimer, A., Dauphin, Y., Liang, P., \BBA\
  Vaughan, J.~W.\BEDS, {\Bem Advances in Neural Information Processing
  Systems}, \lowercase{\BVOL}~34, \BPGS\ 14567--14579. Curran Associates, Inc.

\bibitem[\protect\BCAY{Ogryczak\ \BBA\ {\'S}liwi{\'n}ski}{Ogryczak\ \BBA\
  {\'S}liwi{\'n}ski}{2003}]{ogryczak2003solving}
Ogryczak, W.\BBACOMMA\  \BBA\ {\'S}liwi{\'n}ski, T. \BBOP2003\BBCP.
\newblock \BBOQ On solving linear programs with the ordered weighted averaging
  objective\BBCQ\
\newblock {\Bem European Journal of Operational Research}, {\Bem 148\/}(1),
  80--91.

\bibitem[\protect\BCAY{Papandreou\ \BBA\ Yuille}{Papandreou\ \BBA\
  Yuille}{2011}]{PapandreouY11}
Papandreou, G.\BBACOMMA\  \BBA\ Yuille, A.~L. \BBOP2011\BBCP.
\newblock \BBOQ Perturb-and-map random fields: Using discrete optimization to
  learn and sample from energy models\BBCQ\
\newblock In Metaxas, D.~N., Quan, L., Sanfeliu, A., \BBA\ Gool, L.~V.\BEDS,
  {\Bem {IEEE} International Conference on Computer Vision, {ICCV} 2011,
  Barcelona, Spain, November 6-13, 2011}, \BPGS\ 193--200. {IEEE} Computer
  Society.

\bibitem[\protect\BCAY{Paszke, Gross, Massa, Lerer, Bradbury, Chanan, Killeen,
  Lin, Gimelshein, Antiga, et~al.}{Paszke et~al.}{2019}]{paszke2019pytorch}
Paszke, A., Gross, S., Massa, F., Lerer, A., Bradbury, J., Chanan, G., Killeen,
  T., Lin, Z., Gimelshein, N., Antiga, L., et~al. \BBOP2019\BBCP.
\newblock \BBOQ Pytorch: An imperative style, high-performance deep learning
  library\BBCQ\
\newblock {\Bem Advances in neural information processing systems}, {\Bem 32}.

\bibitem[\protect\BCAY{Paulus, Rol{\'\i}nek, Musil, Amos,\ \BBA\
  Martius}{Paulus et~al.}{2021}]{paulus2021comboptnet}
Paulus, A., Rol{\'\i}nek, M., Musil, V., Amos, B., \BBA\ Martius, G.
  \BBOP2021\BBCP.
\newblock \BBOQ Comboptnet: Fit the right np-hard problem by learning integer
  programming constraints\BBCQ\
\newblock In {\Bem International Conference on Machine Learning}, \BPGS\
  8443--8453. PMLR.

\bibitem[\protect\BCAY{Paulus, Choi, Tarlow, Krause,\ \BBA\ Maddison}{Paulus
  et~al.}{2020}]{SSTPaulusCT0M20}
Paulus, M., Choi, D., Tarlow, D., Krause, A., \BBA\ Maddison, C.~J.
  \BBOP2020\BBCP.
\newblock \BBOQ Gradient estimation with stochastic softmax tricks\BBCQ\
\newblock In Larochelle, H., Ranzato, M., Hadsell, R., Balcan, M., \BBA\ Lin,
  H.\BEDS, {\Bem Advances in Neural Information Processing Systems},
  \lowercase{\BVOL}~33, \BPGS\ 5691--5704. Curran Associates, Inc.

\bibitem[\protect\BCAY{Perron\ \BBA\ Furnon}{Perron\ \BBA\
  Furnon}{2020}]{ortools}
Perron, L.\BBACOMMA\  \BBA\ Furnon, V. \BBOP2020\BBCP.
\newblock \BBOQ Or-tools\BBCQ.

\bibitem[\protect\BCAY{Pflug\ \BBA\ Pichler}{Pflug\ \BBA\
  Pichler}{2014}]{pflug2014multistage}
Pflug, G.~C.\BBACOMMA\  \BBA\ Pichler, A. \BBOP2014\BBCP.
\newblock {\Bem Multistage stochastic optimization}, \lowercase{\BVOL}\ 1104.
\newblock Springer.

\bibitem[\protect\BCAY{Pisinger}{Pisinger}{2005}]{PISINGER20052271}
Pisinger, D. \BBOP2005\BBCP.
\newblock \BBOQ Where are the hard knapsack problems?\BBCQ\
\newblock {\Bem Computers \& Operations Research}, {\Bem 32\/}(9), 2271--2284.

\bibitem[\protect\BCAY{Pisinger\ \BBA\ Toth}{Pisinger\ \BBA\
  Toth}{1998}]{pisinger1998knapsack}
Pisinger, D.\BBACOMMA\  \BBA\ Toth, P. \BBOP1998\BBCP.
\newblock \BBOQ Knapsack problems\BBCQ\
\newblock In {\Bem Handbook of combinatorial optimization}, \BPGS\ 299--428.
  Springer.

\bibitem[\protect\BCAY{Pogančić, Paulus, Musil, Martius,\ \BBA\
  Rolinek}{Pogančić et~al.}{2020}]{PogancicPMMR20}
Pogančić, M.~V., Paulus, A., Musil, V., Martius, G., \BBA\ Rolinek, M.
  \BBOP2020\BBCP.
\newblock \BBOQ Differentiation of blackbox combinatorial solvers\BBCQ\
\newblock In {\Bem International Conference on Learning Representations}.

\bibitem[\protect\BCAY{{PyTorch}}{{PyTorch}}{2017}]{ReduceLROnPlateau}
{PyTorch} \BBOP2017\BBCP.
\newblock \BBOQ Pytorch: Reducelronplateau — pytorch 1.9.0
  documentation\BBCQ.
\newblock
  \url{https://pytorch.org/docs/stable/generated/torch.optim.lr_scheduler.ReduceLROnPlateau.html\#torch.optim.lr_scheduler.ReduceLROnPlateau}.

\bibitem[\protect\BCAY{Qi, Grigas,\ \BBA\ Shen}{Qi
  et~al.}{2023}]{qi2023integrated}
Qi, M., Grigas, P., \BBA\ Shen, Z.-J.~M. \BBOP2023\BBCP.
\newblock \BBOQ Integrated conditional estimation-optimization\BBCQ.

\bibitem[\protect\BCAY{Qi\ \BBA\ Shen}{Qi\ \BBA\
  Shen}{2022}]{Qi2022Integrating}
Qi, M.\BBACOMMA\  \BBA\ Shen, Z.-J. \BBOP2022\BBCP.
\newblock {\Bem Integrating prediction/estimation and optimization with
  applications in operations management}, \BPGS\ 36--58.
\newblock INFORMS.

\bibitem[\protect\BCAY{Rezende, Mohamed,\ \BBA\ Wierstra}{Rezende
  et~al.}{2014}]{rezende14}
Rezende, D.~J., Mohamed, S., \BBA\ Wierstra, D. \BBOP2014\BBCP.
\newblock \BBOQ Stochastic backpropagation and approximate inference in deep
  generative models\BBCQ\
\newblock In Xing, E.~P.\BBACOMMA\  \BBA\ Jebara, T.\BEDS, {\Bem Proceedings of
  the 31st International Conference on Machine Learning}, \lowercase{\BVOL}~32
  of {\Bem Proceedings of Machine Learning Research}, \BPGS\ 1278--1286,
  Bejing, China. PMLR.

\bibitem[\protect\BCAY{Rolinek, Musil, Paulus, Vlastelica, Michaelis,\ \BBA\
  Martius}{Rolinek et~al.}{2020a}]{RolinekMPPMM20}
Rolinek, M., Musil, V., Paulus, A., Vlastelica, M., Michaelis, C., \BBA\
  Martius, G. \BBOP2020a\BBCP.
\newblock \BBOQ Optimizing rank-based metrics with blackbox
  differentiation\BBCQ\
\newblock In {\Bem Proceedings of the IEEE/CVF Conference on Computer Vision
  and Pattern Recognition (CVPR)}.

\bibitem[\protect\BCAY{Rol{\'{\i}}nek, Swoboda, Zietlow, Paulus, Musil,\ \BBA\
  Martius}{Rol{\'{\i}}nek et~al.}{2020b}]{RolinekSZPMM20}
Rol{\'{\i}}nek, M., Swoboda, P., Zietlow, D., Paulus, A., Musil, V., \BBA\
  Martius, G. \BBOP2020b\BBCP.
\newblock \BBOQ Deep graph matching via blackbox differentiation of
  combinatorial solvers\BBCQ\
\newblock In Vedaldi, A., Bischof, H., Brox, T., \BBA\ Frahm, J.\BEDS, {\Bem
  Computer Vision - {ECCV} 2020 - 16th European Conference, Glasgow, UK, August
  23-28, 2020, Proceedings, Part {XXVIII}}, \lowercase{\BVOL}\ 12373 of {\Bem
  Lecture Notes in Computer Science}, \BPGS\ 407--424. Springer.

\bibitem[\protect\BCAY{Rossi, van Beek,\ \BBA\ Walsh}{Rossi
  et~al.}{2006}]{rossi2006handbook}
Rossi, F., van Beek, P., \BBA\ Walsh, T.\BEDS. \BBOP2006\BBCP.
\newblock {\Bem Handbook of Constraint Programming}, \lowercase{\BVOL}~2 of
  {\Bem Foundations of Artificial Intelligence}.
\newblock Elsevier.

\bibitem[\protect\BCAY{Ruszczy{\'n}ski\ \BBA\ Shapiro}{Ruszczy{\'n}ski\ \BBA\
  Shapiro}{2003}]{stochasticprogrammingmodels}
Ruszczy{\'n}ski, A.\BBACOMMA\  \BBA\ Shapiro, A. \BBOP2003\BBCP.
\newblock \BBOQ Stochastic programming models\BBCQ\
\newblock {\Bem Handbooks in operations research and management science}, {\Bem
  10}, 1--64.

\bibitem[\protect\BCAY{Sadana, Chenreddy, Delage, Forel, Frejinger,\ \BBA\
  Vidal}{Sadana et~al.}{2024}]{sadana2023survey}
Sadana, U., Chenreddy, A., Delage, E., Forel, A., Frejinger, E., \BBA\ Vidal,
  T. \BBOP2024\BBCP.
\newblock \BBOQ A survey of contextual optimization methods for decision-making
  under uncertainty\BBCQ\
\newblock In {\Bem European Journal of Operational Research}.

\bibitem[\protect\BCAY{Sahinidis}{Sahinidis}{2004}]{SAHINIDIS2004971}
Sahinidis, N.~V. \BBOP2004\BBCP.
\newblock \BBOQ Optimization under uncertainty: state-of-the-art and
  opportunities\BBCQ\
\newblock {\Bem Computers \& Chemical Engineering}, {\Bem 28\/}(6), 971--983.

\bibitem[\protect\BCAY{Sahoo, Paulus, Vlastelica, Musil, Kuleshov,\ \BBA\
  Martius}{Sahoo et~al.}{2023}]{sahoo2022gradient}
Sahoo, S.~S., Paulus, A., Vlastelica, M., Musil, V., Kuleshov, V., \BBA\
  Martius, G. \BBOP2023\BBCP.
\newblock \BBOQ Backpropagation through combinatorial algorithms: Identity with
  projection works\BBCQ\
\newblock In {\Bem The Eleventh International Conference on Learning
  Representations}.

\bibitem[\protect\BCAY{Sang, Xu, Long, Hu,\ \BBA\ Sun}{Sang
  et~al.}{2022}]{sang2022electricity}
Sang, L., Xu, Y., Long, H., Hu, Q., \BBA\ Sun, H. \BBOP2022\BBCP.
\newblock \BBOQ Electricity price prediction for energy storage system
  arbitrage: A decision-focused approach\BBCQ\
\newblock {\Bem IEEE Transactions on Smart Grid}, {\Bem 13\/}(4), 2822--2832.

\bibitem[\protect\BCAY{Sen, Namata, Bilgic, Getoor, Galligher,\ \BBA\
  Eliassi-Rad}{Sen et~al.}{2008}]{cora2008}
Sen, P., Namata, G., Bilgic, M., Getoor, L., Galligher, B., \BBA\ Eliassi-Rad,
  T. \BBOP2008\BBCP.
\newblock \BBOQ Collective classification in network data\BBCQ\
\newblock {\Bem AI Magazine}, {\Bem 29\/}(3), 93.

\bibitem[\protect\BCAY{Settles}{Settles}{2009}]{settles.tr09}
Settles, B. \BBOP2009\BBCP.
\newblock \BBOQ Active learning literature survey\BBCQ\
\newblock Computer sciences technical report\ 1648, University of
  Wisconsin--Madison.

\bibitem[\protect\BCAY{Shah, Wang, Wilder, Perrault,\ \BBA\ Tambe}{Shah
  et~al.}{2022}]{shahdecision}
Shah, S., Wang, K., Wilder, B., Perrault, A., \BBA\ Tambe, M. \BBOP2022\BBCP.
\newblock \BBOQ Decision-focused learning without decision-making: Learning
  locally optimized decision losses\BBCQ\
\newblock In Oh, A.~H., Agarwal, A., Belgrave, D., \BBA\ Cho, K.\BEDS, {\Bem
  Advances in Neural Information Processing Systems}.

\bibitem[\protect\BCAY{Shah, Wilder, Perrault,\ \BBA\ Tambe}{Shah
  et~al.}{2024}]{leavingthenest}
Shah, S., Wilder, B., Perrault, A., \BBA\ Tambe, M. \BBOP2024\BBCP.
\newblock \BBOQ Leaving the nest: Going beyond local loss functions for
  predict-then-optimize\BBCQ\
\newblock In {\Bem Proceedings of the AAAI Conference on Artificial
  Intelligence}, \lowercase{\BVOL}~38, \BPGS\ 14902--14909.

\bibitem[\protect\BCAY{Silvestri, Berden, Mandi, İrfan Mahmutoğulları,
  Mulamba, Filippo, Guns,\ \BBA\ Lombardi}{Silvestri
  et~al.}{2023}]{silvestri2023score}
Silvestri, M., Berden, S., Mandi, J., İrfan Mahmutoğulları, A., Mulamba, M.,
  Filippo, A.~D., Guns, T., \BBA\ Lombardi, M. \BBOP2023\BBCP.
\newblock \BBOQ Score function gradient estimation to widen the applicability
  of decision-focused learning\BBCQ.

\bibitem[\protect\BCAY{Simonis, O’Sullivan, Mehta, Hurley,\ \BBA\
  Cauwer}{Simonis et~al.}{1999}]{csplib:prob059}
Simonis, H., O’Sullivan, B., Mehta, D., Hurley, B., \BBA\ Cauwer, M.~D.
  \BBOP1999\BBCP.
\newblock \BBOQ {CSPLib} problem 059: Energy-cost aware scheduling\BBCQ\
\newblock \url{http://www.csplib.org/Problems/prob059}.

\bibitem[\protect\BCAY{Singh\ \BBA\ Joachims}{Singh\ \BBA\
  Joachims}{2018}]{singh2018fairness}
Singh, A.\BBACOMMA\  \BBA\ Joachims, T. \BBOP2018\BBCP.
\newblock \BBOQ Fairness of exposure in rankings\BBCQ\
\newblock In {\Bem Proceedings of the 24th ACM SIGKDD international conference
  on knowledge discovery \& data mining}, \BPGS\ 2219--2228.

\bibitem[\protect\BCAY{Sun, Liu,\ \BBA\ Li}{Sun
  et~al.}{2023}]{NEURIPS2023_PredictthenCalibrate}
Sun, C., Liu, L., \BBA\ Li, X. \BBOP2023\BBCP.
\newblock \BBOQ Predict-then-calibrate: A new perspective of robust contextual
  lp\BBCQ\
\newblock In Oh, A., Naumann, T., Globerson, A., Saenko, K., Hardt, M., \BBA\
  Levine, S.\BEDS, {\Bem Advances in Neural Information Processing Systems},
  \lowercase{\BVOL}~36, \BPGS\ 17713--17741. Curran Associates, Inc.

\bibitem[\protect\BCAY{Tan, Delong,\ \BBA\ Terekhov}{Tan
  et~al.}{2019}]{TanDT19}
Tan, Y., Delong, A., \BBA\ Terekhov, D. \BBOP2019\BBCP.
\newblock \BBOQ Deep inverse optimization\BBCQ\
\newblock In Rousseau, L.\BBACOMMA\  \BBA\ Stergiou, K.\BEDS, {\Bem Integration
  of Constraint Programming, Artificial Intelligence, and Operations Research -
  16th International Conference, {CPAIOR} 2019, Thessaloniki, Greece, June 4-7,
  2019, Proceedings}, \lowercase{\BVOL}\ 11494 of {\Bem Lecture Notes in
  Computer Science}, \BPGS\ 540--556. Springer.

\bibitem[\protect\BCAY{Tan, Terekhov,\ \BBA\ Delong}{Tan
  et~al.}{2020}]{TanNEURIPS2020}
Tan, Y., Terekhov, D., \BBA\ Delong, A. \BBOP2020\BBCP.
\newblock \BBOQ Learning linear programs from optimal decisions\BBCQ\
\newblock In Larochelle, H., Ranzato, M., Hadsell, R., Balcan, M., \BBA\ Lin,
  H.\BEDS, {\Bem Advances in Neural Information Processing Systems},
  \lowercase{\BVOL}~33, \BPGS\ 19738--19749.

\bibitem[\protect\BCAY{Tang\ \BBA\ Khalil}{Tang\ \BBA\
  Khalil}{2023a}]{multi-task-DFL}
Tang, B.\BBACOMMA\  \BBA\ Khalil, E.~B. \BBOP2023a\BBCP.
\newblock \BBOQ Multi-task predict-then-optimize\BBCQ\
\newblock In Sellmann, M.\BBACOMMA\  \BBA\ Tierney, K.\BEDS, {\Bem Learning and
  Intelligent Optimization}, \BPGS\ 506--522. Springer International
  Publishing.

\bibitem[\protect\BCAY{Tang\ \BBA\ Khalil}{Tang\ \BBA\ Khalil}{2023b}]{pyepo}
Tang, B.\BBACOMMA\  \BBA\ Khalil, E.~B. \BBOP2023b\BBCP.
\newblock \BBOQ Pyepo: A pytorch-based end-to-end predict-then-optimize library
  for linear and integer programming\BBCQ.

\bibitem[\protect\BCAY{Tian, Yan, Liu,\ \BBA\ Wang}{Tian
  et~al.}{2023}]{TIAN202332}
Tian, X., Yan, R., Liu, Y., \BBA\ Wang, S. \BBOP2023\BBCP.
\newblock \BBOQ A smart predict-then-optimize method for targeted and
  cost-effective maritime transportation\BBCQ\
\newblock {\Bem Transportation Research Part B: Methodological}, {\Bem 172},
  32--52.

\bibitem[\protect\BCAY{Toth\ \BBA\ Vigo}{Toth\ \BBA\
  Vigo}{2015}]{toth2014vehicle}
Toth, P.\BBACOMMA\  \BBA\ Vigo, D. \BBOP2015\BBCP.
\newblock {\Bem Vehicle routing: Problems, methods, and applications}.
\newblock Society for Industrial and Applied Mathematics.

\bibitem[\protect\BCAY{Tschiatschek, Sahin,\ \BBA\ Krause}{Tschiatschek
  et~al.}{2018}]{ijcaiTschiatschekS018}
Tschiatschek, S., Sahin, A., \BBA\ Krause, A. \BBOP2018\BBCP.
\newblock \BBOQ Differentiable submodular maximization\BBCQ\
\newblock In Lang, J.\BED, {\Bem Proceedings of the Twenty-Seventh
  International Joint Conference on Artificial Intelligence, {IJCAI} 2018, July
  13-19, 2018, Stockholm, Sweden}, \BPGS\ 2731--2738. ijcai.org.

\bibitem[\protect\BCAY{Tschora, Guns, Pierre, Plantevit,\ \BBA\
  Robardet}{Tschora et~al.}{2023}]{TschoraPriceForecasting}
Tschora, L., Guns, T., Pierre, E., Plantevit, M., \BBA\ Robardet, C.
  \BBOP2023\BBCP.
\newblock \BBOQ Electricity price forecasting based on order books: a
  differentiable optimization approach\BBCQ\
\newblock In {\Bem 2023 IEEE 10th International Conference on Data Science and
  Advanced Analytics (DSAA)}, \BPGS\ 1--10.

\bibitem[\protect\BCAY{Vapnik}{Vapnik}{1999}]{Vapnik99}
Vapnik, V. \BBOP1999\BBCP.
\newblock \BBOQ An overview of statistical learning theory\BBCQ\
\newblock {\Bem {IEEE} Trans. Neural Networks}, {\Bem 10\/}(5), 988--999.

\bibitem[\protect\BCAY{Wahdany, Schmitt,\ \BBA\ Cremer}{Wahdany
  et~al.}{2023}]{WAHDANY2023109384}
Wahdany, D., Schmitt, C., \BBA\ Cremer, J.~L. \BBOP2023\BBCP.
\newblock \BBOQ More than accuracy: end-to-end wind power forecasting that
  optimises the energy system\BBCQ\
\newblock {\Bem Electric Power Systems Research}, {\Bem 221}, 109384.

\bibitem[\protect\BCAY{Wang, Shah, Chen, Perrault, Doshi-Velez,\ \BBA\
  Tambe}{Wang et~al.}{2021}]{wang2021learning}
Wang, K., Shah, S., Chen, H., Perrault, A., Doshi-Velez, F., \BBA\ Tambe, M.
  \BBOP2021\BBCP.
\newblock \BBOQ Learning mdps from features: Predict-then-optimize for
  sequential decision making by reinforcement learning\BBCQ\
\newblock {\Bem Advances in Neural Information Processing Systems}, {\Bem 34},
  8795--8806.

\bibitem[\protect\BCAY{Wang, Donti, Wilder,\ \BBA\ Kolter}{Wang
  et~al.}{2019}]{wang2019satnet}
Wang, P.-W., Donti, P., Wilder, B., \BBA\ Kolter, Z. \BBOP2019\BBCP.
\newblock \BBOQ Satnet: Bridging deep learning and logical reasoning using a
  differentiable satisfiability solver\BBCQ\
\newblock In {\Bem International Conference on Machine Learning}, \BPGS\
  6545--6554. PMLR.

\bibitem[\protect\BCAY{Wang, Zhang, Guo, Chen, Yang,\ \BBA\ Yan}{Wang
  et~al.}{2023}]{linsatnet}
Wang, R., Zhang, Y., Guo, Z., Chen, T., Yang, X., \BBA\ Yan, J. \BBOP2023\BBCP.
\newblock \BBOQ {L}in{SATN}et: The positive linear satisfiability neural
  networks\BBCQ\
\newblock In Krause, A., Brunskill, E., Cho, K., Engelhardt, B., Sabato, S.,
  \BBA\ Scarlett, J.\BEDS, {\Bem Proceedings of the 40th International
  Conference on Machine Learning}, \lowercase{\BVOL}\ 202 of {\Bem Proceedings
  of Machine Learning Research}, \BPGS\ 36605--36625. PMLR.

\bibitem[\protect\BCAY{Wilder, Dilkina,\ \BBA\ Tambe}{Wilder
  et~al.}{2019a}]{aaai/WilderDT19}
Wilder, B., Dilkina, B., \BBA\ Tambe, M. \BBOP2019a\BBCP.
\newblock \BBOQ Melding the data-decisions pipeline: Decision-focused learning
  for combinatorial optimization\BBCQ\
\newblock In {\Bem The Thirty-Third {AAAI} Conference on Artificial
  Intelligence, {AAAI} 2019, The Thirty-First Innovative Applications of
  Artificial Intelligence Conference, {IAAI} 2019, The Ninth {AAAI} Symposium
  on Educational Advances in Artificial Intelligence, {EAAI} 2019, Honolulu,
  Hawaii, USA, January 27 - February 1, 2019}, \BPGS\ 1658--1665. {AAAI} Press.

\bibitem[\protect\BCAY{Wilder, Ewing, Dilkina,\ \BBA\ Tambe}{Wilder
  et~al.}{2019b}]{WilderEDT19}
Wilder, B., Ewing, E., Dilkina, B., \BBA\ Tambe, M. \BBOP2019b\BBCP.
\newblock \BBOQ End to end learning and optimization on graphs\BBCQ\
\newblock In Wallach, H.~M., Larochelle, H., Beygelzimer, A.,
  d'Alch{\'{e}}{-}Buc, F., Fox, E.~B., \BBA\ Garnett, R.\BEDS, {\Bem Advances
  in Neural Information Processing Systems 32: Annual Conference on Neural
  Information Processing Systems 2019, NeurIPS 2019, December 8-14, 2019,
  Vancouver, BC, Canada}, \BPGS\ 4674--4685.

\bibitem[\protect\BCAY{Williams}{Williams}{1992}]{williams1992score}
Williams, R.~J. \BBOP1992\BBCP.
\newblock \BBOQ Simple statistical gradient-following algorithms for
  connectionist reinforcement learning\BBCQ\
\newblock {\Bem Mach. Learn.}, {\Bem 8}, 229--256.

\bibitem[\protect\BCAY{Wong, Tong, Zhang,\ \BBA\ Zhongbin}{Wong
  et~al.}{2020}]{wong2020fluid}
Wong, K.-K., Tong, K.-F., Zhang, Y., \BBA\ Zhongbin, Z. \BBOP2020\BBCP.
\newblock \BBOQ Fluid antenna system for 6g: When bruce lee inspires wireless
  communications\BBCQ\
\newblock {\Bem Electronics Letters}, {\Bem 56\/}(24), 1288--1290.

\bibitem[\protect\BCAY{Yang, Yan,\ \BBA\ Wang}{Yang
  et~al.}{2022}]{yang2022pairwise}
Yang, Y., Yan, R., \BBA\ Wang, H. \BBOP2022\BBCP.
\newblock \BBOQ Pairwise-comparison based semi-spo method for ship inspection
  planning in maritime transportation\BBCQ\
\newblock {\Bem Journal of Marine Science and Engineering}, {\Bem 10\/}(11),
  1696.

\bibitem[\protect\BCAY{Zou, Zhu,\ \BBA\ Hastie}{Zou et~al.}{2008}]{zou2008new}
Zou, H., Zhu, J., \BBA\ Hastie, T. \BBOP2008\BBCP.
\newblock \BBOQ New multicategory boosting algorithms based on multicategory
  fisher-consistent losses\BBCQ\
\newblock {\Bem The Annals of Applied Statistics}, {\Bem 2\/}(4), 1290.

\end{thebibliography}
\bibliographystyle{theapa}

\end{document}